\documentclass[acmtog,screen,authorversion,nonacm]{acmart}  

\AtBeginDocument{%
  \providecommand\BibTeX{{%
    \normalfont B\kern-0.5em{\scshape i\kern-0.25em b}\kern-0.8em\TeX}}}

\usepackage{cleveref}
\usepackage{subcaption}
\usepackage{wrapfig}
\usepackage{etoolbox}
\usepackage{multirow}
\usepackage{array}
\usepackage{stackengine}
\usepackage{graphicx}
\usepackage{tikz}
\usetikzlibrary{spy,calc}
\usepackage{xspace}
\newcommand{\ourmethod}[0]{\textit{PractiLight}\xspace}
\makeatletter
\patchcmd\WF@putfigmaybe{\lower\intextsep}{}{}{\fail}%
\AddToHook{env/wrapfigure/begin}{\setlength{\intextsep}{0pt} \setlength{\columnsep}{0pt}}
\makeatother
\usepackage{xcolor}
\Crefname{figure}{Fig.}{Figs.}
\Crefname{table}{Tab.}{Tabs.}
\Crefname{equation}{Eq.}{Eqs.}
\Crefname{section}{Sec.}{Secs.}

\begin{document}

\title[\ourmethod]{\ourmethod: Practical Light Control Using Foundational Diffusion Models}

\author{Yotam Erel}
\email{erelyotam@gmail.com}
\affiliation{%
  \institution{Tel Aviv University}
  \city{Tel Aviv}
  \country{Israel}
}
\author{Rishabh Dabral}
\affiliation{%
  \institution{Max Planck Institute for Informatics}
  \city{Saarbrücken}
  \country{Germany}
}
\author{Vladislav Golyanik}
\affiliation{%
  \institution{Max Planck Institute for Informatics}
  \city{Saarbrücken}
  \country{Germany}
}
\author{Amit H. Bermano}
\affiliation{%
  \institution{Tel Aviv University}
  \city{Tel Aviv}
  \country{Israel}
}
\author{Christian Theobalt}
\affiliation{%
  \institution{Max Planck Institute for Informatics}
  \city{Saarbrücken}
  \country{Germany}
}

\renewcommand{\shortauthors}{Erel et al.}

\begin{abstract}
Light control in generated images is a difficult task, posing specific challenges, spanning over the entire image and frequency spectrum. Most approaches tackle this problem by training on extensive yet domain-specific datasets, limiting the inherent generalization and applicability of the foundational backbones used. Instead, \ourmethod is a practical approach, effectively leveraging foundational understanding of recent generative models for the task. Our key insight is that lighting relationships in an image are similar in nature to token interaction in self-attention layers, and hence are best represented there. Based on this and other analyses regarding the importance of early diffusion iterations, \ourmethod trains a lightweight LoRA regressor to produce the direct-irradiance map for a given image, using a small set of training images. We then employ this regressor to incorporate the desired lighting into the generation process of another image using Classifier Guidance. This careful design generalizes well to diverse conditions and image domains. We demonstrate state-of-the-art performance in terms of quality and control with proven parameter and data efficiency compared to leading works over a wide variety of scene types. We hope this work affirms that image lighting can feasibly be controlled by tapping into foundational knowledge, enabling practical and general relighting.
\end{abstract}

\begin{teaserfigure}
  \centering
  \begin{subfigure}{.16\textwidth}
  \centering
  \stackinset{r}{0pt}{b}{0pt}{\color{white}\frame{\includegraphics[width=0.33\textwidth]{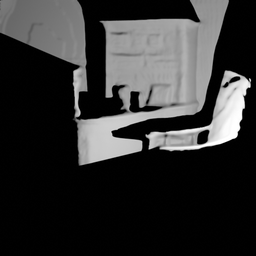}}}{\includegraphics[width=1.0\textwidth]{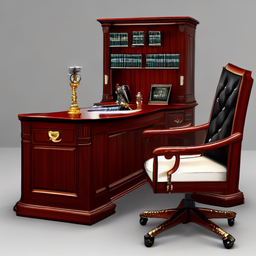}}
  \end{subfigure}
  \begin{subfigure}{.16\textwidth}
  \centering
  \stackinset{r}{0pt}{b}{0pt}{\color{white}\frame{\includegraphics[width=0.33\textwidth]{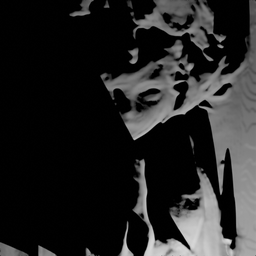}}}{\includegraphics[width=1.0\textwidth]{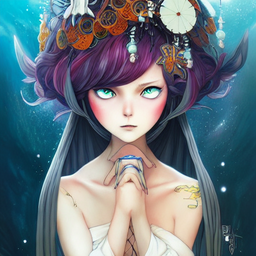}}
  \end{subfigure}
  \begin{subfigure}{.16\textwidth}
  \centering
  \stackinset{r}{0pt}{b}{0pt}{\color{white}\frame{\includegraphics[width=0.33\textwidth]{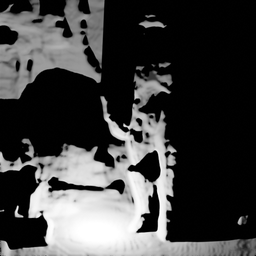}}}{\includegraphics[width=1.0\textwidth]{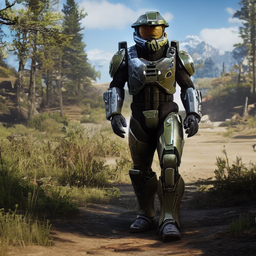}}
  \end{subfigure}
  \begin{subfigure}{.16\textwidth}
  \centering
  \stackinset{r}{0pt}{b}{0pt}{\color{white}\frame{\includegraphics[width=0.33\textwidth]{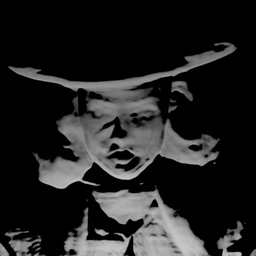}}}{\includegraphics[width=1.0\textwidth]{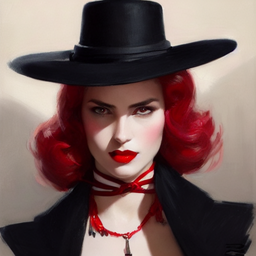}}
  \end{subfigure}
  \begin{subfigure}{.16\textwidth}
  \centering
  \stackinset{r}{0pt}{b}{0pt}{\color{white}\frame{\includegraphics[width=0.33\textwidth]{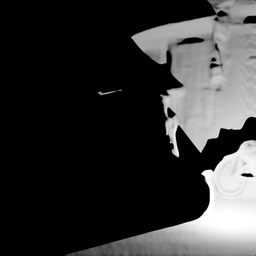}}}{\includegraphics[width=1.0\textwidth]{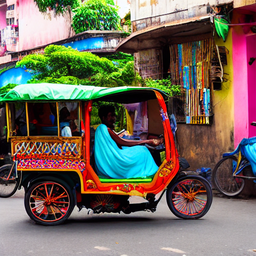}}
  \end{subfigure}
  \begin{subfigure}{.16\textwidth}
  \centering
    \stackinset{r}{0pt}{b}{0pt}{\color{white}\frame{\includegraphics[width=0.33\textwidth]{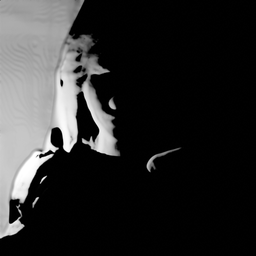}}}{\includegraphics[width=1.0\textwidth]{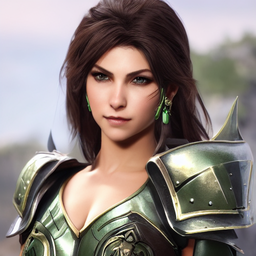}}
  \end{subfigure}
    \begin{subfigure}{.16\textwidth}
  \centering
    \includegraphics[width=1.0\textwidth]{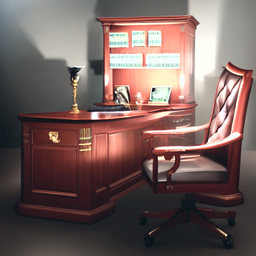}
  \end{subfigure}
    \begin{subfigure}{.16\textwidth}
  \centering
    \includegraphics[width=1.0\textwidth]{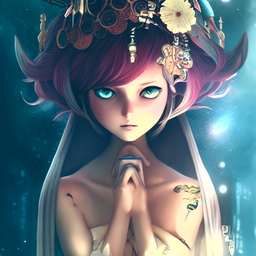}
  \end{subfigure}
    \begin{subfigure}{.16\textwidth}
  \centering
    \includegraphics[width=1.0\textwidth]{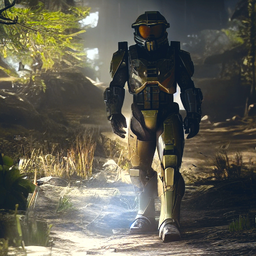}
  \end{subfigure}
  \begin{subfigure}{.16\textwidth}
  \centering
    \includegraphics[width=1.0\textwidth]{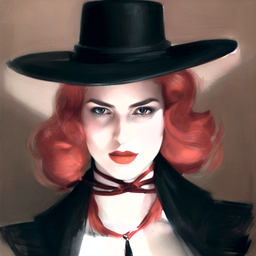}
  \end{subfigure}
  \begin{subfigure}{.16\textwidth}
  \centering
  \includegraphics[width=1.0\textwidth]{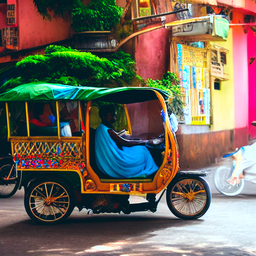}
  \end{subfigure}
  \begin{subfigure}{.16\textwidth}
  \centering
    \includegraphics[width=1.0\textwidth]{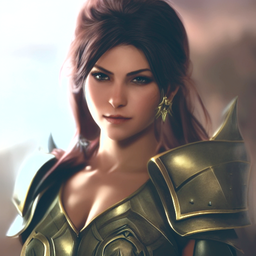}
  \end{subfigure}
  \caption{\textbf{Relighting using \ourmethod}. \textbf{Top:} original images (inset: light condition), \textbf{Bottom:} relighting results. Our method allows efficiently controlling light and shadows in generated images across diverse domains using light transport priors within pre-trained foundational diffusion models.}
  \label{fig:teaser}
\end{teaserfigure}

\maketitle

\section{Introduction}
\label{sec:intro}
Achieving plausible control over light in arbitrarily generated images is both compelling and desirable. It is a fundamental image editing tool essential for establishing the overall mood, tone and atmosphere of the scene, and also guiding local emphasis through shadows, highlights, and object-level illumination. However, it is far from solved; light has a dramatic influence over the entire image and its interactions involve complex relationships between elements in the scene, which must stay consistent in shape during relighting. Spanning all spatial image regions and frequencies, this differentiates lighting control from other control types or editing operations. For these reasons, most recent learning-based approaches tackle light control in generated images by requiring extensive and dedicated in-domain datasets, limiting generalization to unseen domains \cite{dreambooth} and real-world application due to intensive data collection \cite{iclight,lightlab}.
These include methods designed for specific domains such as portraits \cite{diffusion_face_relight, switchlight}, object-centric relighting \cite{dilightnet, neural-gaffer}, outdoor relighting \cite{outcast, outdoor_relight}, indoors \cite{latent_intrinsics, luminet} and recent intrinsic channel decomposition-based approaches \cite{rgbx, ordinal, cd-iid}.
Even when based on pretrained models that are capable of generalizing, such an intensive training process distorts the learned manifold distribution, harming its generalization capabilities (\Cref{sec:results}).

At the same time, the ever-growing power of foundational generative models suggests that light-related reasoning is already understood by these models to a great extent \cite{ilora, latent_intrinsics}.
Motivated by this, in this paper we present \ourmethod~ --- a practical approach to light control in generative art, effectively leveraging the implicit knowledge of light transport within recent foundational generative models. As we demonstrate, \ourmethod~ generalizes to diverse conditions and image domains with very little training and data (\Cref{fig:teaser}). Our key insight is that lighting relationships are of similar nature to the interactions encoded in the self-attention layers; analogously to the attention mechanism, the properties (e.g., color) of an illuminating pixel should be considered by other pixels if they are relevant to each other in terms of visibility and reflectance. As we demonstrate in our analysis (\Cref{sec:analysis}), the activations of the self-attention layers in a latent diffusion model are highly correlated with light phenomena, much more than other layers. We also show that light-related generation spans different timesteps starting very early on in the diffusion process. We speculate this is due to the wide range of frequencies affected by lighting conditions, starting from overall low-frequency light intensity and reflections, and up to high-frequency light phenomena such as hard shadow edges and caustics.

Building upon these insights, we design a generative image editing approach for light control (\Cref{fig:overview}). We first employ a diffusion model to extract the light features from a given image. To do this, we train a lightweight Low-Rank Adapter (LoRA \cite{lora}) to produce a direct-irradiance map based on the self-attention layers and specific iteration steps deduced from our analysis (\Cref{sec:analysis}). This adapter transforms the U-Net backbone of the diffusion model into an image-to-image-translator, that predicts the light features for any timestep directly, bootstrapped by the knowledge of the foundation model. Then, given a generated image and a new light condition as control, we use the adapter to guide the generation of a new image using Classifier Guidance \cite{dhariwal2021diffusion}, while preserving identity using self-attention query injection \cite{cia} and an off-the-shelf edge conditioned ControlNet \cite{controlnet}. The result depicts the desired scene geometry and style with the new lighting conditions.
Similar to \citet{dilightnet}, we control the light by predicting the depth of the scene using an off-the-shelf method \cite{zoe_depth}. Using this estimate, we lift the image to 3D and assign it a GGX material \cite{ggx}. One or more light probes are then added manually or automatically into this pseudo-scene, and a direct-irradiance map is rendered using ray casting. 
To train the LoRA, we create a modest set of $4000$ synthetic images, depicting simple scenes of basic geometric primitives, with random material and color properties.
Despite being simple, this process is sufficient for dramatic relighting effects including hard shadows and specular highlights editing.

Through comparisons with state-of-the-art diffusion-based lighting control methods, we demonstrate human subjects strongly prefer \ourmethod in terms of relighting quality and identity preservation, over a wide variety of scenes and image domains, with significantly better parameter and data efficiency. We hope this work affirms that image lighting can feasibly be controlled by tapping into foundational knowledge, and believe it paves the way to simple, intuitive, and high-quality light editing for novice end-users.

The code, models, and datasets used will be made publicly available at \url{https://yoterel.github.io/PractiLight-project-page/}.

\section{Related Studies}
\label{sec:related}
Relighting a single image is a popular field of research, as lighting is the foundation of scene creation and artistic expression. Since it is also a wide spanning task, most literature tackles specific domains.
\paragraph{Domain Specific} For specific content domains, we focus on related studies that address relighting of low dynamic range (LDR) imagery with image space control. For outdoors, typical techniques impose strong priors over the light sources, and achieve outstanding results (e.g., \cite{outcast, lightit, outdoor_relight}). For portrait and full body relighting, accurate results are achieved using a light stage \cite{light_stage}, or a large collection of portraits photos \cite{deep_portrait}. More recently, generative models trained over light stage datasets have been employed \cite{dir_light_estimator, switchlight, diffusion_face_relight}). Other works that use diffusion based models and focus on object-centric relighting were also shown to achieve state of the art performance \cite{dilightnet, neural-gaffer}. However, their performance in domains such as cel-shading, indoors, or surrealistic imagery significantly deteriorates. For indoors scenes, which are considered to be a challenging domain, similar recent progress using generative models is quite successful \cite{latent_intrinsics, retinex, scribble_light, luminet, lightlab}. That said, accurately relighting indoor scenes is still an active field of research, because of the wide variety of light interactions phenomena occurring in the scene. All of these works employ domain specific data, which requires impractical collection, and binds the capabilities to the specific domains. These techniques are fairly limited in generalization to different domains, and more often than not, limited in their ability to faithfully relight an entire scene. Even methods that do employ a powerful foundational model, still use large training sets, that distort the knowledge of the prior, and hurt its native generalization strength \cite{dreambooth}. The most relevant work to this paper employs a combination of light stage data, synthetic and in-the-wild images, trained over a foundational model \cite{iclight}. This work employs roughly $10$ million images by exploiting the linearity of light transport and achieving impressive results, especially for portraits, but as we demonstrate, inferior performance to our method in general. This is due to finetuning the entire model or a large copy of it, while we find a small subset of parameters that are important for the task at hand and only optimize them.

\paragraph{Generic Relighting} Addressing more complex and diverse scenes, recent image decomposition techniques train neural networks on vasts amount of data and allow (in theory) for realistically relighting a scene by separating a shading channel from the rest of the image \cite{ordinal, rgbx, cd-iid}. These techniques are very flexible in their application, but as we show, again suffer from lack of generalizability. In addition, these models are typically sensitive to the choice of the novel shading signal. More relevantly, \citet{ilora} repurposed the backbone U-Net model of pretrained diffusion models into extracting intrinsic channels, showing the capacity exists for this sort of task. However, this was done for a single timestep, drastically limiting the frequency bandwidth it needs to work with, and as we show is not suitable for the task of relighting (\Cref{sec:analysis}). In Sketch-Guidance \cite{sketch} as well as Readout Guidance \cite{readout_guidance}, the generation process is guided using any image modality with minimal data and memory requirements, by training small MLP-based regressor networks. The common issue is they are tailored to simpler spatial control types, such as pose, or edges, which does not fit the light editing setting as they do not deal well with identity preservation nor leverage existing priors efficiently. We adopt a similar guiding approach, with changes made using our light related insights (\Cref{sec:analysis}). Specifically, we have a parameter count reduction by a factor of 10, and substantially higher generation quality and identity preservation, even when trained on the same dataset (\Cref{sec:quantitative_results}). We believe our work fills an important gap of practical light control, as it requires very little data to train, operates on many different image domains out-of-the-box and, since it's adapter-based, integrates well with the ``eco-system'' of tools and editing capabilities in-use by artists and practitioners.

\section{Preliminaries}
\label{sec:classifier_guidance}
\label{sec:lora}
In this work, we hypothesize that foundational diffusion models~\cite{sd} possess a certain level of intrinsic understanding of light transport.
To effectively leverage this prior knowledge for our image relighting task, we use the following techniques, briefly revised below. 

\paragraph{Classifier Guidance}
Given a pretrained diffusion-based generative model that learned a score function $S_\theta = \nabla_x{log(p(x_t))}$, where $\theta$ are the learned model parameters, and $x_t$ are noisy latents at diffusion timestep $t$. 
A condition $y$ can be introduced using Bayes' theorem:
\vspace{-0.1in}
\begin{equation}
    \nabla_x{log(p(x|y))} = \overbrace{\nabla_x{log(p(x))}}^{S_\theta} + \overbrace{\nabla_x{log(p(y|x))}}^{E}
\label{eq:guidance}
\end{equation}
where $E$ is an energy term that can be used to \textit{guide} the generation process, usually in the form of a classifier. 
In our case $E$ is a regressor, trained on a continuous distribution.
Specifically, we train a regressor to translate a given input image $x$ into its direct-irradiance component $y$.
This regressor can then be used as an estimator of $p(y|x)$. More explicitly, if a model is tasked with noise prediction $\hat{\epsilon}$, we have:
\begin{equation}
    \hat{\epsilon} = (1+w)\epsilon_\theta(x_t, t, y,c) - w\epsilon_\theta(x_t,t,\varnothing,c) + \gamma_t E_t
\label{eq:update_rule}
\end{equation}
Where $\hat{\epsilon}$ is the final noise prediction, $\epsilon_\theta$ is the model prediction, $x_t$ is the current latents, $t$ is the current timestep, $y$ is text conditioning, $c$ is other forms of conditions if applicable (such as ControlNet), $w$ is the classifier-free guidance scale, $\gamma_t$ is the classifier guidance scale, and $E_t$ is the energy term as before, which could be timestep dependent in general.
\paragraph{LoRA}
Low Rank Adaptation (LoRA) \cite{lora} allows for fine tuning large models with minimal distortion by decreasing the rank of the updated parameters. Instead of updating the network weights $W$ using: $W' = W + \Delta W$, where $W,\Delta W\in \mathbb{R}^{m \times n}$, one can perform a decomposition: $\Delta W = BA$ where $\space B\in \mathbb{R}^{m \times r}, \space A\in \mathbb{R}^{r \times n}$, and $r << min(m,n)$. This allows for a controllable trade-off between parameter count and accuracy. LoRAs have been used in the context of image generation for concept learning \cite{concept_lora, concept_lora2}, style transferring \cite{ziplora}, personalization \cite{personalized_lora}, and many other downstream tasks.

\subsection{Light Transport Analysis}
\label{sec:analysis}
\begin{figure}[h]
    \centering
    \includegraphics[width=1.0\linewidth]{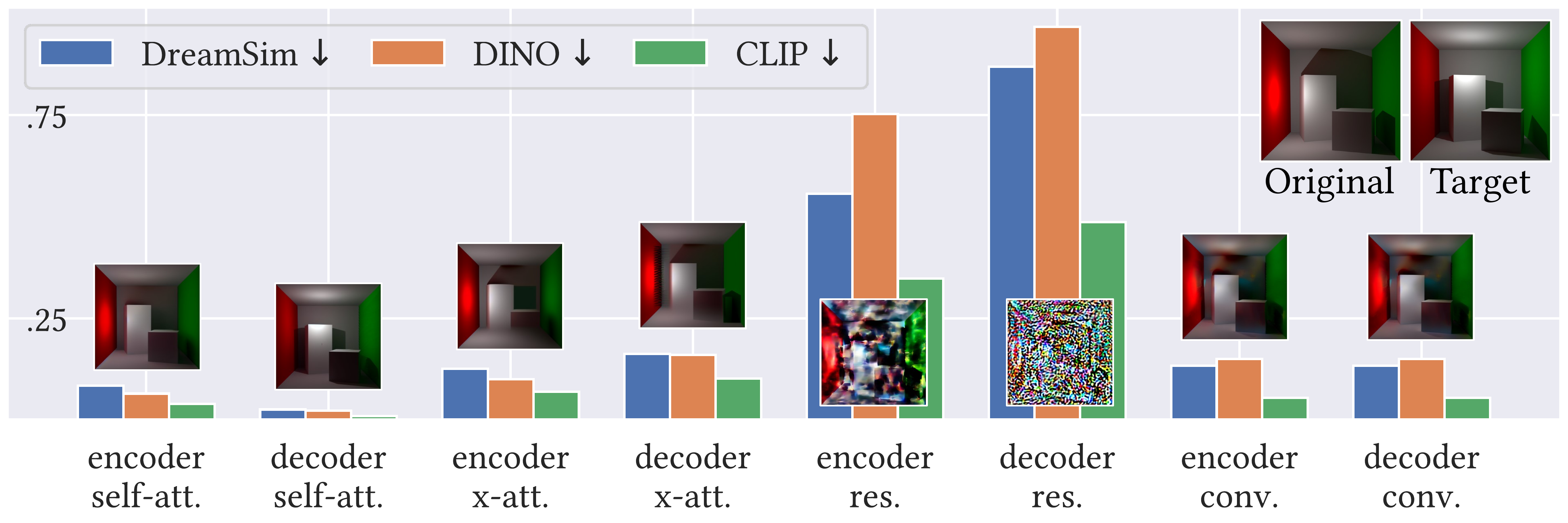} \\
    \caption{\textbf{Layer Type Analysis}. Generation quality is assessed while features from the \textit{Target} (top right) are directly injected into the \textit{Original} generation for different layer types. Self-attention layers performs best for all metrics (lower is better). Typical results per layer type are shown in insets.}
    \label{fig:layer_analysis}
\end{figure}
\begin{figure}[h]
    \centering
    {
    \setlength{\tabcolsep}{0pt} 
    \renewcommand{\arraystretch}{0.1} 
    
    \begin{subfigure}[T]{0.58\linewidth}
        \centering
        \begin{tabular}{*{3}{>{\centering\arraybackslash}m{0.33\textwidth}}}
    Generated & $t=900$ & $t=100$ \\
        \\[0.5em]
        \includegraphics[width=\linewidth]{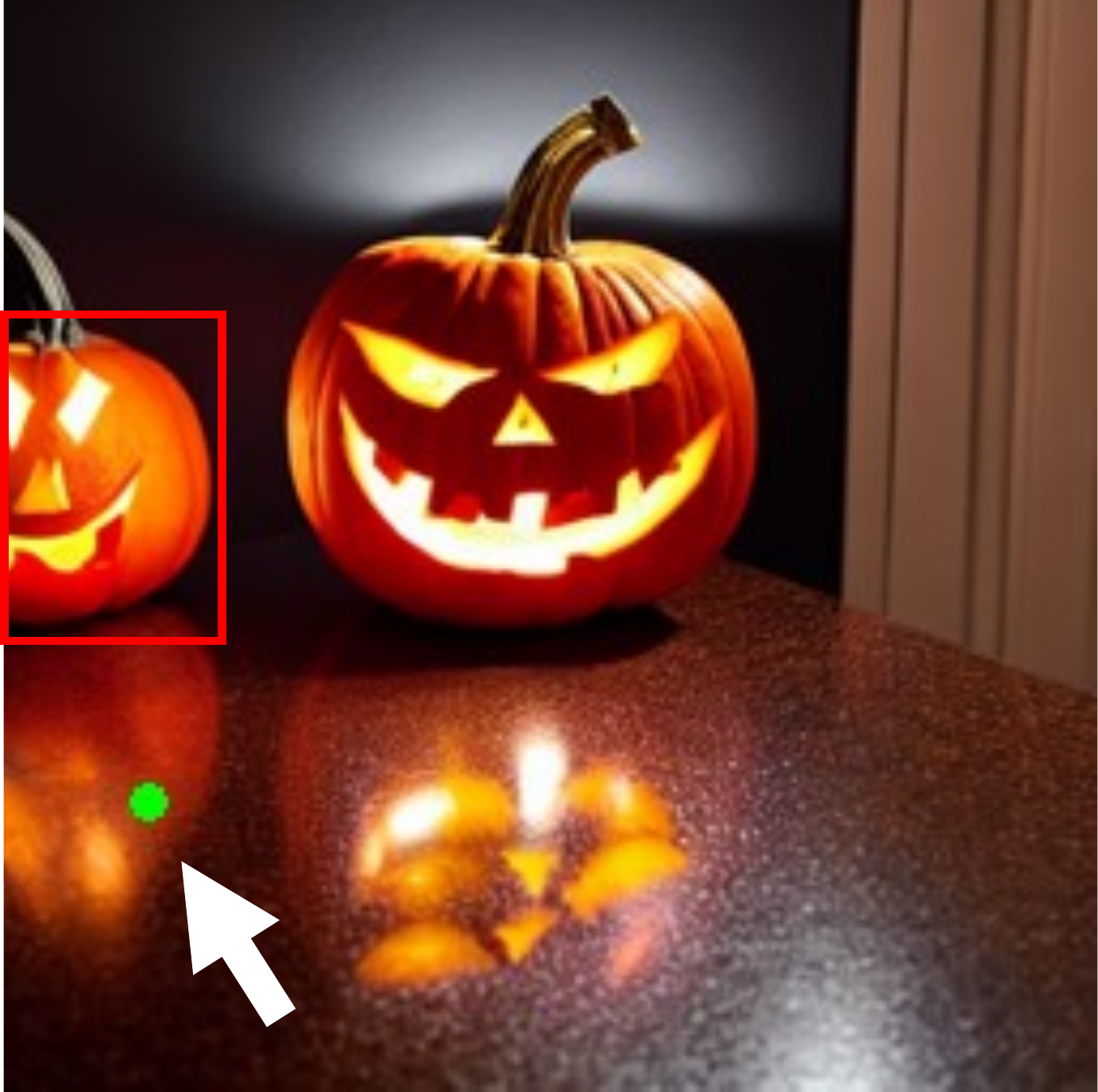} & 
        \includegraphics[width=\linewidth]{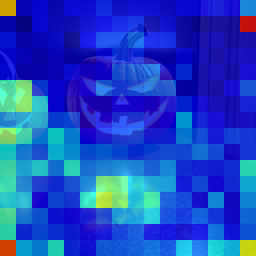} & 
        \includegraphics[width=\linewidth]{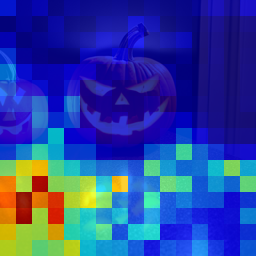} \\ 
        \includegraphics[width=\linewidth]{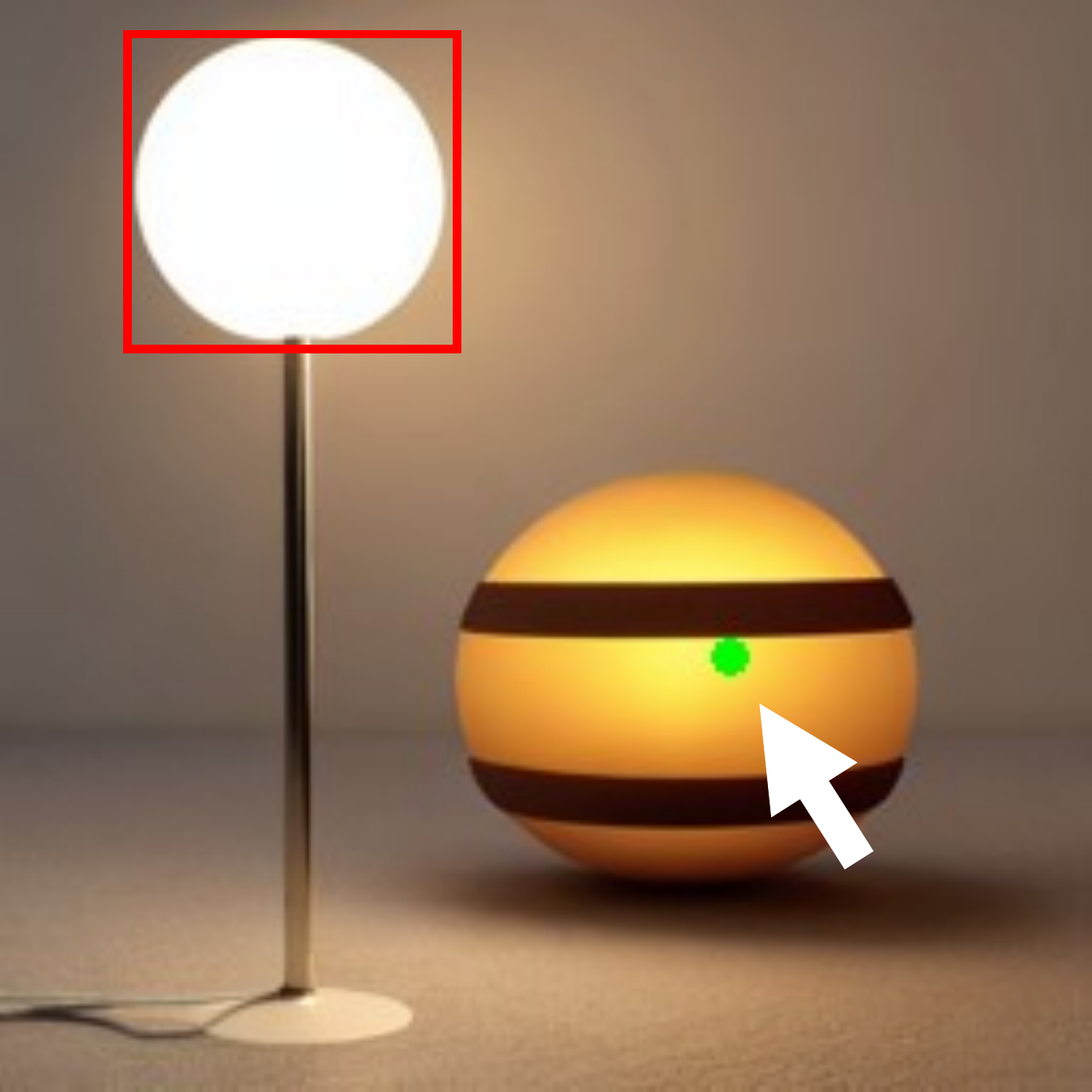} & 
        \includegraphics[width=\linewidth]{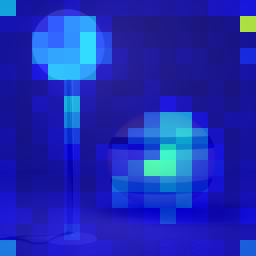} & 
        \includegraphics[width=\linewidth]{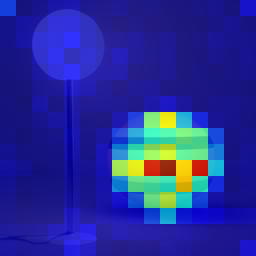}
    \end{tabular}
    \end{subfigure}%
    ~ 
    \begin{subfigure}[T]{0.42\linewidth}
        \centering
        {Attention vs Timesteps} \\ [0.35em]
        
        {\includegraphics[width=\linewidth]{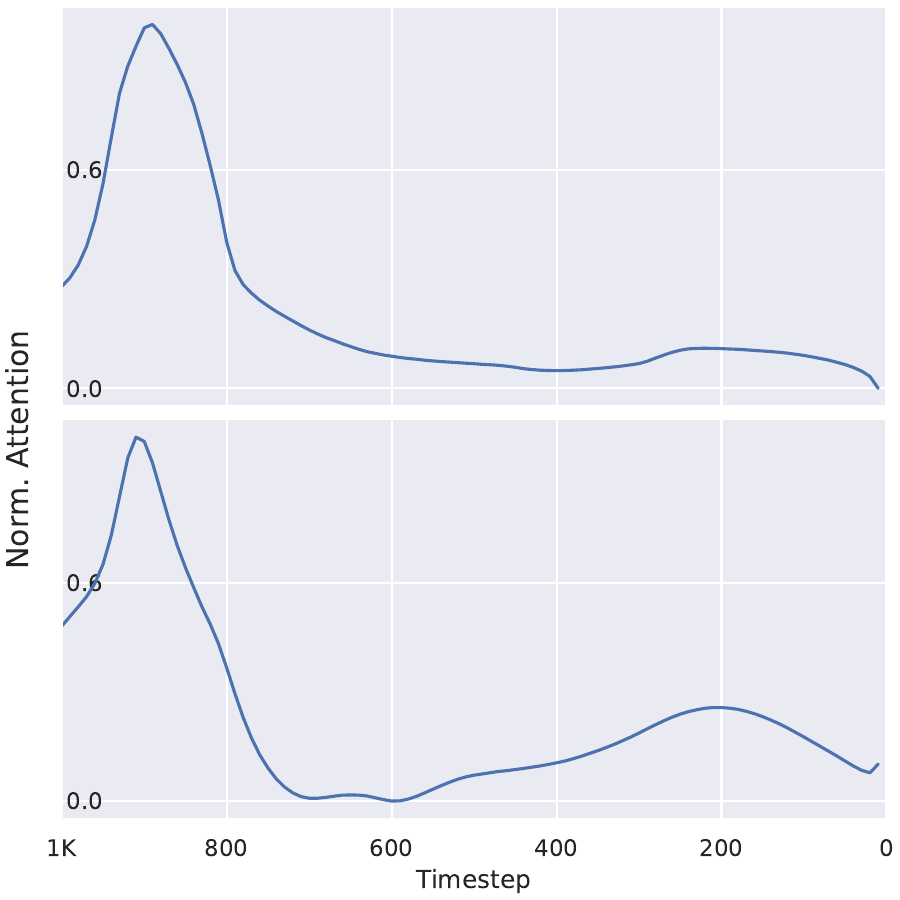}}
    \end{subfigure}
    \caption{\textbf{Attention and Timesteps}. \textbf{Left:} a generated image with a specific pixel marked in green, and a region of interest (ROI) in a red box (top: the source of reflection, bottom: a light source). \textbf{Middle:} The self-attention row corresponding to that pixel (1st attention block in decoder, last layer, SD1.5). The pixels attend the ROI in early timesteps (t=900), while attending more semantically similar areas later on (t=100). \textbf{Right:} the normalized attention given by the pixel to the ROI as a function of timesteps. Similar behavior was observed for different self-attention layers and scenes.}
    \label{fig:time_analysis}
    }
\end{figure}
Foundational generative models such as SD1.5 approximate aspects of light properties as part of their overall understanding of the world \cite{ilora}. However, fine-tuning all parameters of the model towards a new task can lead to loss of generality, and requires vasts amounts of data. 
This motivates our analysis to better understand where and how light transport is encoded in diffusion models.
\begin{figure*}[!t] 
  \centering 
  \includegraphics[width=1\linewidth]{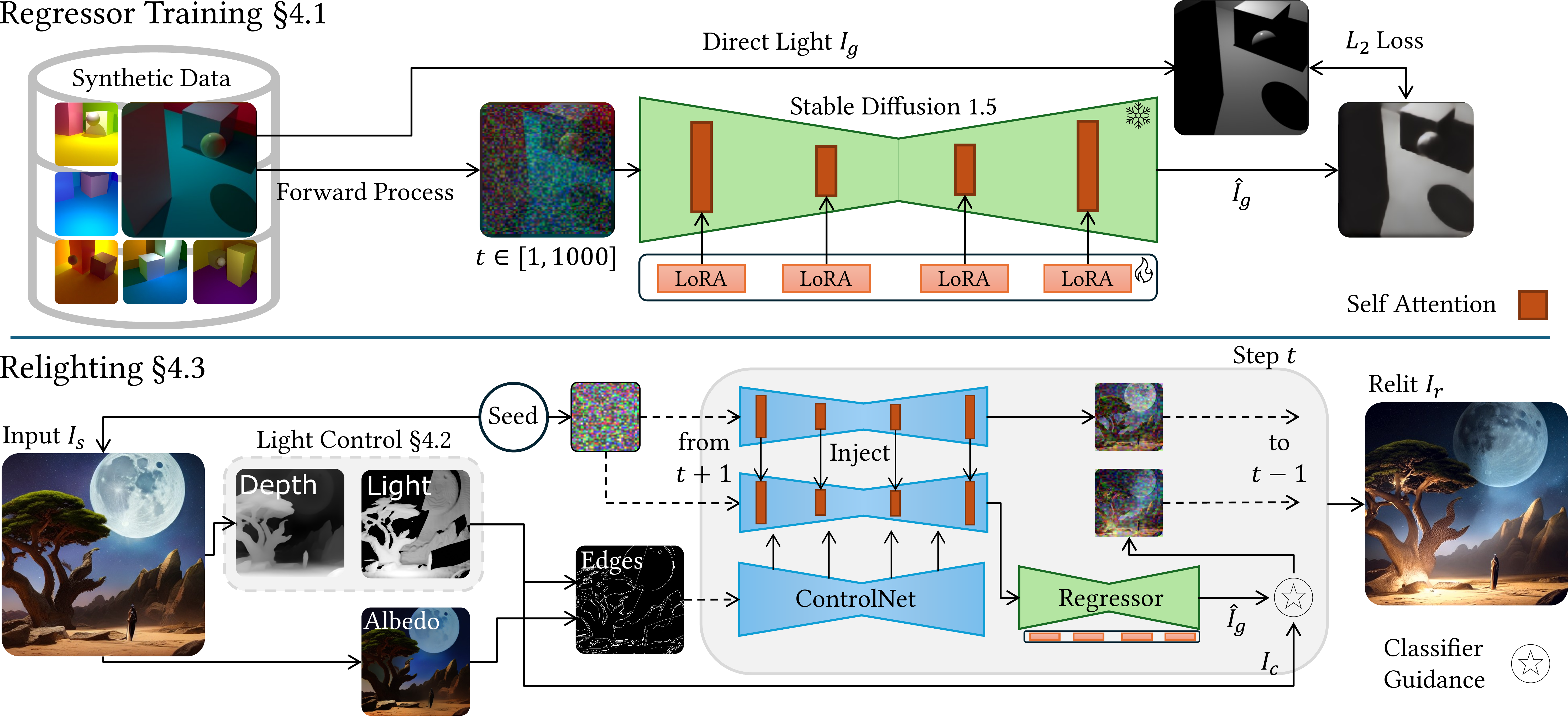} 
  \caption{\textbf{Overview of \ourmethod}. \textbf{Top:} we train a regressor on synthetic data (\Cref{sec:training}) to directly predict the clean direct-irradiance component $\hat{I}_g$ from images. \textbf{Bottom:} for relighting (\Cref{sec:relighting}), given a generated image $I_s$ we edit the light by applying classifier guidance using the regressor, and preserving identity using query injection and an edge-condition ControlNet. The user controls the light using a reconstructed pseudo-scene (\Cref{sec:lantern_control}) that makes use of a 3D renderer.} 
  \label{fig:overview} 
\end{figure*} 
\paragraph{Layer Analysis} For conciseness, consider a simple scene illuminated by two different light conditions \textit{Original} and \textit{Target} (\Cref{fig:layer_analysis}, top-right). 
We first perform DDIM inversion on both images~\cite{ddim}, to recover the latents that can reconstruct the images back. 
To investigate which components capture properties related to light transport, we inject activations from the reconstructed Target into the denoising process for reconstructing Original, and measure the quality of the \textit{Result} using semantic similarity metrics with respect to the Target (DreamSim \cite{dreamsim}, DINO \cite{dino}, and CLIP \cite{clip}).
Since the only difference between the initial images is the lighting, we argue that injecting activations should yield higher quality results when using layers that disentangle light better. 
We test this for different layer types in the SD1.5 backbone U-Net model. Results can be seen in \Cref{fig:layer_analysis}, where we performed this experiment over 100 different image pairs and all timesteps.

Indeed, the analysis suggests that \textit{self-attention layers encode features that are more associated with light transport}. We also observe that the self-attention layers in the decoder perform best under this experiment, perhaps due to their abstraction capabilities shown in other studies \cite{p2p, cia}. In the self-attention layers themselves, we also observe differences in performance between different heads, which indicates different heads have distinct roles when dealing with light (see supplement).

\paragraph{Time Step Analysis.} We also investigate the effects of light transport over different diffusion timesteps. 
In \Cref{fig:time_analysis}, two example scenes show that attention is given to different phenomena in different timesteps: we observe that regions depicting reflections (including diffuse and specular reflections) attend the relevant light sources in the beginning of the diffusion process, and transition to attending more semantically similar regions as the process evolves, roughly until half the total number of steps. This aligns with the idea that diffusion models resolve of low-frequency components (e.g., shading) earlier than high-frequencies details (e.g., fine textures). Therefore, \textit{when dealing with lighting, editing should begin as early as possible but cease around the midpoint}, as further changes become ineffective and risk degrading the image rather than enhancing the lighting (see supplemental material for an example).

\section{Method} 
Given a prompt generated source image $I_s$ and a new light condition given as another image $I_c$, \ourmethod synthesizes the relit image $I_r$. Our full pipeline is shown in \Cref{fig:overview}.
Our implementation is based on training a lightweight adapter (\Cref{sec:training}) that is tasked with generating an estimate of the direct-irradiance component $\hat{I}_g$, given $I_s$. Our core premise is that a low-rank adapter unlocks the foundational knowledge already encoded in the backbone it is applied to (Stable Diffusion 1.5 in our case), instead of accumulating knowledge on its own, or overfitting to the training examples' domain. This differentiable regressor is later used to guide the relighting process by defining a guidance objective between $\hat{I}_g$ and $I_c$. For control, we create $I_c$ by constructing a synthetic 3D scene from $I_s$ (\Cref{sec:lantern_control}), followed by a detailed description of the relighting process that leverages the regressor to create the relit image $I_r$ (\Cref{sec:relighting}).

\subsection{Regressor Training}
\label{sec:training}

\begin{figure*}[ht] 
  \centering 
  \includegraphics[width=1\linewidth]{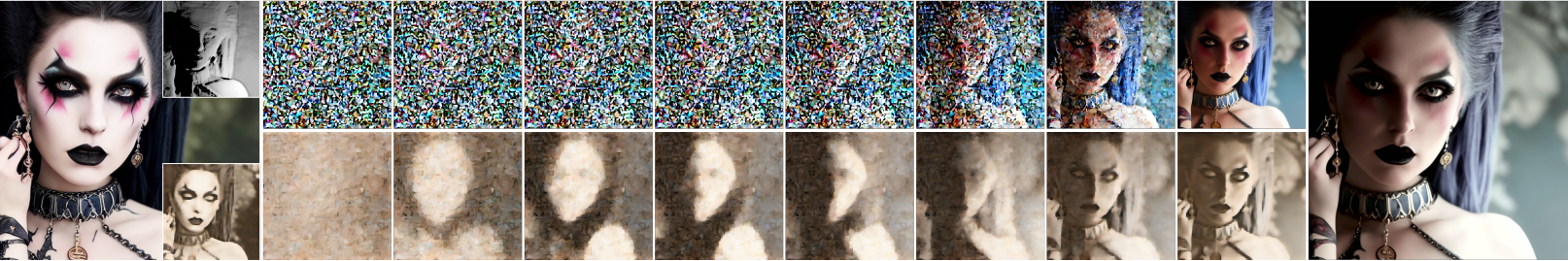} 
  \caption{\textbf{Relighting Process}. \textbf{Left:} a generated image and in inset the control signal (top) and the raw prediction of the direct-irradiance map from the regressor $\hat{I}_g$ (bottom). \textbf{Middle:} denoising using our method (top) and the regressor predictions for those timesteps (bottom). \textbf{Right:} the result after color-correction with original image statistics.} 
  \label{fig:diffusion_process} 
\end{figure*}

We start by training a regressor (\Cref{fig:overview}, top), used solely for defining an energy term for classifier guidance (\Cref{eq:update_rule}). This regressor is trained to extract an estimate of the direct-irradiance component $\hat{I}_g$ from images, and is trained over a synthetic dataset.

The synthetic dataset is created using Blender \cite{blender}, using three 3D primitives (sphere, cube, cuboid) and a single point light source, placed in a large cube-like room. The location of the shapes, light and camera is randomized, avoiding any intersections. We also randomize the GGX material roughness and albedo parameters of the shapes, as well as the light intensity. Each scene is rendered regularly to create $I_s$, and a separate render pass is performed for the direct diffuse and direct specular channels (added together) to create ground truth labels $I_g$.

We train the regressor by fine-tuning a pre-trained diffusion model using a LoRA on its self-attention layers.
This has two benefits over the more straight forward approach of training a regressor from scratch fed by features from all layers (e.g., \cite{hyper_features, readout_guidance}): first, the priors learned by the foundation model are better utilized, as we surgically tap into specific layers (see \Cref{sec:analysis}). We demonstrate significantly better results with 10 times less parameters, when trained on the same data (\Cref{fig:zoo1,fig:zoo2}).
Second, this allows simple control over the trade-off between accuracy and parameter count (by adjusting the LoRA rank).

In a previous study, LoRAs were used to transform the backbone of a diffusion model into an image-to-image translator \cite{ilora} for a single fixed timestep, which means they are unusable for classifier guidance.
In contrast, we train the regressor over \textit{all} timesteps, similar to how the diffusion model is trained.
We task the regressor to predict the clean direct-irradiance signal using the $L_2$ loss: $||\hat{I}_g - I_g||^2_2$. We note that the loss is computed in latent space, after encoding both $\hat{I}_g, I_g$, and notation is omitted for brevity.

\subsection{Light Control}
\label{sec:lantern_control}
\begin{figure}[h]
    \centering
    {
    \setlength{\tabcolsep}{0pt} 
    \renewcommand{\arraystretch}{1} 
    \begin{tabular}{ccc}
    Illumination & Condition $I_c$ & Result  \\[0.5em]
    \includegraphics[width=0.3\linewidth]{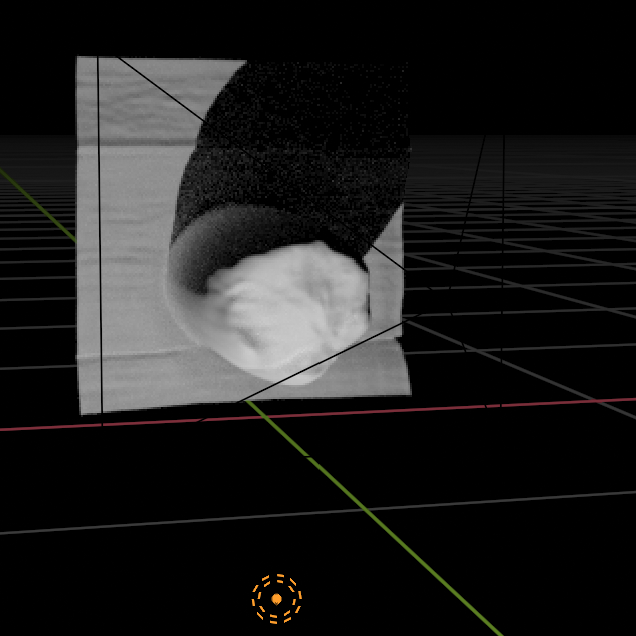} & 
    \includegraphics[width=0.3\linewidth]{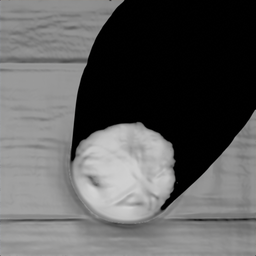} & 
    \includegraphics[width=0.3\linewidth]{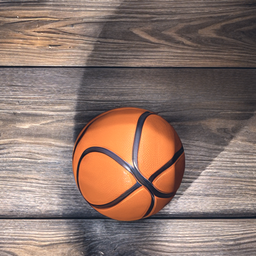} \\ 
    \includegraphics[width=0.3\linewidth]{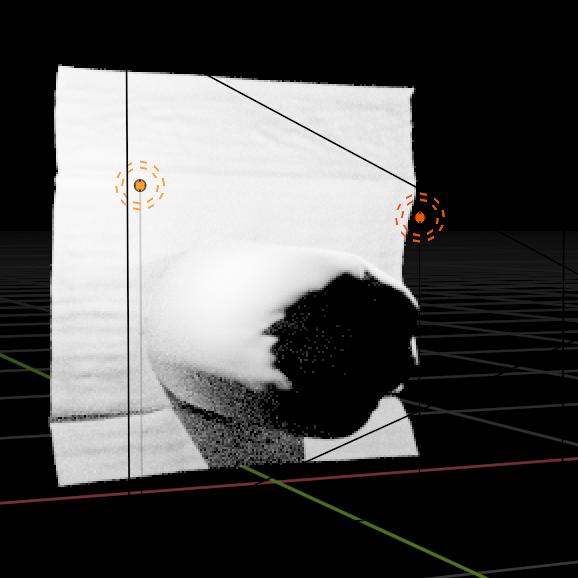} & 
    \includegraphics[width=0.3\linewidth]{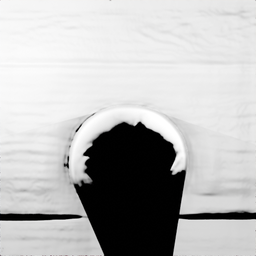} & 
    \includegraphics[width=0.3\linewidth]{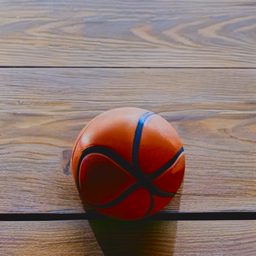} \\
    \includegraphics[width=0.3\linewidth]{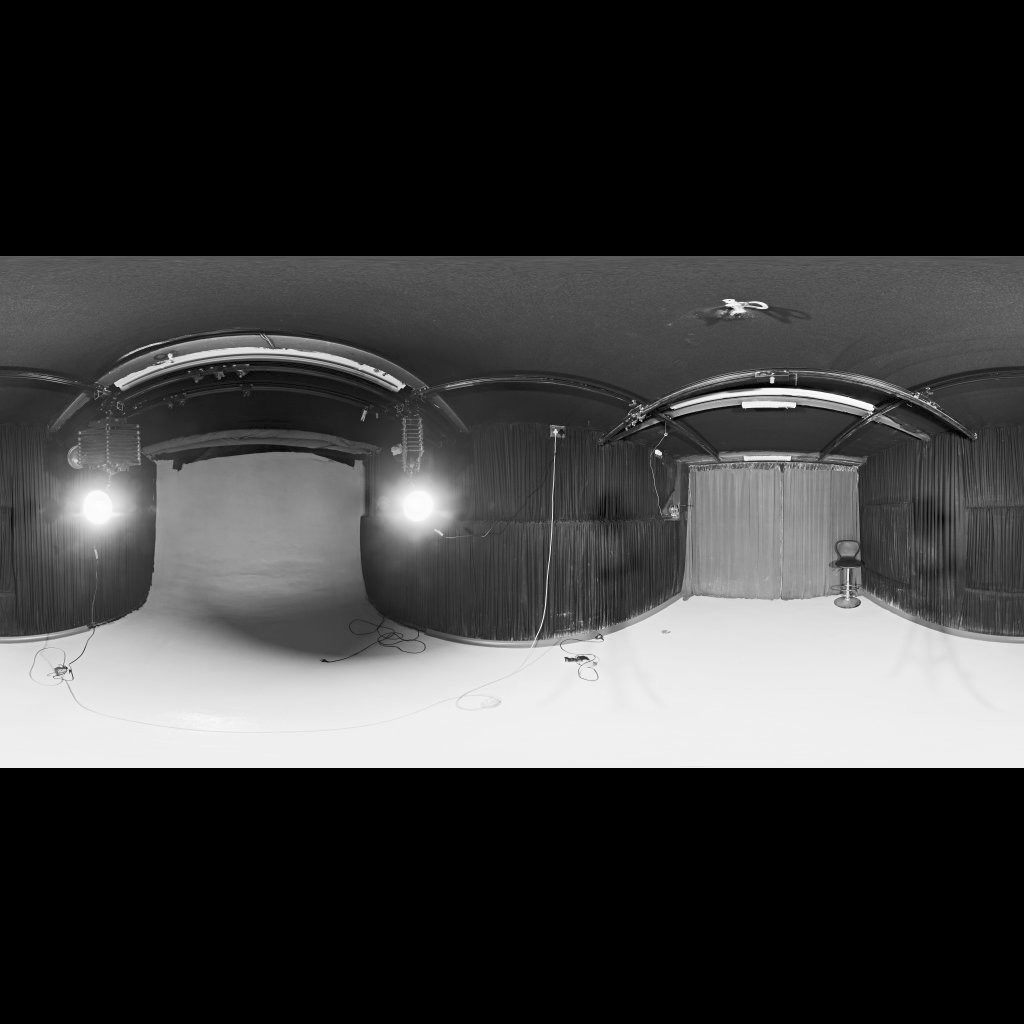} & 
    \includegraphics[width=0.3\linewidth]{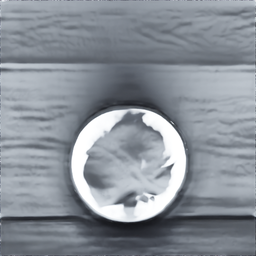} & 
    \includegraphics[width=0.3\linewidth]{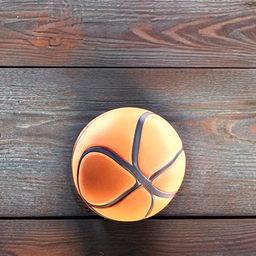} \\ 
    \end{tabular}
    }
    \caption{\textbf{Light Conditions.} \ourmethod allows setting arbitrarily gray scale light sources including one (top) or more (middle) point light sources and an environment map (bottom).}
    \label{fig:light_conditions}
\end{figure}

To allow the users to create a novel light condition, we use a scheme similar to the radiance hints proposed by \citet{dilightnet}: we first estimate a depth map for the source input $I_s$ using an off-the-shelf tool \cite{zoe_depth}, and create a 3D height-map of the scene without texture. We assign a GGX material to the geometry, allowing the user to control the roughness.
Then, any light source(s) can be inserted to create a desired effect, and a single direct-irradiance map (1\textsuperscript{st} bounce) is rendered using an orthographic camera which serves as $I_c$. The whole process, illustrated in \Cref{fig:overview} and the supplemental material, is quick and user-friendly, and can be fully automated to create random light conditions if needed. This allows \ourmethod to support arbitrary gray scale light conditions as shown in \Cref{fig:light_conditions}.

\subsection{Relighting}
\label{sec:relighting}
Recall that our goal is to edit a prompt generated source image $I_s$ into a relit image $I_r$, based on a control image, $I_c$. The relighting process described in this section is illustrated in \Cref{fig:overview}, bottom.

We guide the sampling process of $I_r$ with a guidance energy term (\Cref{eq:guidance}): $$E = \frac{\nabla_x \mathcal{L}}{\left\| \nabla_x \mathcal{L} \right\|_2^2}, \quad \text{where } \mathcal{L} = \left\| \hat{I}_g - I_c \right\|_2^2$$ which ensures that for each timestep the current direct-irradiance estimate $\hat{I}_g$ are similar to the control image $I_c$. As a reminder, the energy term is computed in latent space, and notations are omitted for brevity.

Because of an imperfect regressor and a domain gap between synthetic training data and realistic generated images, in addition to noisy gradient updates and disentanglement between light and other phenomena, applying guidance na\"ively results in significant loss of structure, identity and style preservation of the source $I_s$ as we show in \Cref{sec:ablations}.
To better preserve image content, we take two complementary steps: \textit{First}, we apply a pretrained ControlNet \cite{controlnet} conditioned on an edge-map of the condition $I_c$ added to an edge map of the albedo estimate of the original image \cite{cd-iid}. 
This helps conserve the structure of $I_s$, but avoids reconstructing hard shadows or highlights appearing in it, while also transferring hard shadow edges from the light condition.
However, solely using a ControlNet introduces cartoonification and color saturation artifacts to the results (see supplemental material for an example).
We found this to persist over different ControlNets, even when they were trained on realistic datasets only. 
We hence further encourage identity preservation through injection of self-attention queries, that were shown to encode structure \cite{cia}. 
To this end, during the sampling process of $I_r$, we inject the queries of all self-attention layers from the sampling process of $I_s$, reducing cartoonification and ensuring preservation of identity. The final $I_r$ is created by normalizing each channel with the mean and the standard deviation of the original image $I_s$ \cite{color_correction}, assuring any global color shifts are corrected (see \Cref{fig:diffusion_process}).

\section{Experiments}

\label{sec:results}
We start with implementation details (\Cref{sec:implementation}), followed by describing an evaluation dataset used for all experiments (\Cref{sec:eval_dataset}). We then show both qualitative (\Cref{sec:qualitative_results}) and quantitative (\Cref{sec:quantitative_results}) results, followed by a subjective user study (\Cref{sec:user_study}). Finally we show an ablation and design choice study (\Cref{sec:ablations}).

\begin{table*}[th]
\caption{\textbf{Quantitative Evaluation.} We compare against RGB$\leftrightarrow$x \cite{rgbx} and IC-Light \cite{iclight} over our evaluation dataset for aesthetics, adherence to control, identity preservation and efficiency.}
\label{tab:quantitative}
\centering
\begin{tabular}{||c|c|c|c|c|c|c|c|c|c|c|c|c||}
\hline
\multirow{2}{*}{} & Aesthetic &  \multicolumn{4}{|c|}{Control} & \multicolumn{4}{|c|}{Identity} & \multicolumn{3}{|c|}{Efficiency}\\
\cline{2-13}
 & HPSv2 $\uparrow$ & $L_2 \downarrow$ & LPIPS $\downarrow$ & CLIP $\downarrow$ & DINO $\downarrow$ & $L_2 \downarrow$ & LPIPS $\downarrow$ & CLIP $\downarrow$ & DINO $\downarrow$ & Params $\downarrow$ & Data $\downarrow$ & Compute $\downarrow$\\ 
\hline
RGB$\leftrightarrow$x & 0.2334 & 0.17 & \underline{0.74} & \underline{0.53} & 0.72 & \underline{0.100} & 0.45 & 0.160 & 0.21 & 1.7e9 & \underline{2e5} & 1600h\\ 
\hline
IC-Light & \underline{0.2462} & \textbf{0.10} & \textbf{0.63} & \textbf{0.50} & \textbf{0.68} & \underline{0.100} & \underline{0.38} & \underline{0.140} & \underline{0.19} & \underline{8.6e8} & 1e7 & \underline{800h} \\
\hline
Ours & \textbf{0.2493} & \underline{0.13} & \underline{0.74} & 0.56 & \underline{0.71} & \textbf{0.056} & \textbf{0.30} & \textbf{0.088} & \textbf{0.14} & \textbf{7.9e5} & \textbf{4e3} & \textbf{1h} \\
\hline
\end{tabular}
\end{table*}

\subsection{Implementation}
\label{sec:implementation}
We use a pretrained SD1.5 as the foundational model and a Quadro RTX 8000 NVidia GPU with 48 GB of RAM for all our experiments. We train on 4000 synthetic image pairs of resolution 512x512, and optimize the $q,k,v$ and linear output projection matrices of all self-attention layers by decomposing them into rank 8 matrices. For scheduling the guidance (\Cref{eq:update_rule}), we use: $$ \gamma_t = 
\begin{cases}
2.2, & t\in[0.05, 0.5] \\
0,   & \text{otherwise}
\end{cases}$$ where $t$ is the normalized timestep. Ideally we would start at $t=0$ as suggested in \Cref{sec:analysis}, but starting too early tends to result in loss of structure. We also observe that going beyond the half point tends to degenerate $\hat{I}_r$ into a blending scheme between $I_s$ and $I_c$ and exhibits severe loss of identity (see supplement). Sampling with \ourmethod takes roughly twice as long as sampling with SD1.5 out-of-the-box due to the backward pass used to compute gradients for guidance.

\subsection{Evaluation Dataset}
\label{sec:eval_dataset}
Our main contribution is controlling light generally, across various image domains. 
However, existing datasets for relighting do not span several domains, and ground truth lighting annotations are very difficult to obtain for in-the-wild imagery. 
There is also no consensus on how to faithfully evaluate unlabeled plausible relighting. 

For these reasons, we construct our own evaluation dataset, that spans several domains, and is suitable for evaluating light control. We draw prompts from DiffusionDB \cite{diffusiondb}, which contains 2 million prompts-image pairs and hyper parameters for their generations. After basic filtering for unsafe content and duplication, we manually group them into nine categories: Portrait, Portrait CG, Anime, Animal, Fantasy Landscape, Indoors, Outdoors, Painting, Sketch/Cartoon.
In each category, we sort the pairs according to an aesthetic score \cite{aesthetic_score}, and select the top 20 images. This results in 180 prompt-image pairs and their generation hyper parameters (seed, diffusion timesteps, and CFG value).

We then create a novel light condition for each image using a randomly placed point light source as described in \Cref{sec:lantern_control}.
This dataset is designed for generative comparisons, as the generation parameters for each prompt allow for consistent generation. Importantly, this dataset spans multiple image domains with relatively high-quality and aesthetic images, allowing for human subjects to perform evaluations without being distracted by the content.

\subsection{Qualitative Results}
\label{sec:qualitative_results}

\begin{figure*}
    \centering
    {
    \setlength{\tabcolsep}{0pt} 
    \renewcommand{\arraystretch}{0} 
    \begin{tabular}{*{2}{>{\centering\arraybackslash}m{0.13\textwidth}} @{\hskip 5pt} *{5}{>{\centering\arraybackslash}m{0.13\textwidth}}}
    Source $I_s$& Condition $I_c$ & Readouts & ControlNet & RGB$\leftrightarrow$x & IC-Light & Ours  
    \end{tabular}
        \begin{tabular}{*{2}{>{\centering\arraybackslash}m{0.13\textwidth}} @{\hskip 5pt} *{5}{>{\centering\arraybackslash}m{0.13\textwidth}}}
        \\[0.5em]
        \includegraphics[width=\linewidth]{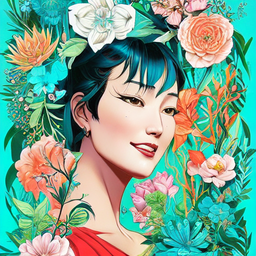} & 
        \includegraphics[width=\linewidth]{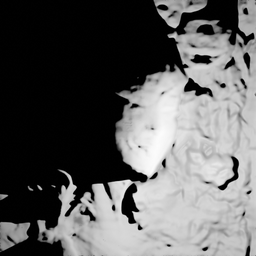} & 
        \includegraphics[width=\linewidth]{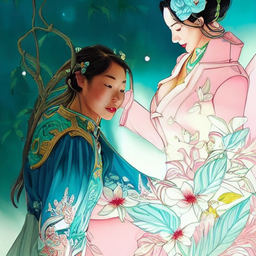} & 
        \includegraphics[width=\linewidth]{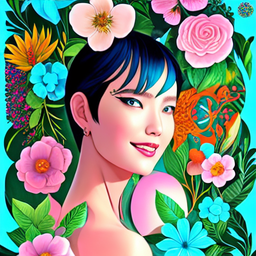} &
        \includegraphics[width=\linewidth]{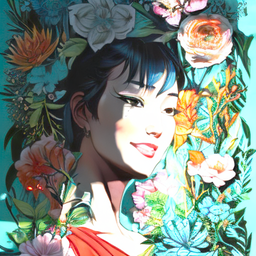} & 
        \includegraphics[width=\linewidth]{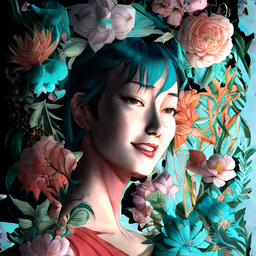} & 
        \includegraphics[width=\linewidth]{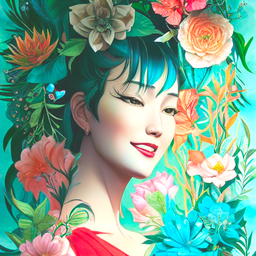} \\ 
        \includegraphics[width=\linewidth]{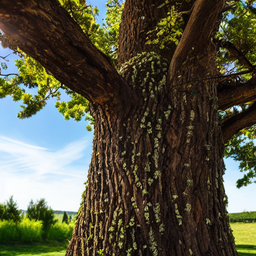} & 
        \includegraphics[width=\linewidth]{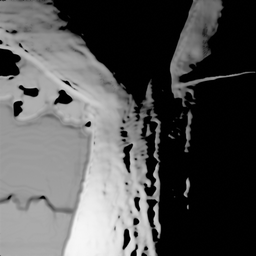} & 
        \includegraphics[width=\linewidth]{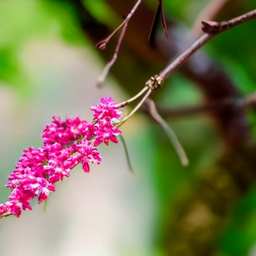} & 
        \includegraphics[width=\linewidth]{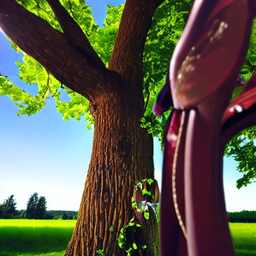} & 
        \includegraphics[width=\linewidth]{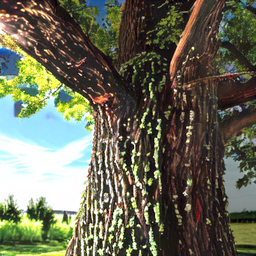} & 
        \includegraphics[width=\linewidth]{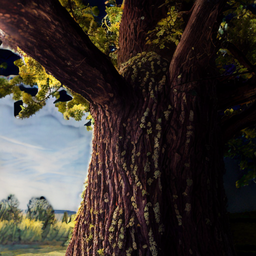} & 
        \includegraphics[width=\linewidth]{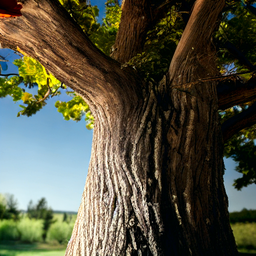} \\
        \includegraphics[width=\linewidth]{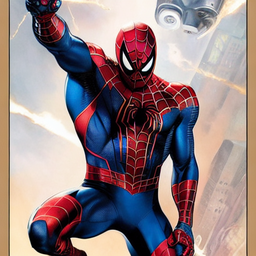} & 
        \includegraphics[width=\linewidth]{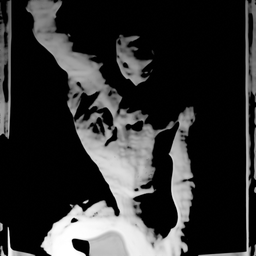} & 
        \includegraphics[width=\linewidth]{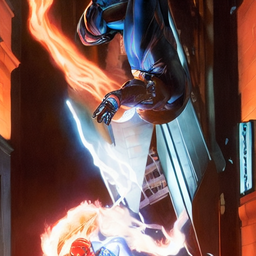} & 
        \includegraphics[width=\linewidth]{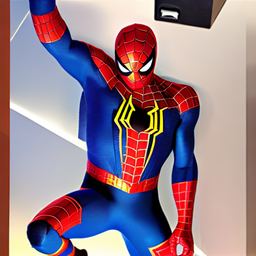} &
        \includegraphics[width=\linewidth]{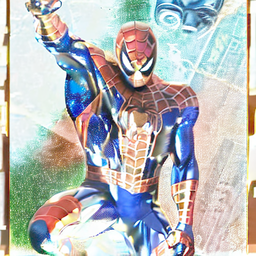} & 
        \includegraphics[width=\linewidth]{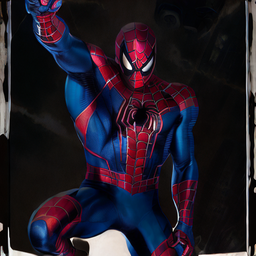} & 
        \includegraphics[width=\linewidth]{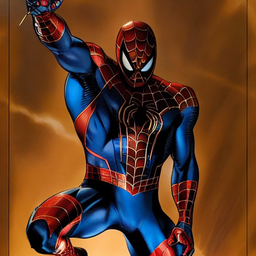} \\
        \includegraphics[width=\linewidth]{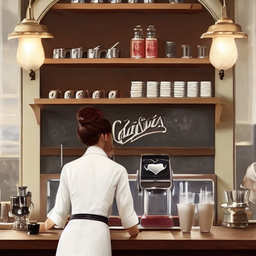} & 
        \includegraphics[width=\linewidth]{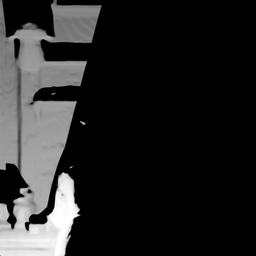} & 
        \includegraphics[width=\linewidth]{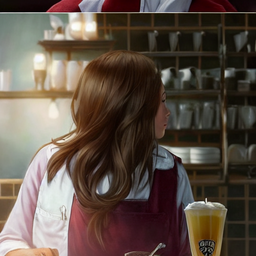} & 
        \includegraphics[width=\linewidth]{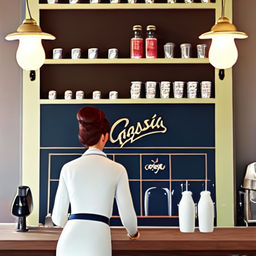} &
        \includegraphics[width=\linewidth]{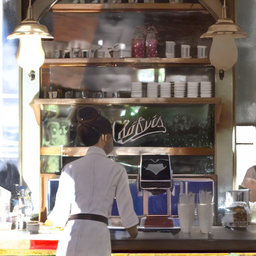} & 
        \includegraphics[width=\linewidth]{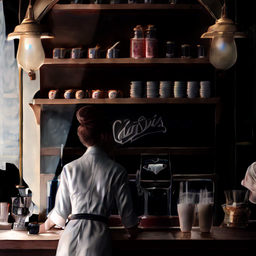} & 
        \includegraphics[width=\linewidth]{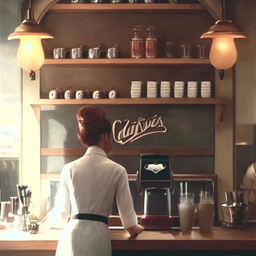} \\
        \includegraphics[width=\linewidth]{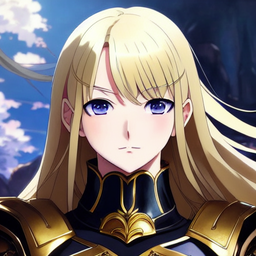} & 
        \includegraphics[width=\linewidth]{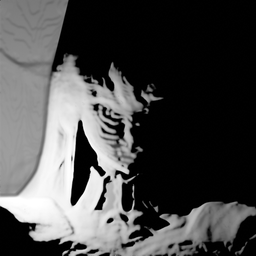} & 
        \includegraphics[width=\linewidth]{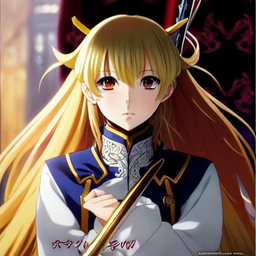} & 
        \includegraphics[width=\linewidth]{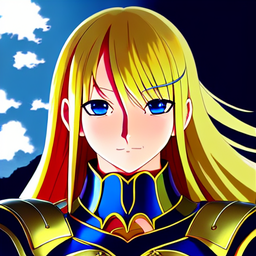} &
        \includegraphics[width=\linewidth]{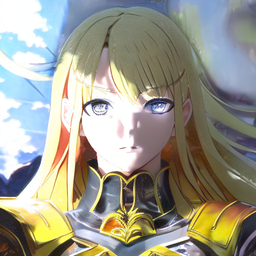} & 
        \includegraphics[width=\linewidth]{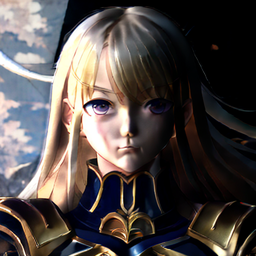} & 
        \includegraphics[width=\linewidth]{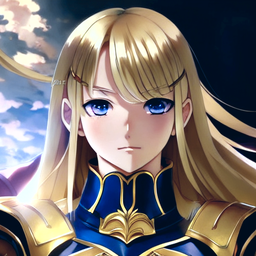} \\
        \includegraphics[width=\linewidth]{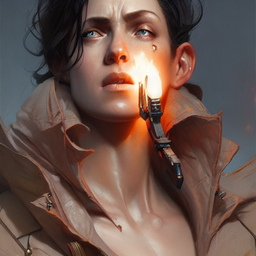} & 
        \includegraphics[width=\linewidth]{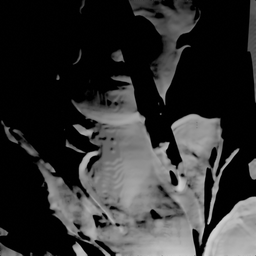} & 
        \includegraphics[width=\linewidth]{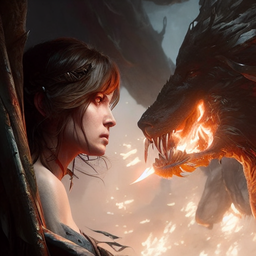} & 
        \includegraphics[width=\linewidth]{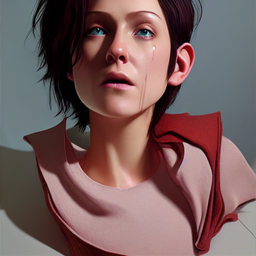} &
        \includegraphics[width=\linewidth]{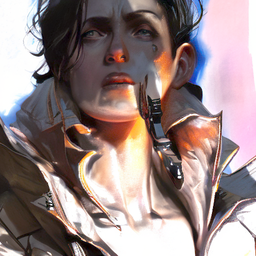} & 
        \includegraphics[width=\linewidth]{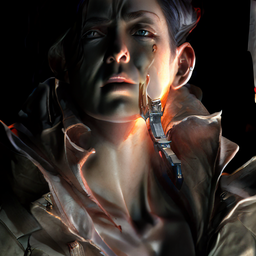} & 
        \includegraphics[width=\linewidth]{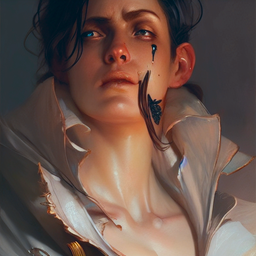} \\ 
        \includegraphics[width=\linewidth]{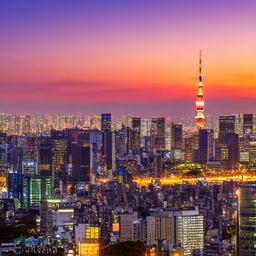} & 
        \includegraphics[width=\linewidth]{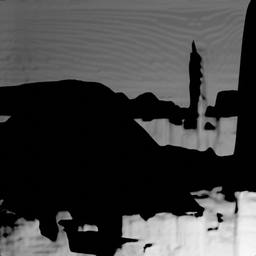} & 
        \includegraphics[width=\linewidth]{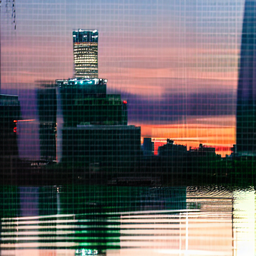} & 
        \includegraphics[width=\linewidth]{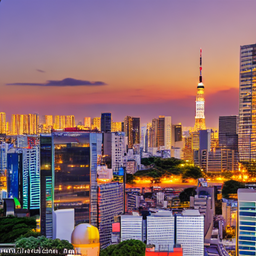} &
        \includegraphics[width=\linewidth]{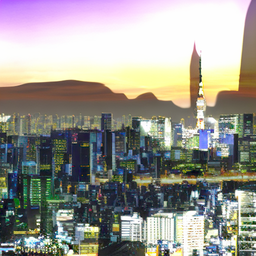} & 
        \includegraphics[width=\linewidth]{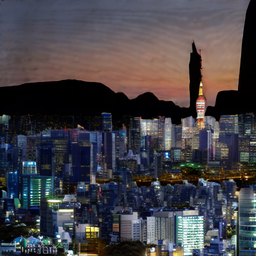} & 
        \includegraphics[width=\linewidth]{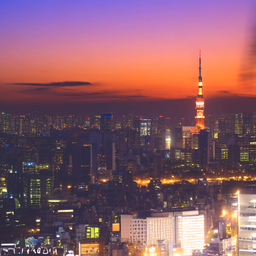} \\
        \includegraphics[width=\linewidth]{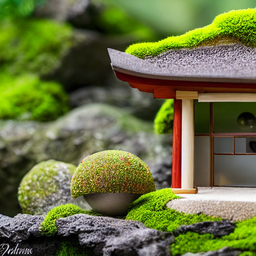} & 
        \includegraphics[width=\linewidth]{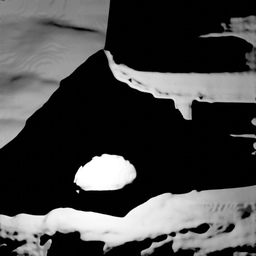} & 
        \includegraphics[width=\linewidth]{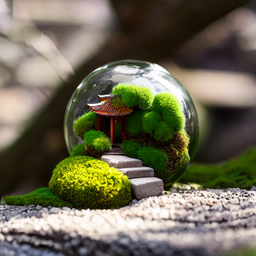} & 
        \includegraphics[width=\linewidth]{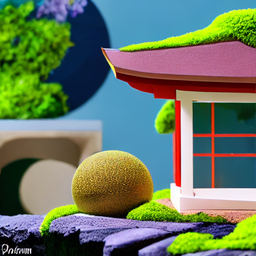} &
        \includegraphics[width=\linewidth]{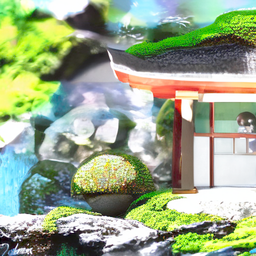} & 
        \includegraphics[width=\linewidth]{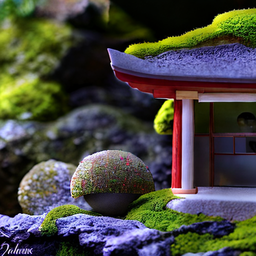} & 
        \includegraphics[width=\linewidth]{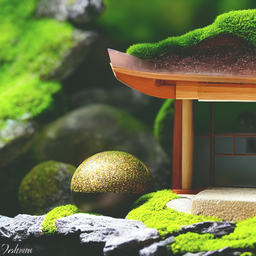} \\
        \includegraphics[width=\linewidth]{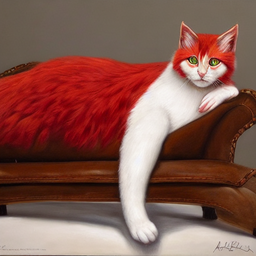} & 
        \includegraphics[width=\linewidth]{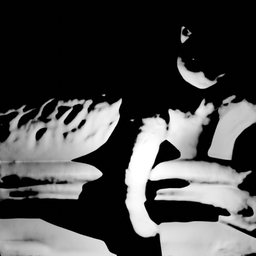} & 
        \includegraphics[width=\linewidth]{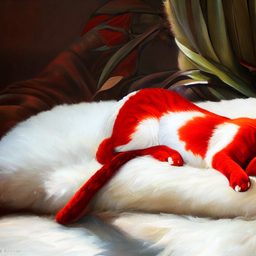} & 
        \includegraphics[width=\linewidth]{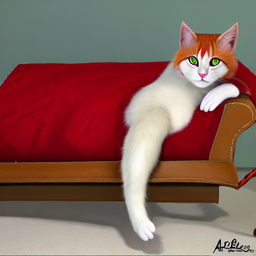} &
        \includegraphics[width=\linewidth]{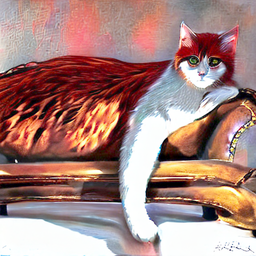} & 
        \includegraphics[width=\linewidth]{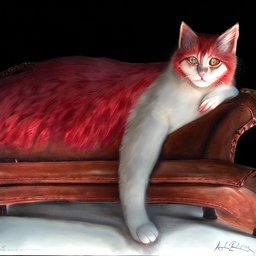} & 
        \includegraphics[width=\linewidth]{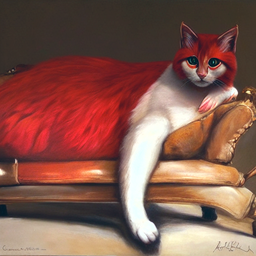} \\ 
    \end{tabular}
    }
    \caption{\textbf{Qualitative Results.} Relighting results on the evaluation dataset. Zoomed-in viewing recommended.}
    \label{fig:zoo1}
\end{figure*}

\begin{figure*}
    \centering
    {
    \setlength{\tabcolsep}{0pt} 
    \renewcommand{\arraystretch}{0} 
    \begin{tabular}{*{2}{>{\centering\arraybackslash}m{0.12\textwidth}} @{\hskip 5pt} *{5}{>{\centering\arraybackslash}m{0.12\textwidth}}}
    Source $I_s$& Condition $I_c$ & Readouts & ControlNet & RGB$\leftrightarrow$x & IC-Light & Ours  
    \end{tabular}
        \begin{tabular}{*{2}{>{\centering\arraybackslash}m{0.12\textwidth}} @{\hskip 5pt} *{5}{>{\centering\arraybackslash}m{0.12\textwidth}}}
        \\[0.5em]
        \includegraphics[width=\linewidth]{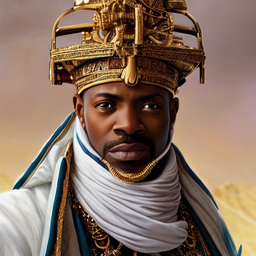} & 
        \includegraphics[width=\linewidth]{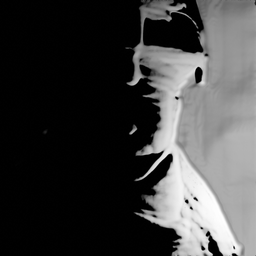} & 
        \includegraphics[width=\linewidth]{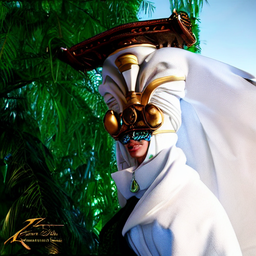} & 
        \includegraphics[width=\linewidth]{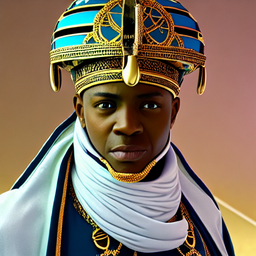} &
        \includegraphics[width=\linewidth]{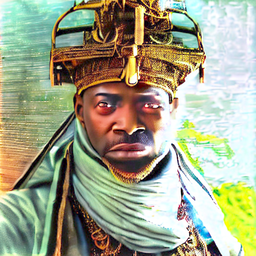} & 
        \includegraphics[width=\linewidth]{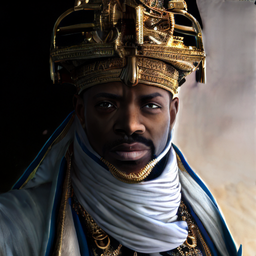} & 
        \includegraphics[width=\linewidth]{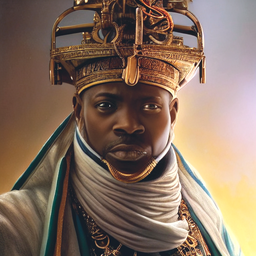} \\ 
        \includegraphics[width=\linewidth]{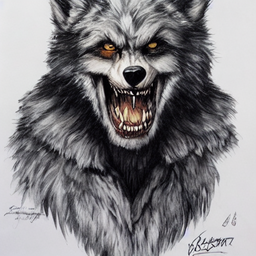} & 
        \includegraphics[width=\linewidth]{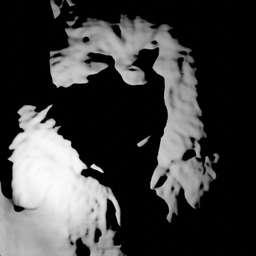} & 
        \includegraphics[width=\linewidth]{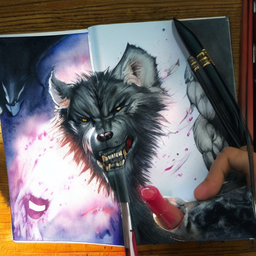} & 
        \includegraphics[width=\linewidth]{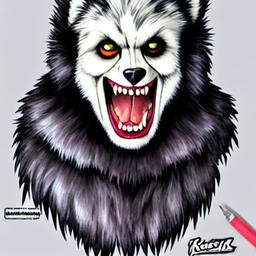} & 
        \includegraphics[width=\linewidth]{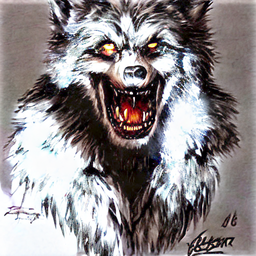} & 
        \includegraphics[width=\linewidth]{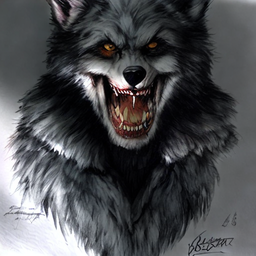} & 
        \includegraphics[width=\linewidth]{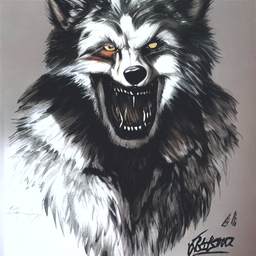} \\
        \includegraphics[width=\linewidth]{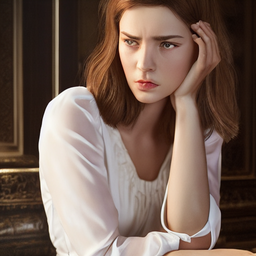} & 
        \includegraphics[width=\linewidth]{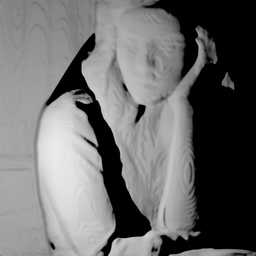} & 
        \includegraphics[width=\linewidth]{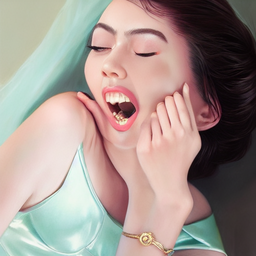} & 
        \includegraphics[width=\linewidth]{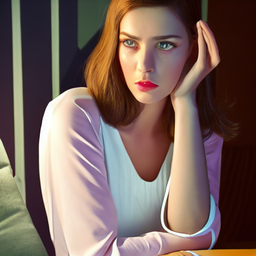} &
        \includegraphics[width=\linewidth]{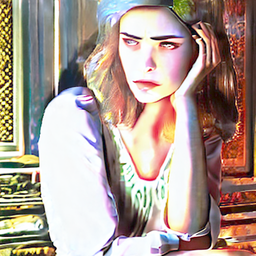} & 
        \includegraphics[width=\linewidth]{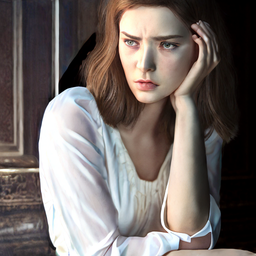} & 
        \includegraphics[width=\linewidth]{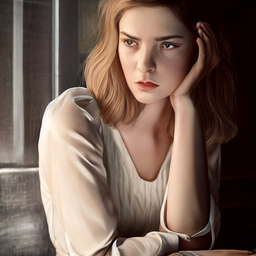} \\
        \includegraphics[width=\linewidth]{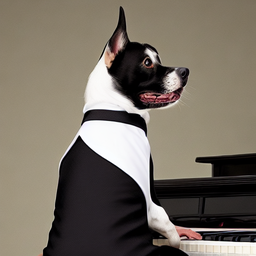} & 
        \includegraphics[width=\linewidth]{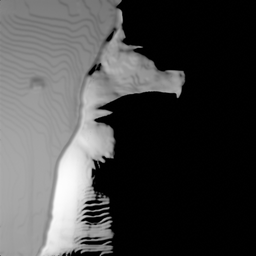} & 
        \includegraphics[width=\linewidth]{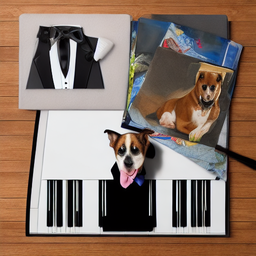} & 
        \includegraphics[width=\linewidth]{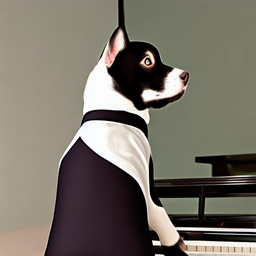} &
        \includegraphics[width=\linewidth]{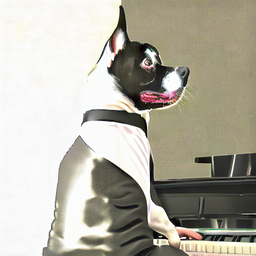} & 
        \includegraphics[width=\linewidth]{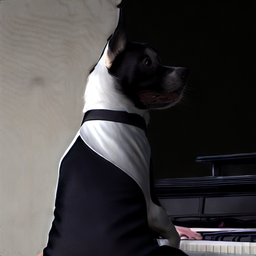} & 
        \includegraphics[width=\linewidth]{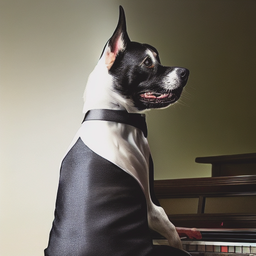} \\
        \includegraphics[width=\linewidth]{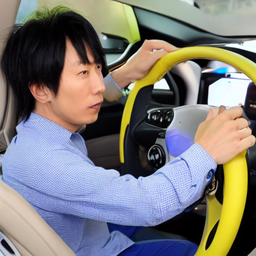} & 
        \includegraphics[width=\linewidth]{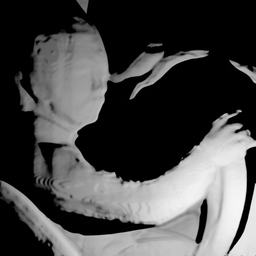} & 
        \includegraphics[width=\linewidth]{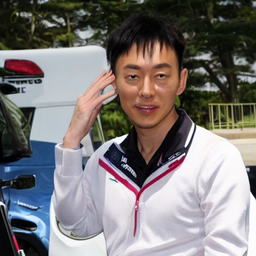} & 
        \includegraphics[width=\linewidth]{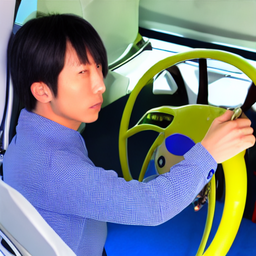} &
        \includegraphics[width=\linewidth]{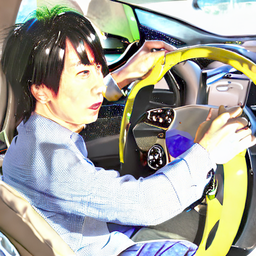} & 
        \includegraphics[width=\linewidth]{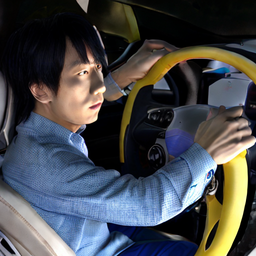} & 
        \includegraphics[width=\linewidth]{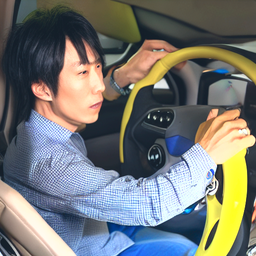} \\
        \includegraphics[width=\linewidth]{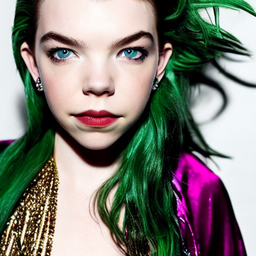} & 
        \includegraphics[width=\linewidth]{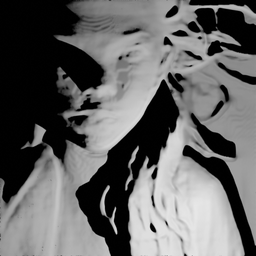} & 
        \includegraphics[width=\linewidth]{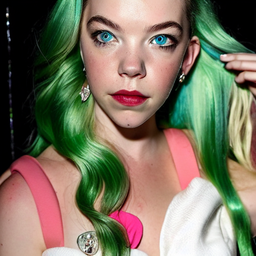} & 
        \includegraphics[width=\linewidth]{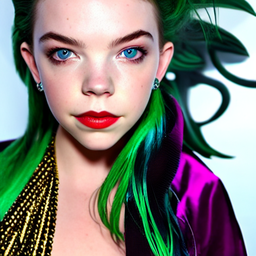} &
        \includegraphics[width=\linewidth]{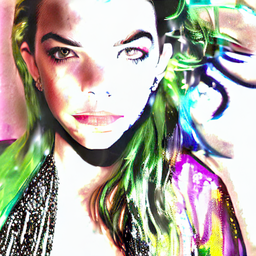} & 
        \includegraphics[width=\linewidth]{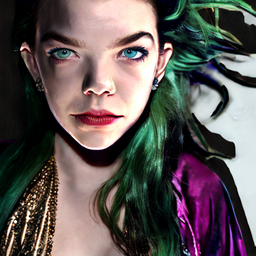} & 
        \includegraphics[width=\linewidth]{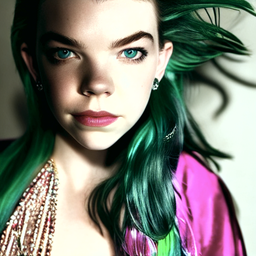} \\
    \end{tabular}
    }
    \caption{\textbf{Qualitative Results.} Relighting results on the evaluation dataset. Zoomed-in viewing recommended.}
    \label{fig:zoo2}
\end{figure*}

\begin{figure*}
    \centering
    {
    \setlength{\tabcolsep}{0pt} 
    \renewcommand{\arraystretch}{0} 
    \begin{tabular}{*{2}{>{\centering\arraybackslash}m{0.10\textwidth}} @{\hskip 1pt} *{2}{>{\centering\arraybackslash}m{0.10\textwidth}} @{\hskip 1pt}*{2}{>{\centering\arraybackslash}m{0.10\textwidth}} @{\hskip 1pt}*{2}{>{\centering\arraybackslash}m{0.10\textwidth}} @{\hskip 1pt}}
    DiLightNet & Ours & DiLightNet & Ours & DiLightNet & Ours & DiLightNet & Ours
    \end{tabular}
        \begin{tabular}{*{2}{>{\centering\arraybackslash}m{0.10\textwidth}} @{\hskip 1pt} *{2}{>{\centering\arraybackslash}m{0.10\textwidth}} @{\hskip 1pt}*{2}{>{\centering\arraybackslash}m{0.10\textwidth}} @{\hskip 1pt}*{2}{>{\centering\arraybackslash}m{0.10\textwidth}} @{\hskip 1pt}}
        \\[0.5em]
        \includegraphics[width=\linewidth]{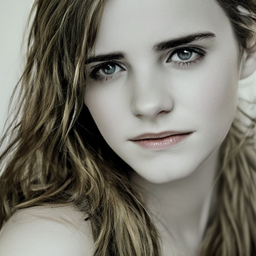} & 
        \includegraphics[width=\linewidth]{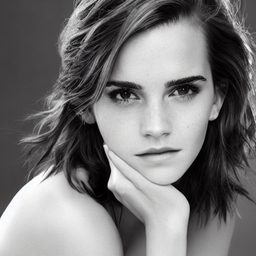} & 
        \includegraphics[width=\linewidth]{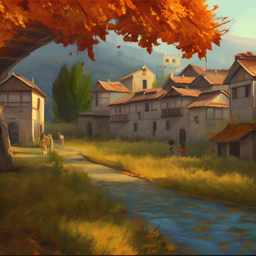} &
        \includegraphics[width=\linewidth]{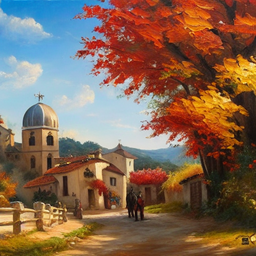} & 
        \includegraphics[width=\linewidth]{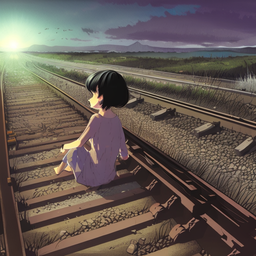} & 
        \includegraphics[width=\linewidth]{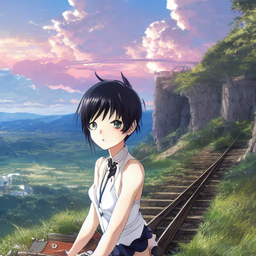} & 
        \includegraphics[width=\linewidth]{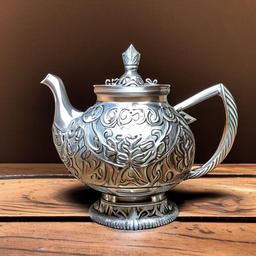} &
        \includegraphics[width=\linewidth]{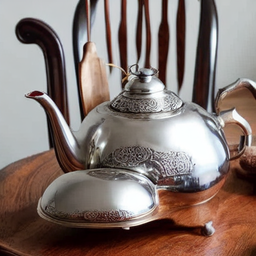} \\ 
        \includegraphics[width=\linewidth]{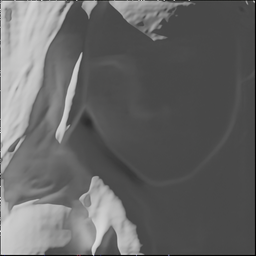} & 
        \includegraphics[width=\linewidth]{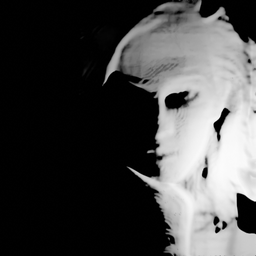} & 
        \includegraphics[width=\linewidth]{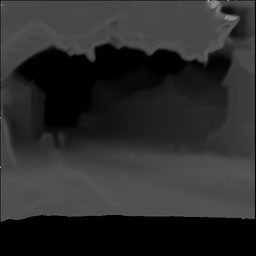} &
        \includegraphics[width=\linewidth]{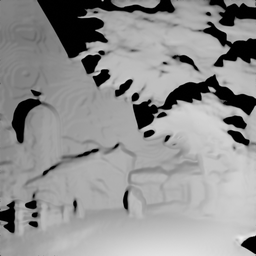} & 
        \includegraphics[width=\linewidth]{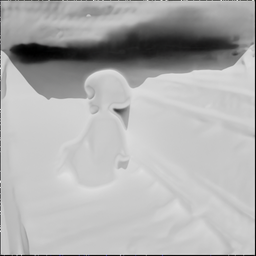} & 
        \includegraphics[width=\linewidth]{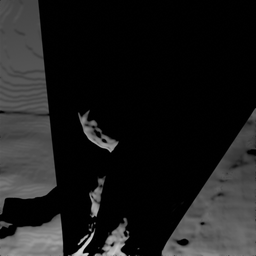} & 
        \includegraphics[width=\linewidth]{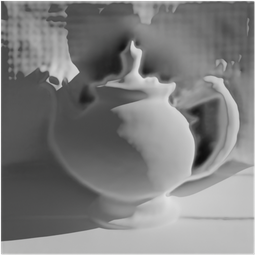} &
        \includegraphics[width=\linewidth]{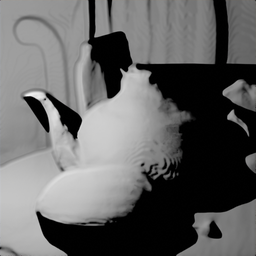} \\
        \includegraphics[width=\linewidth]{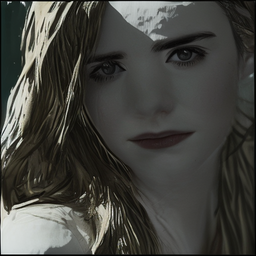} & 
        \includegraphics[width=\linewidth]{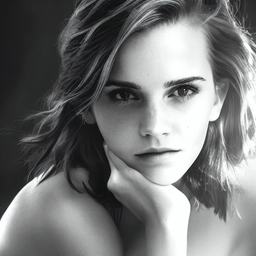} & 
        \includegraphics[width=\linewidth]{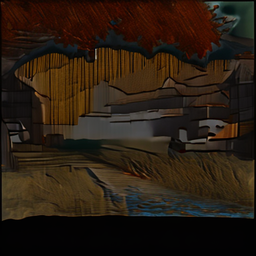} &
        \includegraphics[width=\linewidth]{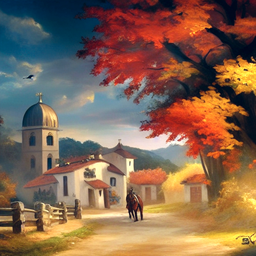} & 
        \includegraphics[width=\linewidth]{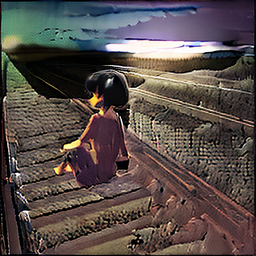} & 
        \includegraphics[width=\linewidth]{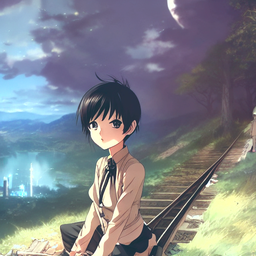} & 
        \includegraphics[width=\linewidth]{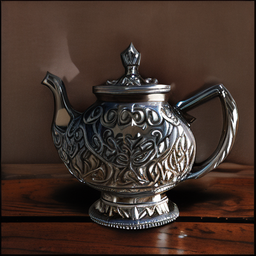} &
        \includegraphics[width=\linewidth]{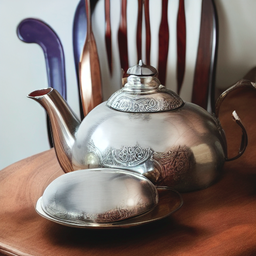} \\
    \end{tabular}
    }
    \caption{\textbf{DiLightNet Comparisons.} Top: source $I_s$, middle: condition $I_c$, bottom: relit result. Zoomed-in viewing recommended.}
    \label{fig:zoo3}
\end{figure*}

We show qualitative results comparing our method to other competitive approaches in \Cref{fig:zoo1,fig:zoo2,fig:zoo3} over prompts from our evaluation dataset (\Cref{sec:eval_dataset}). We carefully selected and fixed hyper parameters for each method. The selected approaches are: IC-Light \cite{iclight} and RGB$\leftrightarrow$x \cite{rgbx}; both using the original implementation and weights, and Readout Guidance \cite{readout_guidance} and ControlNet \cite{controlnet}; both trained from scratch on our synthetic image dataset. Note ControlNet is usually trained on significantly larger and diverse datasets, and thus in our case fails to preserve identity or relight. For comparisons with DiLightNet \cite{dilightnet}, the source image and guidance signals are different due to difference in foundational models and required control signal. For full implementation details see supplemental material.
As can be seen, our approach is highly competitive with other methods, and is able to relight photorealistic imagery, perform cel-shading, deal with sketches, CG art, surrealistic paintings and more. In contrast, other methods specialize in one specific domain such as portraits or object-centric scenes, and fail to deliver reasonable results in general.

\begin{table*}[th]
\caption{\textbf{Ablation and Design Choice Study.} Best results are in \textbf{bold}, second-best are \underline{underlined}, and third-best are \textit{italicized}. See \Cref{sec:ablations} for more details.}
\label{tab:ablations}
\centering
\begin{tabular}{||c|c|c|c|c|c|c|c|c|c||}
\hline
\multirow{2}{*}{} & Aesthetic &  \multicolumn{4}{|c|}{Control} & \multicolumn{4}{|c|}{Identity} \\
\cline{2-10}
 & HPSv2 $\uparrow$ & $L_2 \downarrow$ & LPIPS $\downarrow$ & CLIP $\downarrow$ & DINO $\downarrow$ & $L_2 \downarrow$ & LPIPS $\downarrow$ & CLIP $\downarrow$ & DINO $\downarrow$\\ 
\hline
$I_s$ & 0.2607 & 0.23 & 0.75 & 0.60 & 0.73 & - & - & - & - \\
\hline
Blending & 0.2153 & 0.057 & 0.61 & 0.47 & 0.59 & 0.057 & 0.22 & 0.130 & 0.180 \\
\hline
\hline
With Cross Attention & 0.2483 & \textit{0.12} & \underline{0.74} & 0.56 & \textit{0.70} & 0.057 & \textit{0.30} & 0.090 & \textit{0.140} \\
\hline
No Scheduling & \textbf{0.2546} & 0.21 & \textit{0.75} & 0.59 & 0.72 & \textbf{0.004} & \textbf{0.11} & \textbf{0.021} & \textbf{0.028} \\
\hline
$\gamma_t=4.0$ & 0.2363 & \textbf{0.10} & \textbf{0.73} & \textbf{0.52} & \textbf{0.67} & 0.076 & 0.39 & 0.140 & 0.220 \\
\hline
No Query Injection & 0.2441 & \underline{0.11} & \underline{0.74} & \textit{0.55} & \underline{0.68} & 0.086 & 0.46 & 0.150 & 0.240 \\
\hline
No ControlNet & 0.2444 & \textit{0.12} & \textit{0.75} & \textit{0.55} & 0.71 & 0.066 & 0.36 & 0.120 & 0.180 \\
\hline
LoRA Rank=4 & \underline{0.2517} & 0.14 & \underline{0.74} & 0.57& 0.71 & \underline{0.042} & \underline{0.25} & \underline{0.067} & \underline{0.100} \\
\hline
LoRA Rank=16 & 0.2413 & \textit{0.12} & \textit{0.75} & \underline{0.54} & 0.69 & 0.063 & 0.34 & 0.110 & 0.160 \\
\hline
Noisy Labels & 0.2350 & \underline{0.11} & \textbf{0.73} & \textbf{0.52} & \underline{0.68} & 0.064 & 0.38 & 0.130 & 0.204\\
\hline
Ours Full & \textit{0.2493} & 0.13 & \underline{0.74} & 0.56 & 0.71 & \textit{0.056} &\textit{0.30} & \textit{0.088} & \textit{0.140} \\
\hline
\end{tabular}
\end{table*}

\begin{table}[h]
    \caption{\textbf{User Study Results.} Users predominantly preferred our method in terms of identity and style preservation, and better relighting overall.}
    \label{tab:user_study}
    \centering
    \begin{tabular}{||c|c|c||c|c||}
        \hline
        \textbf{Metric $\uparrow$} & \textbf{Ours} & \textbf{IC-Light} & \textbf{Ours} & \textbf{DiLightNet}\\
        \hline
        Identity \& Style & \textbf{73.7\%} & 26.3\% & \textbf{77.3}\% & 22.7\% \\
        \hline
        Relighting & \textbf{67.0\%} & 33.0\% & \textbf{70.4}\% & 29.6\%\\
        \hline
    \end{tabular}

\end{table}

\subsection{Quantitative Evaluation}
\label{sec:quantitative_results}
Using the evaluation dataset, we performed a quantitative comparison with two state-of-the-art relighting techniques (See \Cref{tab:quantitative}). The selected methods are IC-Light \cite{iclight} and RGB$\leftrightarrow$x \cite{rgbx}, chosen as the most relevant and competitive techniques using similar control signals and exhibiting some generalizability to other image domains. All methods used the same prompt and seed derived from the dataset. CFG values and diffusion timesteps were chosen between dataset-provided values and manual tuning (the best of the two options). For IC-Light, we use the official background-conditioned model finetuned from SD1.5. For RGB$\leftrightarrow$x, we used the official code and decomposed each image using the RGB$\rightarrow$x pipeline, replaced its irradiance channel with a diffuse render pass (not our direct-irradiance map), and recomposed it back using the x$\rightarrow$RGB pipeline. We tuned and fixed the hyper parameters for all methods to the best of our abilities. See  for hyper-parameters and other details.

In \Cref{tab:quantitative}, the Human Preference Score (HPSv2 \cite{hpsv2}), is used to compare images generated with the same prompt, estimating (``Aesthetics'') according to human preference. 
Other metrics ($L_2$, LPIPS \cite{lpips}, CLIP score \cite{clip}, DINO score \cite{dino}) are computed between the result and the control signal (``Control''), or the original image (``Identity''). This allows evaluating how the control signal translated to the relit result, and the trade off it entails with identity preservation. Lastly, we show performance in terms of number of optimized parameters, training data points, and GPU hours (``Efficiency''). As the table shows, our method achieves a higher aesthetic score and identity preservation, while being on par in terms of control adherence, with three orders of magnitude less parameters and data. We attribute this success to effectively leveraging the existing knowledge in the pretrained backbone model.

\subsection{Subjective Evaluation}
\label{sec:user_study}
We conducted a subjective user study comparing our method and two other competitive relighting methods (separately) over the evaluation dataset with results shown in \Cref{tab:user_study}. We selected IC-Light \cite{iclight} and DiLightNet \cite{dilightnet} for comparisons, as the leading approaches for 
general purpose light control (IC-Light) as well as object-centric relighting (DilightNet), that use similar inputs to ours. Subjects showed high preference to our technique, validating further the generelizability of our approach. Users were asked to answer which method preserves the identity and style of the original image better, and which method produces an overall better relighting result (two questions per datapoint). Note that for comparisons with DiLightNet, the generated images and their condition signal differ per prompt due to the usage of a different foundational model (SD2.1) and required control.
See  for more details on the user study with exemplar questions, additional qualitative results, and hyper parameters used.

\subsection{Ablations and Design Choice Study}
\label{sec:ablations}
\begin{figure}[h]
    \centering
    \includegraphics[width=0.5\textwidth]{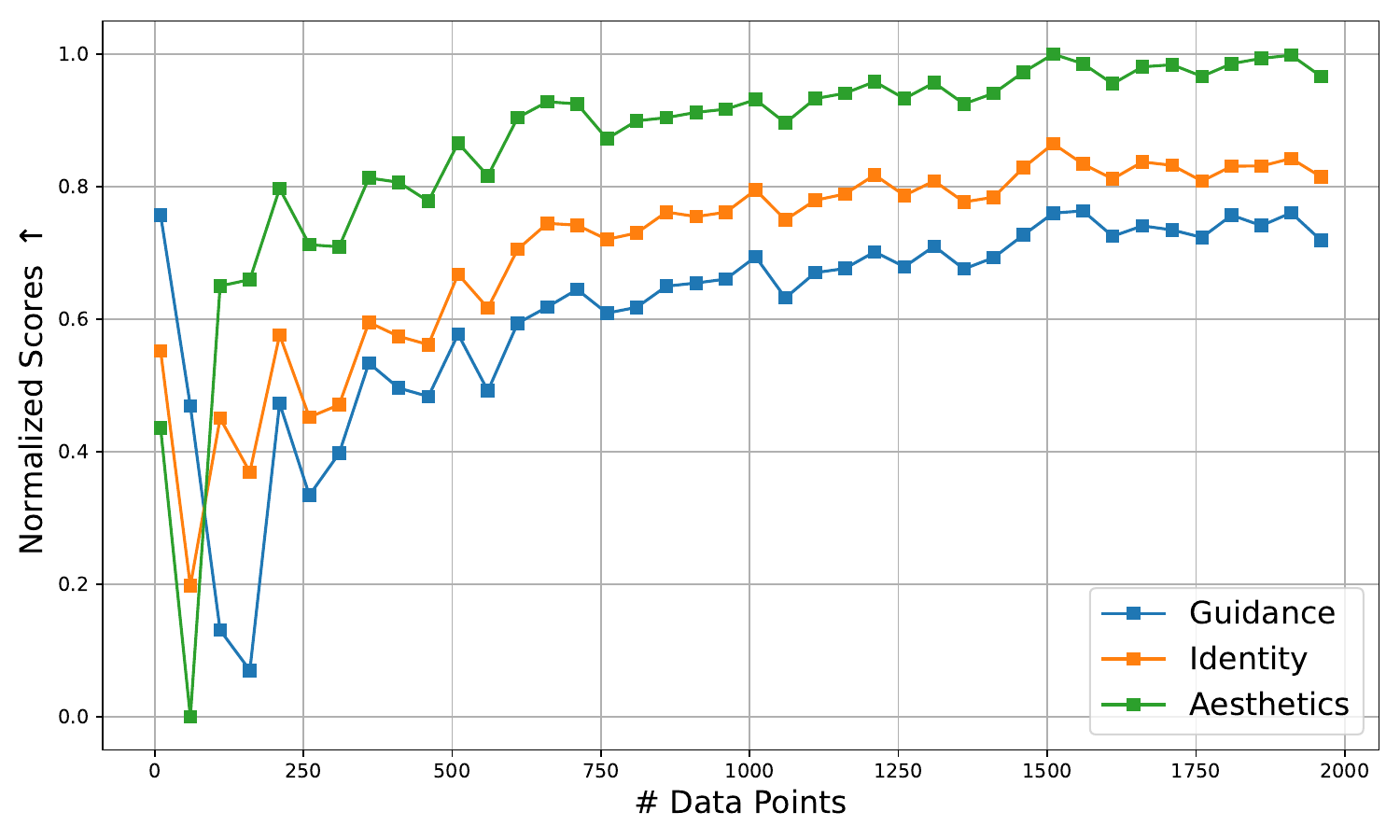}
    \caption{\textbf{Dataset Size Effect.} Quality results can be achieved with only several hundreds of examples. Increasing the number of synthetic data points returns diminishing merits for over 2000 examples, demonstrating data efficiency. Scores were computed by normalizing and averaging all metrics from \Cref{tab:quantitative}, with a LoRA rank of 32.}
    \label{fig:data_scaling}
\end{figure}

We performed an ablation and design choice study, by measuring the relighting performance on the evaluation dataset (\Cref{sec:eval_dataset}) using the same metrics used in the quantitative evaluation (\Cref{sec:quantitative_results}). The results are shown in \Cref{tab:ablations}. $I_s$ indicates metric measurements on the input images, shown for reference, and ``Blending'' is a naive 50\% blending scheme between $I_s$ and $I_c$, yielding highly non-aesthetic images that resemble both the control and the source. We provide this as another reference point.
``With Cross Attention'' leverages the cross-attention layers in addition to the self-attention ones . This yields on par performance despite requiring double the parameters, indicating light information is not encoded as well in these layers, as also suggested by our analysis. 
``No Scheduling'' removes all scheduling from the guidance during inference (no gradient normalization, $\gamma_t=1, \space\forall t$). This results in images that are barely edited at all as indicated by the lower control metrics. $\gamma_t=4.0$ uses a larger classifier guidance scaling factor (our full solution uses $2.2$). While this results in better adherence to the control, it also degrades the aesthetics and identity preservation significantly. ``No Query Injection'' removes the query injection used during inference, resulting in sever loss of identity. Similarly, ``No ControlNet'' removes the edge-conditioned ControlNet also used for identity preservation. ``LoRA Rank=4'' and ``LoRA Rank=16'' are versions of the regressor with different LoRA ranks (our solution uses rank=$8$). While the rank 4 version seems promising, in practice we found the control adherence to be less convincing. ``Noisy Labels'' is a regressor trained to predict noisy labels $I_g$ instead of the clean labels. Additionally, we show the effects of using different number of data points to train the regressor in \Cref{fig:data_scaling}. The regressor was trained with LoRA rank=32, with different number of synthetic instances, and performance was measured by normalizing and averaging all metrics in \Cref{tab:quantitative}. Results suggest the model is data efficient, with useful signals extracted already from as little as several hundreds of examples, and saturates quickly with a few thousands.

\section{Discussion}

This paper presented a practical relighting method that relies on the light transport priors learned by a pre-trained latent diffusion model. We showed that large diffusion models learned important components of light transport and more importantly, we show how to extract this information and use it for relighting. Our data- and parameter-efficient solution generalizes better compared to related state-of-the-art approaches, relying on extensive in-domain training. In the future, we believe this line of investigation could further inject light interaction and physical properties into the generation process.

\paragraph{Limitations and Future Work.}
While our method exhibits a moderate level of control over high frequency elements such as shadows and highlights, we found it much simpler to add them rather than removing them completely. An interesting future work could explore how to address this issue more thoroughly.

Although we present relighting results using a UNet-based diffusion model, our approach is readily extensible to DiT-based architectures. This follows from the central premise that attention layers provide a natural inductive bias for modeling light transport through their many-to-many interactions (see supplemental material for preliminary evidence). 

Our method does not currently support colored light. We found colored light to require significantly more data, likely due to requiring the regressor to disentangle material reflectance and light color. We also found this to yield global color drifts in the resulting relit image, an artifact that was observed in previous studies \cite{readout_guidance}. We also note that we found the regressor predictions to consistently be tinted in brown, which may stem from limitations in the diffusion model’s dynamic range \cite{flaw}, or simply a bias from the training data and process.

Additionally, controlling global illumination phenomena (e.g., inter-reflections) is still an open question. We believe it could be useful to include higher-order illumination effects in the training set to address this problem. 

Furthermore, the analysis suggests that there may be a simpler way to manipulate light in the image without optimization, as some heads are highly associated with light in the scene (see supplemental material). Manipulating them directly for surgical control could be an interesting future direction.

\section{Acknowledgments}
This research was supported in part by the Len Blavatnik and the Blavatnik family foundation and ISF number 1337/22.

\bibliographystyle{ACM-Reference-Format}
\bibliography{egbib}


\begin{thebibliography}{49}


\ifx \showCODEN    \undefined \def \showCODEN     #1{\unskip}     \fi
\ifx \showDOI      \undefined \def \showDOI       #1{#1}\fi
\ifx \showISBNx    \undefined \def \showISBNx     #1{\unskip}     \fi
\ifx \showISBNxiii \undefined \def \showISBNxiii  #1{\unskip}     \fi
\ifx \showISSN     \undefined \def \showISSN      #1{\unskip}     \fi
\ifx \showLCCN     \undefined \def \showLCCN      #1{\unskip}     \fi
\ifx \shownote     \undefined \def \shownote      #1{#1}          \fi
\ifx \showarticletitle \undefined \def \showarticletitle #1{#1}   \fi
\ifx \showURL      \undefined \def \showURL       {\relax}        \fi
\providecommand\bibfield[2]{#2}
\providecommand\bibinfo[2]{#2}
\providecommand\natexlab[1]{#1}
\providecommand\showeprint[2][]{arXiv:#2}

\bibitem[Alaluf et~al\mbox{.}(2024)]%
        {cia}
\bibfield{author}{\bibinfo{person}{Yuval Alaluf}, \bibinfo{person}{Daniel Garibi}, \bibinfo{person}{Or Patashnik}, \bibinfo{person}{Hadar Averbuch-Elor}, {and} \bibinfo{person}{Daniel Cohen-Or}.} \bibinfo{year}{2024}\natexlab{}.
\newblock \showarticletitle{Cross-Image Attention for Zero-Shot Appearance Transfer}. In \bibinfo{booktitle}{\emph{ACM SIGGRAPH 2024 Conference Papers}} (Denver, CO, USA) \emph{(\bibinfo{series}{SIGGRAPH '24})}. \bibinfo{publisher}{Association for Computing Machinery}, \bibinfo{address}{New York, NY, USA}, Article \bibinfo{articleno}{132}, \bibinfo{numpages}{12}~pages.
\newblock
\showISBNx{9798400705250}
\urldef\tempurl%
\url{https://doi.org/10.1145/3641519.3657423}
\showDOI{\tempurl}


\bibitem[Bhat et~al\mbox{.}(2023)]%
        {zoe_depth}
\bibfield{author}{\bibinfo{person}{Shariq~Farooq Bhat}, \bibinfo{person}{Reiner Birkl}, \bibinfo{person}{Diana Wofk}, \bibinfo{person}{Peter Wonka}, {and} \bibinfo{person}{Matthias Müller}.} \bibinfo{year}{2023}\natexlab{}.
\newblock \bibinfo{title}{ZoeDepth: Zero-shot Transfer by Combining Relative and Metric Depth}.
\newblock
\newblock
\showeprint[arxiv]{2302.12288}~[cs.CV]
\urldef\tempurl%
\url{https://arxiv.org/abs/2302.12288}
\showURL{%
\tempurl}


\bibitem[Careaga and Aksoy(2023)]%
        {ordinal}
\bibfield{author}{\bibinfo{person}{Chris Careaga} {and} \bibinfo{person}{Ya\u{g}{\i}z Aksoy}.} \bibinfo{year}{2023}\natexlab{}.
\newblock \showarticletitle{Intrinsic Image Decomposition via Ordinal Shading}.
\newblock \bibinfo{journal}{\emph{ACM Trans. Graph.}} \bibinfo{volume}{43}, \bibinfo{number}{1}, Article \bibinfo{articleno}{12} (\bibinfo{year}{2023}), \bibinfo{numpages}{24}~pages.
\newblock


\bibitem[Careaga and Aksoy(2024)]%
        {cd-iid}
\bibfield{author}{\bibinfo{person}{Chris Careaga} {and} \bibinfo{person}{Ya\u{g}{\i}z Aksoy}.} \bibinfo{year}{2024}\natexlab{}.
\newblock \showarticletitle{Colorful Diffuse Intrinsic Image Decomposition in the Wild}.
\newblock \bibinfo{journal}{\emph{ACM Trans. Graph.}} \bibinfo{volume}{43}, \bibinfo{number}{6}, Article \bibinfo{articleno}{178} (\bibinfo{year}{2024}), \bibinfo{numpages}{12}~pages.
\newblock


\bibitem[Caron et~al\mbox{.}(2021)]%
        {dino}
\bibfield{author}{\bibinfo{person}{Mathilde Caron}, \bibinfo{person}{Hugo Touvron}, \bibinfo{person}{Ishan Misra}, \bibinfo{person}{Hervé Jegou}, \bibinfo{person}{Julien Mairal}, \bibinfo{person}{Piotr Bojanowski}, {and} \bibinfo{person}{Armand Joulin}.} \bibinfo{year}{2021}\natexlab{}.
\newblock \showarticletitle{Emerging Properties in Self-Supervised Vision Transformers}. In \bibinfo{booktitle}{\emph{2021 IEEE/CVF International Conference on Computer Vision (ICCV)}}. \bibinfo{pages}{9630--9640}.
\newblock
\urldef\tempurl%
\url{https://doi.org/10.1109/ICCV48922.2021.00951}
\showDOI{\tempurl}


\bibitem[Choi et~al\mbox{.}(2024)]%
        {scribble_light}
\bibfield{author}{\bibinfo{person}{Jun~Myeong Choi}, \bibinfo{person}{Annie Wang}, \bibinfo{person}{Pieter Peers}, \bibinfo{person}{Anand Bhattad}, {and} \bibinfo{person}{Roni Sengupta}.} \bibinfo{year}{2024}\natexlab{}.
\newblock \bibinfo{title}{ScribbleLight: Single Image Indoor Relighting with Scribbles}.
\newblock
\newblock
\showeprint[arxiv]{2411.17696}~[cs.CV]
\urldef\tempurl%
\url{https://arxiv.org/abs/2411.17696}
\showURL{%
\tempurl}


\bibitem[Community(2018)]%
        {blender}
\bibfield{author}{\bibinfo{person}{Blender~Online Community}.} \bibinfo{year}{2018}\natexlab{}.
\newblock \bibinfo{booktitle}{\emph{Blender - a 3D modelling and rendering package}}.
\newblock Blender Foundation, Stichting Blender Foundation, Amsterdam.
\newblock
\urldef\tempurl%
\url{http://www.blender.org}
\showURL{%
\tempurl}


\bibitem[Debevec et~al\mbox{.}(2000)]%
        {light_stage}
\bibfield{author}{\bibinfo{person}{Paul Debevec}, \bibinfo{person}{Tim Hawkins}, \bibinfo{person}{Chris Tchou}, \bibinfo{person}{Haarm-Pieter Duiker}, \bibinfo{person}{Westley Sarokin}, {and} \bibinfo{person}{Mark Sagar}.} \bibinfo{year}{2000}\natexlab{}.
\newblock \showarticletitle{Acquiring the reflectance field of a human face}. In \bibinfo{booktitle}{\emph{Proceedings of the 27th annual conference on Computer graphics and interactive techniques}}. \bibinfo{pages}{145--156}.
\newblock


\bibitem[Dhariwal and Nichol(2021)]%
        {dhariwal2021diffusion}
\bibfield{author}{\bibinfo{person}{Prafulla Dhariwal} {and} \bibinfo{person}{Alexander Nichol}.} \bibinfo{year}{2021}\natexlab{}.
\newblock \showarticletitle{Diffusion models beat gans on image synthesis}.
\newblock \bibinfo{journal}{\emph{Advances in neural information processing systems}}  \bibinfo{volume}{34} (\bibinfo{year}{2021}), \bibinfo{pages}{8780--8794}.
\newblock


\bibitem[Discus0434(2025)]%
        {aesthetic_score}
\bibfield{author}{\bibinfo{person}{Discus0434}.} \bibinfo{year}{2025}\natexlab{}.
\newblock \bibinfo{title}{Aesthetic Predictor v2.5}.
\newblock \bibinfo{howpublished}{\url{https://github.com/discus0434/aesthetic-predictor-v2-5}}.
\newblock
\newblock
\shownote{Accessed: 2025-01-14}.


\bibitem[Du et~al\mbox{.}(2023)]%
        {ilora}
\bibfield{author}{\bibinfo{person}{Xiaodan Du}, \bibinfo{person}{Nicholas Kolkin}, \bibinfo{person}{Greg Shakhnarovich}, {and} \bibinfo{person}{Anand Bhattad}.} \bibinfo{year}{2023}\natexlab{}.
\newblock \showarticletitle{Generative Models: What do they know? Do they know things? Let's find out!}
\newblock \bibinfo{journal}{\emph{arXiv preprint arXiv:2311.17137}} (\bibinfo{year}{2023}).
\newblock


\bibitem[Erel et~al\mbox{.}(2025)]%
        {attention_chains}
\bibfield{author}{\bibinfo{person}{Yotam Erel}, \bibinfo{person}{Olaf D{\"u}nkel}, \bibinfo{person}{Rishabh Dabral}, \bibinfo{person}{Vladislav Golyanik}, \bibinfo{person}{Christian Theobalt}, {and} \bibinfo{person}{Amit~H. Bermano}.} \bibinfo{year}{2025}\natexlab{}.
\newblock \showarticletitle{Attention (as Discrete-Time Markov) Chains}.
\newblock \bibinfo{journal}{\emph{arXiv preprint arXiv:2507.17657}} (\bibinfo{year}{2025}).
\newblock


\bibitem[Fu et~al\mbox{.}(2023)]%
        {dreamsim}
\bibfield{author}{\bibinfo{person}{Stephanie Fu}, \bibinfo{person}{Netanel Tamir}, \bibinfo{person}{Shobhita Sundaram}, \bibinfo{person}{Lucy Chai}, \bibinfo{person}{Richard Zhang}, \bibinfo{person}{Tali Dekel}, {and} \bibinfo{person}{Phillip Isola}.} \bibinfo{year}{2023}\natexlab{}.
\newblock \showarticletitle{DreamSim: Learning New Dimensions of Human Visual Similarity using Synthetic Data}. In \bibinfo{booktitle}{\emph{Advances in Neural Information Processing Systems}}, Vol.~\bibinfo{volume}{36}. \bibinfo{pages}{50742--50768}.
\newblock


\bibitem[Futschik et~al\mbox{.}(2023)]%
        {dir_light_estimator}
\bibfield{author}{\bibinfo{person}{David Futschik}, \bibinfo{person}{Kelvin Ritland}, \bibinfo{person}{James Vecore}, \bibinfo{person}{Sean Fanello}, \bibinfo{person}{Sergio Orts-Escolano}, \bibinfo{person}{Brian Curless}, \bibinfo{person}{Daniel S{\`y}kora}, {and} \bibinfo{person}{Rohit Pandey}.} \bibinfo{year}{2023}\natexlab{}.
\newblock \showarticletitle{Controllable light diffusion for portraits}. In \bibinfo{booktitle}{\emph{Proceedings of the IEEE/CVF Conference on Computer Vision and Pattern Recognition}}. \bibinfo{pages}{8412--8421}.
\newblock


\bibitem[Gandikota et~al\mbox{.}(2025)]%
        {concept_lora}
\bibfield{author}{\bibinfo{person}{Rohit Gandikota}, \bibinfo{person}{Joanna Materzy{\'n}ska}, \bibinfo{person}{Tingrui Zhou}, \bibinfo{person}{Antonio Torralba}, {and} \bibinfo{person}{David Bau}.} \bibinfo{year}{2025}\natexlab{}.
\newblock \showarticletitle{Concept sliders: Lora adaptors for precise control in diffusion models}. In \bibinfo{booktitle}{\emph{European Conference on Computer Vision}}. Springer, \bibinfo{pages}{172--188}.
\newblock


\bibitem[Griffiths et~al\mbox{.}(2022)]%
        {outcast}
\bibfield{author}{\bibinfo{person}{David Griffiths}, \bibinfo{person}{Tobias Ritschel}, {and} \bibinfo{person}{Julien Philip}.} \bibinfo{year}{2022}\natexlab{}.
\newblock \showarticletitle{OutCast: Outdoor Single-image Relighting with Cast Shadows}. In \bibinfo{booktitle}{\emph{Computer Graphics Forum}}, Vol.~\bibinfo{volume}{41}. Wiley Online Library, \bibinfo{pages}{179--193}.
\newblock


\bibitem[Hertz et~al\mbox{.}(2022)]%
        {p2p}
\bibfield{author}{\bibinfo{person}{Amir Hertz}, \bibinfo{person}{Ron Mokady}, \bibinfo{person}{Jay Tenenbaum}, \bibinfo{person}{Kfir Aberman}, \bibinfo{person}{Yael Pritch}, {and} \bibinfo{person}{Daniel Cohen-Or}.} \bibinfo{year}{2022}\natexlab{}.
\newblock \showarticletitle{Prompt-to-prompt image editing with cross attention control}.
\newblock \bibinfo{journal}{\emph{arXiv preprint arXiv:2208.01626}} (\bibinfo{year}{2022}).
\newblock


\bibitem[Hu et~al\mbox{.}(2021)]%
        {lora}
\bibfield{author}{\bibinfo{person}{Edward~J Hu}, \bibinfo{person}{Yelong Shen}, \bibinfo{person}{Phillip Wallis}, \bibinfo{person}{Zeyuan Allen-Zhu}, \bibinfo{person}{Yuanzhi Li}, \bibinfo{person}{Shean Wang}, \bibinfo{person}{Lu Wang}, {and} \bibinfo{person}{Weizhu Chen}.} \bibinfo{year}{2021}\natexlab{}.
\newblock \showarticletitle{Lora: Low-rank adaptation of large language models}.
\newblock \bibinfo{journal}{\emph{arXiv preprint arXiv:2106.09685}} (\bibinfo{year}{2021}).
\newblock


\bibitem[Jin et~al\mbox{.}(2024)]%
        {neural-gaffer}
\bibfield{author}{\bibinfo{person}{Haian Jin}, \bibinfo{person}{Yuan Li}, \bibinfo{person}{Fujun Luan}, \bibinfo{person}{Yuanbo Xiangli}, \bibinfo{person}{Sai Bi}, \bibinfo{person}{Kai Zhang}, \bibinfo{person}{Zexiang Xu}, \bibinfo{person}{Jin Sun}, {and} \bibinfo{person}{Noah Snavely}.} \bibinfo{year}{2024}\natexlab{}.
\newblock \showarticletitle{Neural Gaffer: Relighting Any Object via Diffusion}.
\newblock \bibinfo{journal}{\emph{arXiv preprint arXiv:2406.07520}} (\bibinfo{year}{2024}).
\newblock


\bibitem[Kim et~al\mbox{.}(2024)]%
        {switchlight}
\bibfield{author}{\bibinfo{person}{Hoon Kim}, \bibinfo{person}{Minje Jang}, \bibinfo{person}{Wonjun Yoon}, \bibinfo{person}{Jisoo Lee}, \bibinfo{person}{Donghyun Na}, {and} \bibinfo{person}{Sanghyun Woo}.} \bibinfo{year}{2024}\natexlab{}.
\newblock \showarticletitle{SwitchLight: Co-design of Physics-driven Architecture and Pre-training Framework for Human Portrait Relighting}. In \bibinfo{booktitle}{\emph{Proceedings of the IEEE/CVF Conference on Computer Vision and Pattern Recognition}}. \bibinfo{pages}{25096--25106}.
\newblock


\bibitem[Kocsis et~al\mbox{.}(2024)]%
        {lightit}
\bibfield{author}{\bibinfo{person}{Peter Kocsis}, \bibinfo{person}{Julien Philip}, \bibinfo{person}{Kalyan Sunkavalli}, \bibinfo{person}{Matthias Nie{\ss}ner}, {and} \bibinfo{person}{Yannick Hold-Geoffroy}.} \bibinfo{year}{2024}\natexlab{}.
\newblock \showarticletitle{Lightit: Illumination modeling and control for diffusion models}. In \bibinfo{booktitle}{\emph{Proceedings of the IEEE/CVF Conference on Computer Vision and Pattern Recognition}}. \bibinfo{pages}{9359--9369}.
\newblock


\bibitem[Labs(2024)]%
        {flux}
\bibfield{author}{\bibinfo{person}{Black~Forest Labs}.} \bibinfo{year}{2024}\natexlab{}.
\newblock \bibinfo{title}{FLUX}.
\newblock \bibinfo{howpublished}{\url{https://github.com/black-forest-labs/flux}}.
\newblock


\bibitem[Li et~al\mbox{.}(2024)]%
        {personalized_lora}
\bibfield{author}{\bibinfo{person}{Likun Li}, \bibinfo{person}{Haoqi Zeng}, \bibinfo{person}{Changpeng Yang}, \bibinfo{person}{Haozhe Jia}, {and} \bibinfo{person}{Di Xu}.} \bibinfo{year}{2024}\natexlab{}.
\newblock \showarticletitle{Block-wise LoRA: Revisiting Fine-grained LoRA for Effective Personalization and Stylization in Text-to-Image Generation}.
\newblock \bibinfo{journal}{\emph{arXiv preprint arXiv:2403.07500}} (\bibinfo{year}{2024}).
\newblock


\bibitem[Lin et~al\mbox{.}(2024)]%
        {flaw}
\bibfield{author}{\bibinfo{person}{Shanchuan Lin}, \bibinfo{person}{Bingchen Liu}, \bibinfo{person}{Jiashi Li}, {and} \bibinfo{person}{Xiao Yang}.} \bibinfo{year}{2024}\natexlab{}.
\newblock \showarticletitle{Common diffusion noise schedules and sample steps are flawed}. In \bibinfo{booktitle}{\emph{Proceedings of the IEEE/CVF winter conference on applications of computer vision}}. \bibinfo{pages}{5404--5411}.
\newblock


\bibitem[Liu et~al\mbox{.}(2020)]%
        {outdoor_relight}
\bibfield{author}{\bibinfo{person}{Andrew Liu}, \bibinfo{person}{Shiry Ginosar}, \bibinfo{person}{Tinghui Zhou}, \bibinfo{person}{Alexei~A Efros}, {and} \bibinfo{person}{Noah Snavely}.} \bibinfo{year}{2020}\natexlab{}.
\newblock \showarticletitle{Learning to factorize and relight a city}. In \bibinfo{booktitle}{\emph{Computer Vision--ECCV 2020: 16th European Conference, Glasgow, UK, August 23--28, 2020, Proceedings, Part IV 16}}. Springer, \bibinfo{pages}{544--561}.
\newblock


\bibitem[Lu et~al\mbox{.}(2024)]%
        {concept_lora2}
\bibfield{author}{\bibinfo{person}{Shilin Lu}, \bibinfo{person}{Zilan Wang}, \bibinfo{person}{Leyang Li}, \bibinfo{person}{Yanzhu Liu}, {and} \bibinfo{person}{Adams Wai-Kin Kong}.} \bibinfo{year}{2024}\natexlab{}.
\newblock \showarticletitle{Mace: Mass concept erasure in diffusion models}. In \bibinfo{booktitle}{\emph{Proceedings of the IEEE/CVF Conference on Computer Vision and Pattern Recognition}}. \bibinfo{pages}{6430--6440}.
\newblock


\bibitem[Luo et~al\mbox{.}(2024a)]%
        {readout_guidance}
\bibfield{author}{\bibinfo{person}{Grace Luo}, \bibinfo{person}{Trevor Darrell}, \bibinfo{person}{Oliver Wang}, \bibinfo{person}{Dan~B Goldman}, {and} \bibinfo{person}{Aleksander Holynski}.} \bibinfo{year}{2024}\natexlab{a}.
\newblock \showarticletitle{Readout guidance: Learning control from diffusion features}. In \bibinfo{booktitle}{\emph{Proceedings of the IEEE/CVF Conference on Computer Vision and Pattern Recognition}}. \bibinfo{pages}{8217--8227}.
\newblock


\bibitem[Luo et~al\mbox{.}(2024b)]%
        {hyper_features}
\bibfield{author}{\bibinfo{person}{Grace Luo}, \bibinfo{person}{Lisa Dunlap}, \bibinfo{person}{Dong~Huk Park}, \bibinfo{person}{Aleksander Holynski}, {and} \bibinfo{person}{Trevor Darrell}.} \bibinfo{year}{2024}\natexlab{b}.
\newblock \showarticletitle{Diffusion hyperfeatures: Searching through time and space for semantic correspondence}.
\newblock \bibinfo{journal}{\emph{Advances in Neural Information Processing Systems}}  \bibinfo{volume}{36} (\bibinfo{year}{2024}).
\newblock


\bibitem[Magar et~al\mbox{.}(2025)]%
        {lightlab}
\bibfield{author}{\bibinfo{person}{Nadav Magar}, \bibinfo{person}{Amir Hertz}, \bibinfo{person}{Eric Tabellion}, \bibinfo{person}{Yael Pritch}, \bibinfo{person}{Alex Rav-Acha}, \bibinfo{person}{Ariel Shamir}, {and} \bibinfo{person}{Yedid Hoshen}.} \bibinfo{year}{2025}\natexlab{}.
\newblock \showarticletitle{LightLab: Controlling Light Sources in Images with Diffusion Models}.
\newblock \bibinfo{journal}{\emph{arXiv preprint arXiv:2505.09608}} (\bibinfo{year}{2025}).
\newblock


\bibitem[Ponglertnapakorn et~al\mbox{.}(2023)]%
        {diffusion_face_relight}
\bibfield{author}{\bibinfo{person}{Puntawat Ponglertnapakorn}, \bibinfo{person}{Nontawat Tritrong}, {and} \bibinfo{person}{Supasorn Suwajanakorn}.} \bibinfo{year}{2023}\natexlab{}.
\newblock \showarticletitle{DiFaReli: Diffusion face relighting}. In \bibinfo{booktitle}{\emph{Proceedings of the IEEE/CVF International Conference on Computer Vision}}. \bibinfo{pages}{22646--22657}.
\newblock


\bibitem[Radford et~al\mbox{.}(2021)]%
        {clip}
\bibfield{author}{\bibinfo{person}{Alec Radford}, \bibinfo{person}{Jong~Wook Kim}, \bibinfo{person}{Chris Hallacy}, \bibinfo{person}{Aditya Ramesh}, \bibinfo{person}{Gabriel Goh}, \bibinfo{person}{Sandhini Agarwal}, \bibinfo{person}{Girish Sastry}, \bibinfo{person}{Amanda Askell}, \bibinfo{person}{Pamela Mishkin}, \bibinfo{person}{Jack Clark}, {et~al\mbox{.}}} \bibinfo{year}{2021}\natexlab{}.
\newblock \showarticletitle{Learning transferable visual models from natural language supervision}. In \bibinfo{booktitle}{\emph{International conference on machine learning}}. PMLR, \bibinfo{pages}{8748--8763}.
\newblock


\bibitem[Reinhard et~al\mbox{.}(2001)]%
        {color_correction}
\bibfield{author}{\bibinfo{person}{Erik Reinhard}, \bibinfo{person}{Michael Adhikhmin}, \bibinfo{person}{Bruce Gooch}, {and} \bibinfo{person}{Peter Shirley}.} \bibinfo{year}{2001}\natexlab{}.
\newblock \showarticletitle{Color transfer between images}.
\newblock \bibinfo{journal}{\emph{IEEE Computer graphics and applications}} \bibinfo{volume}{21}, \bibinfo{number}{5} (\bibinfo{year}{2001}), \bibinfo{pages}{34--41}.
\newblock


\bibitem[Rombach et~al\mbox{.}(2022)]%
        {sd}
\bibfield{author}{\bibinfo{person}{Robin Rombach}, \bibinfo{person}{Andreas Blattmann}, \bibinfo{person}{Dominik Lorenz}, \bibinfo{person}{Patrick Esser}, {and} \bibinfo{person}{Bj{\"o}rn Ommer}.} \bibinfo{year}{2022}\natexlab{}.
\newblock \showarticletitle{High-resolution image synthesis with latent diffusion models}. In \bibinfo{booktitle}{\emph{Proceedings of the IEEE/CVF conference on computer vision and pattern recognition}}. \bibinfo{pages}{10684--10695}.
\newblock


\bibitem[Ruiz et~al\mbox{.}(2023)]%
        {dreambooth}
\bibfield{author}{\bibinfo{person}{Nataniel Ruiz}, \bibinfo{person}{Yuanzhen Li}, \bibinfo{person}{Varun Jampani}, \bibinfo{person}{Yael Pritch}, \bibinfo{person}{Michael Rubinstein}, {and} \bibinfo{person}{Kfir Aberman}.} \bibinfo{year}{2023}\natexlab{}.
\newblock \showarticletitle{Dreambooth: Fine tuning text-to-image diffusion models for subject-driven generation}. In \bibinfo{booktitle}{\emph{Proceedings of the IEEE/CVF conference on computer vision and pattern recognition}}. \bibinfo{pages}{22500--22510}.
\newblock


\bibitem[Shah et~al\mbox{.}(2025)]%
        {ziplora}
\bibfield{author}{\bibinfo{person}{Viraj Shah}, \bibinfo{person}{Nataniel Ruiz}, \bibinfo{person}{Forrester Cole}, \bibinfo{person}{Erika Lu}, \bibinfo{person}{Svetlana Lazebnik}, \bibinfo{person}{Yuanzhen Li}, {and} \bibinfo{person}{Varun Jampani}.} \bibinfo{year}{2025}\natexlab{}.
\newblock \showarticletitle{Ziplora: Any subject in any style by effectively merging loras}. In \bibinfo{booktitle}{\emph{European Conference on Computer Vision}}. Springer, \bibinfo{pages}{422--438}.
\newblock


\bibitem[Song et~al\mbox{.}(2020)]%
        {ddim}
\bibfield{author}{\bibinfo{person}{Jiaming Song}, \bibinfo{person}{Chenlin Meng}, {and} \bibinfo{person}{Stefano Ermon}.} \bibinfo{year}{2020}\natexlab{}.
\newblock \showarticletitle{Denoising diffusion implicit models}.
\newblock \bibinfo{journal}{\emph{arXiv preprint arXiv:2010.02502}} (\bibinfo{year}{2020}).
\newblock


\bibitem[Voynov et~al\mbox{.}(2023)]%
        {sketch}
\bibfield{author}{\bibinfo{person}{Andrey Voynov}, \bibinfo{person}{Kfir Aberman}, {and} \bibinfo{person}{Daniel Cohen-Or}.} \bibinfo{year}{2023}\natexlab{}.
\newblock \showarticletitle{Sketch-guided text-to-image diffusion models}. In \bibinfo{booktitle}{\emph{ACM SIGGRApH 2023 conference proceedings}}. \bibinfo{pages}{1--11}.
\newblock


\bibitem[Walter et~al\mbox{.}(2007)]%
        {ggx}
\bibfield{author}{\bibinfo{person}{Bruce Walter}, \bibinfo{person}{Stephen~R Marschner}, \bibinfo{person}{Hongsong Li}, {and} \bibinfo{person}{Kenneth~E Torrance}.} \bibinfo{year}{2007}\natexlab{}.
\newblock \showarticletitle{Microfacet Models for Refraction through Rough Surfaces.}
\newblock \bibinfo{journal}{\emph{Rendering techniques}}  \bibinfo{volume}{2007} (\bibinfo{year}{2007}), \bibinfo{pages}{18th}.
\newblock


\bibitem[Wang et~al\mbox{.}(2022)]%
        {diffusiondb}
\bibfield{author}{\bibinfo{person}{Zijie~J Wang}, \bibinfo{person}{Evan Montoya}, \bibinfo{person}{David Munechika}, \bibinfo{person}{Haoyang Yang}, \bibinfo{person}{Benjamin Hoover}, {and} \bibinfo{person}{Duen~Horng Chau}.} \bibinfo{year}{2022}\natexlab{}.
\newblock \showarticletitle{Diffusiondb: A large-scale prompt gallery dataset for text-to-image generative models}.
\newblock \bibinfo{journal}{\emph{arXiv preprint arXiv:2210.14896}} (\bibinfo{year}{2022}).
\newblock


\bibitem[Wu et~al\mbox{.}(2023)]%
        {hpsv2}
\bibfield{author}{\bibinfo{person}{Xiaoshi Wu}, \bibinfo{person}{Yiming Hao}, \bibinfo{person}{Keqiang Sun}, \bibinfo{person}{Yixiong Chen}, \bibinfo{person}{Feng Zhu}, \bibinfo{person}{Rui Zhao}, {and} \bibinfo{person}{Hongsheng Li}.} \bibinfo{year}{2023}\natexlab{}.
\newblock \showarticletitle{Human preference score v2: A solid benchmark for evaluating human preferences of text-to-image synthesis}.
\newblock \bibinfo{journal}{\emph{arXiv preprint arXiv:2306.09341}} (\bibinfo{year}{2023}).
\newblock


\bibitem[Xing et~al\mbox{.}(2024a)]%
        {luminet}
\bibfield{author}{\bibinfo{person}{Xiaoyan Xing}, \bibinfo{person}{Konrad Groh}, \bibinfo{person}{Sezer Karaoglu}, \bibinfo{person}{Theo Gevers}, {and} \bibinfo{person}{Anand Bhattad}.} \bibinfo{year}{2024}\natexlab{a}.
\newblock \bibinfo{title}{LumiNet: Latent Intrinsics Meets Diffusion Models for Indoor Scene Relighting}.
\newblock
\newblock
\showeprint[arxiv]{2412.00177}~[cs.CV]
\urldef\tempurl%
\url{https://arxiv.org/abs/2412.00177}
\showURL{%
\tempurl}


\bibitem[Xing et~al\mbox{.}(2024b)]%
        {retinex}
\bibfield{author}{\bibinfo{person}{Xiaoyan Xing}, \bibinfo{person}{Vincent~Tao Hu}, \bibinfo{person}{Jan~Hendrik Metzen}, \bibinfo{person}{Konrad Groh}, \bibinfo{person}{Sezer Karaoglu}, {and} \bibinfo{person}{Theo Gevers}.} \bibinfo{year}{2024}\natexlab{b}.
\newblock \showarticletitle{Retinex-diffusion: On controlling illumination conditions in diffusion models via retinex theory}.
\newblock \bibinfo{journal}{\emph{arXiv preprint arXiv:2407.20785}} (\bibinfo{year}{2024}).
\newblock


\bibitem[Zeng et~al\mbox{.}(2024b)]%
        {dilightnet}
\bibfield{author}{\bibinfo{person}{Chong Zeng}, \bibinfo{person}{Yue Dong}, \bibinfo{person}{Pieter Peers}, \bibinfo{person}{Youkang Kong}, \bibinfo{person}{Hongzhi Wu}, {and} \bibinfo{person}{Xin Tong}.} \bibinfo{year}{2024}\natexlab{b}.
\newblock \showarticletitle{DiLightNet: Fine-grained Lighting Control for Diffusion-based Image Generation}.
\newblock \bibinfo{journal}{\emph{arXiv preprint arXiv:2402.11929}} (\bibinfo{year}{2024}).
\newblock


\bibitem[Zeng et~al\mbox{.}(2024a)]%
        {rgbx}
\bibfield{author}{\bibinfo{person}{Zheng Zeng}, \bibinfo{person}{Valentin Deschaintre}, \bibinfo{person}{Iliyan Georgiev}, \bibinfo{person}{Yannick Hold-Geoffroy}, \bibinfo{person}{Yiwei Hu}, \bibinfo{person}{Fujun Luan}, \bibinfo{person}{Ling-Qi Yan}, {and} \bibinfo{person}{Milo\v{s} Ha\v{s}an}.} \bibinfo{year}{2024}\natexlab{a}.
\newblock \showarticletitle{RGBX: Image decomposition and synthesis using material- and lighting-aware diffusion models}. In \bibinfo{booktitle}{\emph{ACM SIGGRAPH 2024 Conference Papers}} (Denver, CO, USA) \emph{(\bibinfo{series}{SIGGRAPH '24})}. \bibinfo{publisher}{Association for Computing Machinery}, \bibinfo{address}{New York, NY, USA}, Article \bibinfo{articleno}{75}, \bibinfo{numpages}{11}~pages.
\newblock
\showISBNx{9798400705250}
\urldef\tempurl%
\url{https://doi.org/10.1145/3641519.3657445}
\showDOI{\tempurl}


\bibitem[Zhang et~al\mbox{.}(2023)]%
        {controlnet}
\bibfield{author}{\bibinfo{person}{Lvmin Zhang}, \bibinfo{person}{Anyi Rao}, {and} \bibinfo{person}{Maneesh Agrawala}.} \bibinfo{year}{2023}\natexlab{}.
\newblock \showarticletitle{Adding conditional control to text-to-image diffusion models}. In \bibinfo{booktitle}{\emph{Proceedings of the IEEE/CVF International Conference on Computer Vision}}. \bibinfo{pages}{3836--3847}.
\newblock


\bibitem[Zhang et~al\mbox{.}(2025)]%
        {iclight}
\bibfield{author}{\bibinfo{person}{Lvmin Zhang}, \bibinfo{person}{Anyi Rao}, {and} \bibinfo{person}{Maneesh Agrawala}.} \bibinfo{year}{2025}\natexlab{}.
\newblock \showarticletitle{Scaling In-the-Wild Training for Diffusion-based Illumination Harmonization and Editing by Imposing Consistent Light Transport}. In \bibinfo{booktitle}{\emph{The Thirteenth International Conference on Learning Representations}}.
\newblock
\urldef\tempurl%
\url{https://openreview.net/forum?id=u1cQYxRI1H}
\showURL{%
\tempurl}


\bibitem[Zhang et~al\mbox{.}(2018)]%
        {lpips}
\bibfield{author}{\bibinfo{person}{Richard Zhang}, \bibinfo{person}{Phillip Isola}, \bibinfo{person}{Alexei~A Efros}, \bibinfo{person}{Eli Shechtman}, {and} \bibinfo{person}{Oliver Wang}.} \bibinfo{year}{2018}\natexlab{}.
\newblock \showarticletitle{The Unreasonable Effectiveness of Deep Features as a Perceptual Metric}. In \bibinfo{booktitle}{\emph{CVPR}}.
\newblock


\bibitem[Zhang et~al\mbox{.}(2024)]%
        {latent_intrinsics}
\bibfield{author}{\bibinfo{person}{Xiao Zhang}, \bibinfo{person}{William Gao}, \bibinfo{person}{Seemandhar Jain}, \bibinfo{person}{Michael Maire}, \bibinfo{person}{David Forsyth}, \bibinfo{person}{Anand Bhattad}, {et~al\mbox{.}}} \bibinfo{year}{2024}\natexlab{}.
\newblock \showarticletitle{Latent Intrinsics Emerge from Training to Relight}.
\newblock \bibinfo{journal}{\emph{arXiv preprint arXiv:2405.21074}} (\bibinfo{year}{2024}).
\newblock


\bibitem[Zhou et~al\mbox{.}(2019)]%
        {deep_portrait}
\bibfield{author}{\bibinfo{person}{Hao Zhou}, \bibinfo{person}{Sunil Hadap}, \bibinfo{person}{Kalyan Sunkavalli}, {and} \bibinfo{person}{David~W Jacobs}.} \bibinfo{year}{2019}\natexlab{}.
\newblock \showarticletitle{Deep single-image portrait relighting}. In \bibinfo{booktitle}{\emph{Proceedings of the IEEE/CVF international conference on computer vision}}. \bibinfo{pages}{7194--7202}.
\newblock


\end{thebibliography}

\appendix

\section{Further Implementation Details}
\subsection{Our hyper parameters}
All our experiments use the same hyper-parameters, unless otherwise noted. For the following description, we note $t$ as the normalized timestep, with \textbf{$t=0$} corresponding to \textbf{pure noise} and \textbf{$t=1$} is the \textbf{final image}.

We used query injection for timesteps $t=[0.0, 0.7]$, and ControlNet guidance for timesteps $t=[0.0, 0.6]$ with a ControlNet guidance scale of $0.6$.

The Classifier Free Guidance scale (CFG) was independently set per experiment (or if not mentioned, set to $7.5$):
\begin{itemize}
    \item For the layer and timestep analysis (Sec 3.1 in the main paper), we used CFG=$7.5$ (and $3.5$ for inverting Origin and Target).
    \item For all other experiments (qualitative results, quantitative results and subjective user study), CFG values were set using the hyper parameters from the DiffusionDB dataset \cite{diffusiondb}.
\end{itemize}

We used a LoRA rank=$8$ and alpha=$8$ throughout all experiments (unless otherwise noted). We initialize the LoRA weights with Gaussian initialization.

We used the DDIM sampler \cite{ddim} and $\eta=0.0$ for all experiments.

For all experiments and all methods, we use the negative prompt (unconditional branch): ``artifacts, slow, ugly, blurry, deformed, multiple limbs, pixelated, static, fog, flat, unclear, distorted, error, still, low resolution, oversaturated, grain, blur, morhping, warping''.

\subsection{RGB$\leftrightarrow$x \cite{rgbx}}
We used the official implementation and followed the recommended parameters. We further tuned the following parameters with a grid search: ``image guidance scale''=$0.15$, ``guidance rescale''=$0.1$. Addtionally, we set CFG=$7.5$, and used a fixed total timesteps of $50$. This was because we observed that using the dataset-provided CFG and timesteps heavily saturated or distorted the results.
Every image was first decomposed into all channels (RGB$\rightarrow$x), followed by replacing the irradiance channel with the control signal $I_c$, and recomposing (x$\rightarrow$RGB). We provide a control signal which is the full diffuse render pass (including higher order bounces), different from our direct-light map. The reason we changed the condition is because RGB$\leftrightarrow$x requires a diffuse irradiance map, rather than the (diffuse and glossy) direct light map. The condition was rendered using the exact same light source, material, and 3D height-map.

\subsection{IC-Light \cite{iclight}}
We used the SD1.5 background-conditioned model from the official repository. We did not use any image up-scaling as opposed to the default configurations. We found letting the pipeline mask the input yields better results on average, hence we left this option on. We tuned the CFG and timesteps parameters to $7.5$ and $20$

\subsection{Readout Guidance \cite{readout_guidance}}
We trained a spatial readout head on the exact same data as our method  with the recommended values from the official code. We set $\eta=0.0$ instead of the recommended $\eta=1.0$ for sampling, as we found this yielded better results.

\subsection{ControlNet \cite{controlnet}}
We trained a ControlNet from scratch, trained for 24 GPU hours on our synthetic dataset. We did not witness a sudden convergence in the loss function as described in the original paper, however results did appear to improve with more iterations, up to some point. We believe this is due to lack of sufficient amount of scenes and diversity in the data. During sampling, we use our own identity preservation on top of the ControlNet, as we found it to completely fail to preserve identity otherwise.

\subsection{DiLightNet \cite{dilightnet}}
We used the official implementation of DiLightNet, and generated the radiance hints for every prompt in our evaluation dataset. The generated images and control type used are different compared to the original images from our evaluation dataset and the conditions we use, hence quantitative evaluations were not possible directly. We used an all-ones mask for all generations, as the automatic masking option yielded significantly worse results. We generated the radiance hints using 10 different random point light sources per prompt, and selected the best looking result from each. We note this was a necessary step due to a high failure rate (many of the results appeared heavily distorted, rendering them not usable for comparisons).

\section{User Study}
\begin{figure*}
    \includegraphics[width=0.49\linewidth]{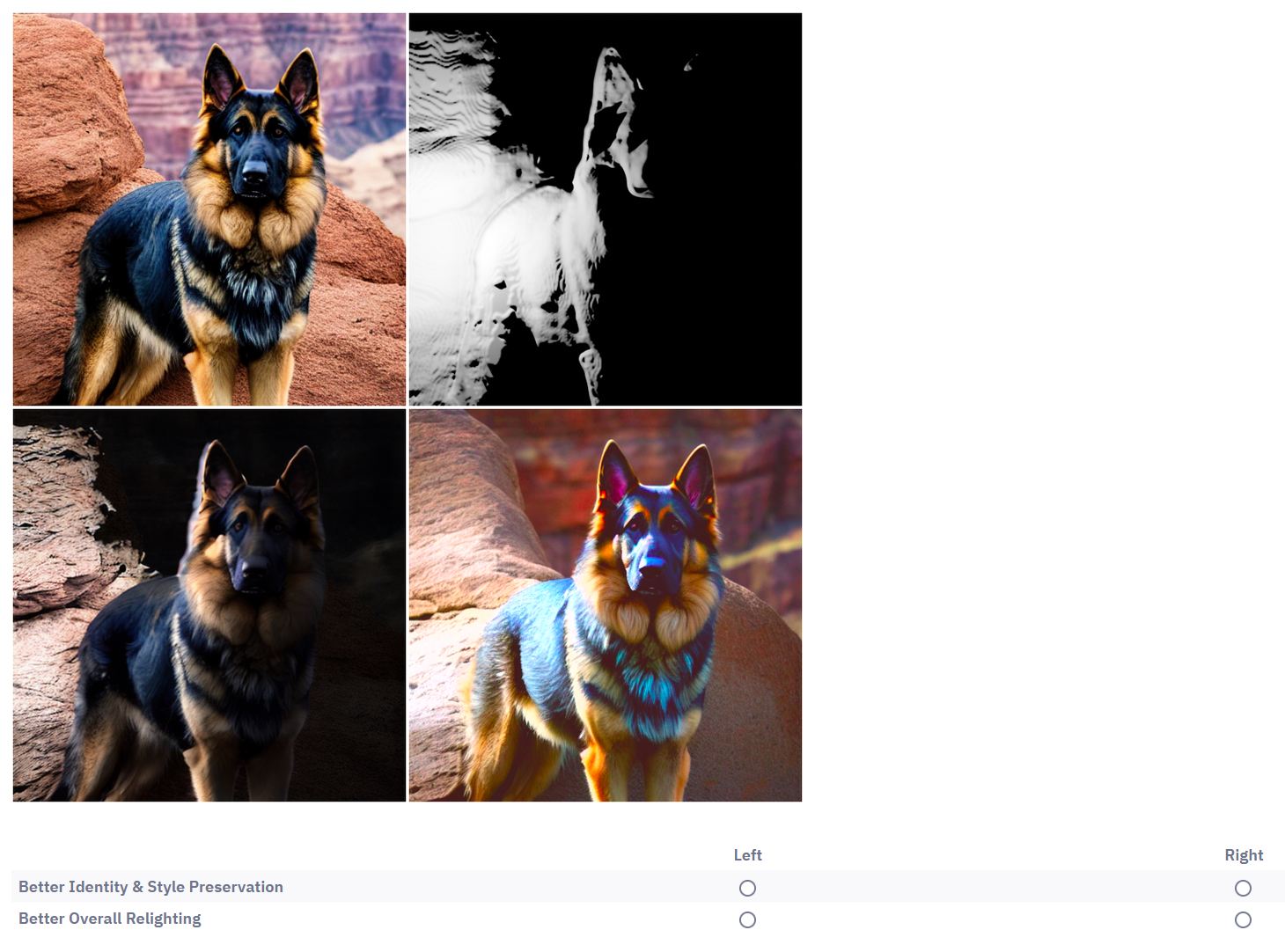}
    \includegraphics[width=0.49\linewidth]{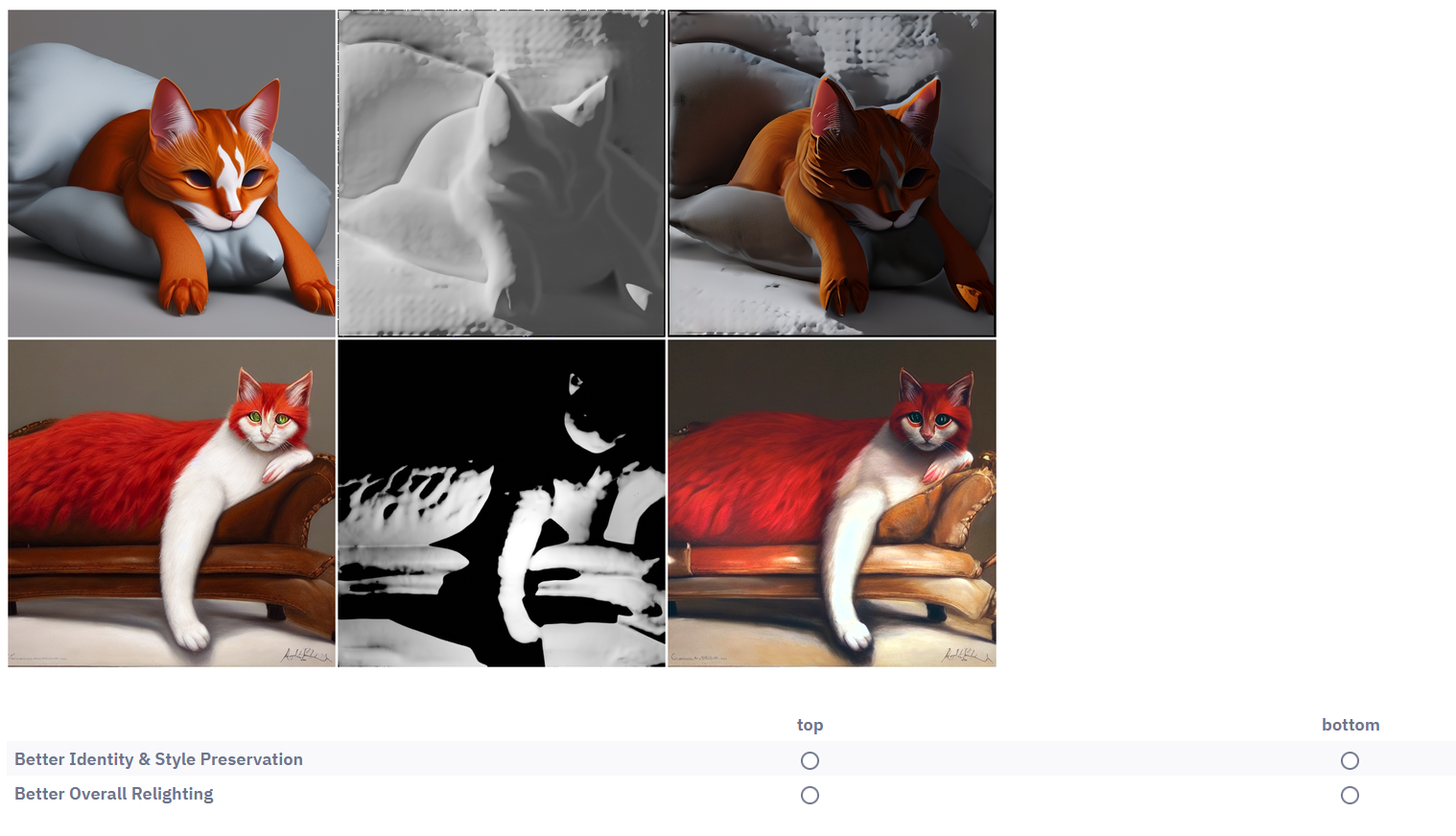}
    \caption{\textbf{User Study Questions. Left:} when comparing against IC-Light \cite{iclight}, subjects were presented with four images, the top two being the source $I_s$ (left) and the light condition $I_c$ (right), while the two bottom images were randomly ordered depicting the relit results $I_r$ (Ours: right, IC-Light: left). \textbf{Right:} For comaprisons with DiLightNet, users were presented with six images, each row consisting of the original $I_s$, the condition $I_c$ and the result $I_r$, where the rows were randomly ordered (Ours: bottom, DiLightNet: top).}
    \label{fig:user_study}
\end{figure*}

See \Cref{fig:user_study} for two exemplar questions posed to subjects who participated in our subjective evaluation.

For comparisons with IC-Light, users were tasked with observing the \textit{source} image generated by the prompt, the target light \textit{condition}, and two relit versions, with their order randomly permuted. They were then asked two binary questions: 
\begin{itemize}
    \item Which image better preserves the identity \& style of the original image?
    \item Which image produces an overall better relighting result?
\end{itemize}
We also introduced two training examples that help explain the task. Overall we had 44 subjects, each comparing a random subset of 15 images.

For DiLightNet, users were tasked with answering the same questions as above while observing two \textit{sources}, two target light \textit{conditions} and two relit results, with their order randomly permuted. This user study had 37 participants, each comparing a random subset of 30 images.

\section{DiT and Light Transport}
\begin{figure}[h]
\centering
{
\setlength{\tabcolsep}{0pt} 
\renewcommand{\arraystretch}{0} 
\begin{tabular}{cccc}
Generated & 1st Bounce & 2nd bounce & 3rd Bounce \\
[0.5em]
\includegraphics[width=0.12\textwidth]{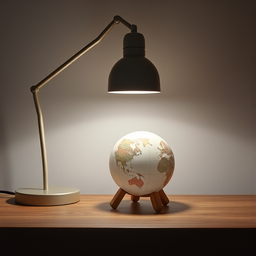} &
\includegraphics[width=0.12\textwidth]{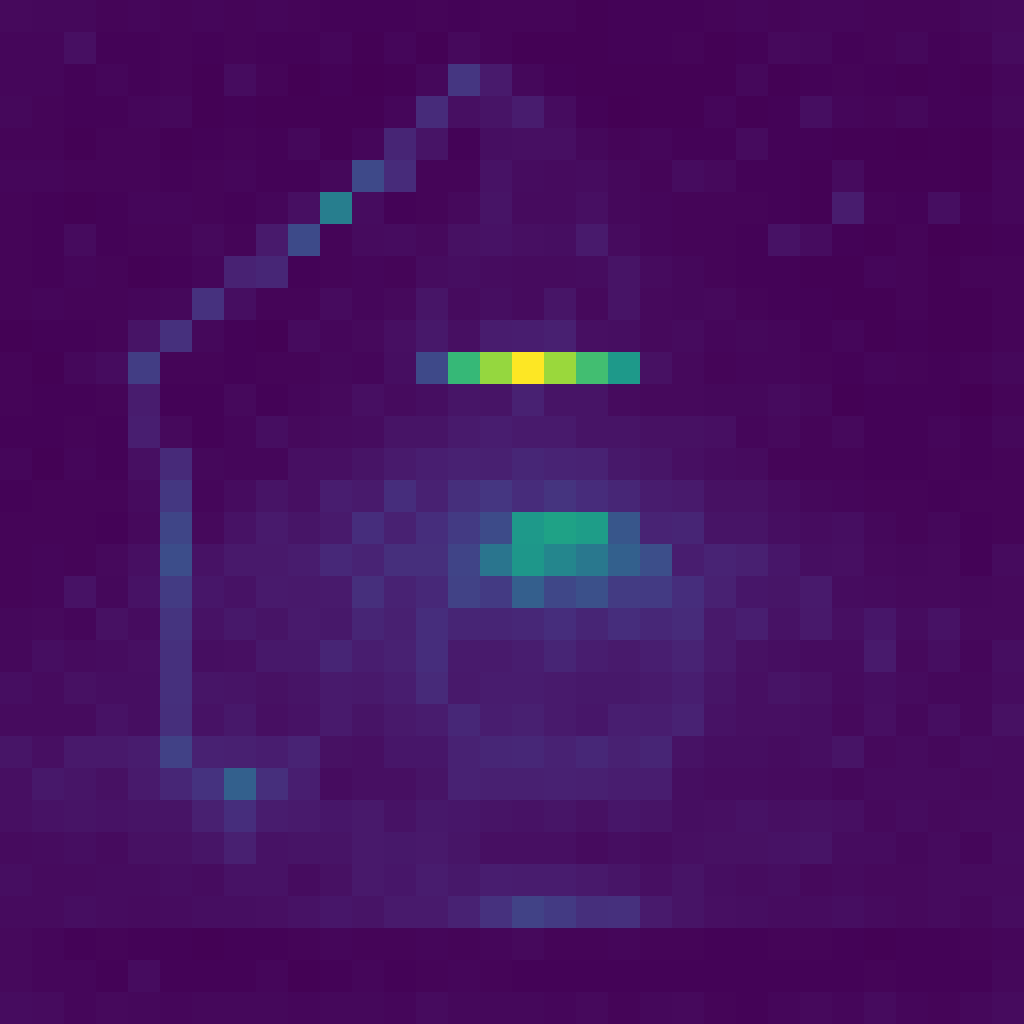} &
\includegraphics[width=0.12\textwidth]{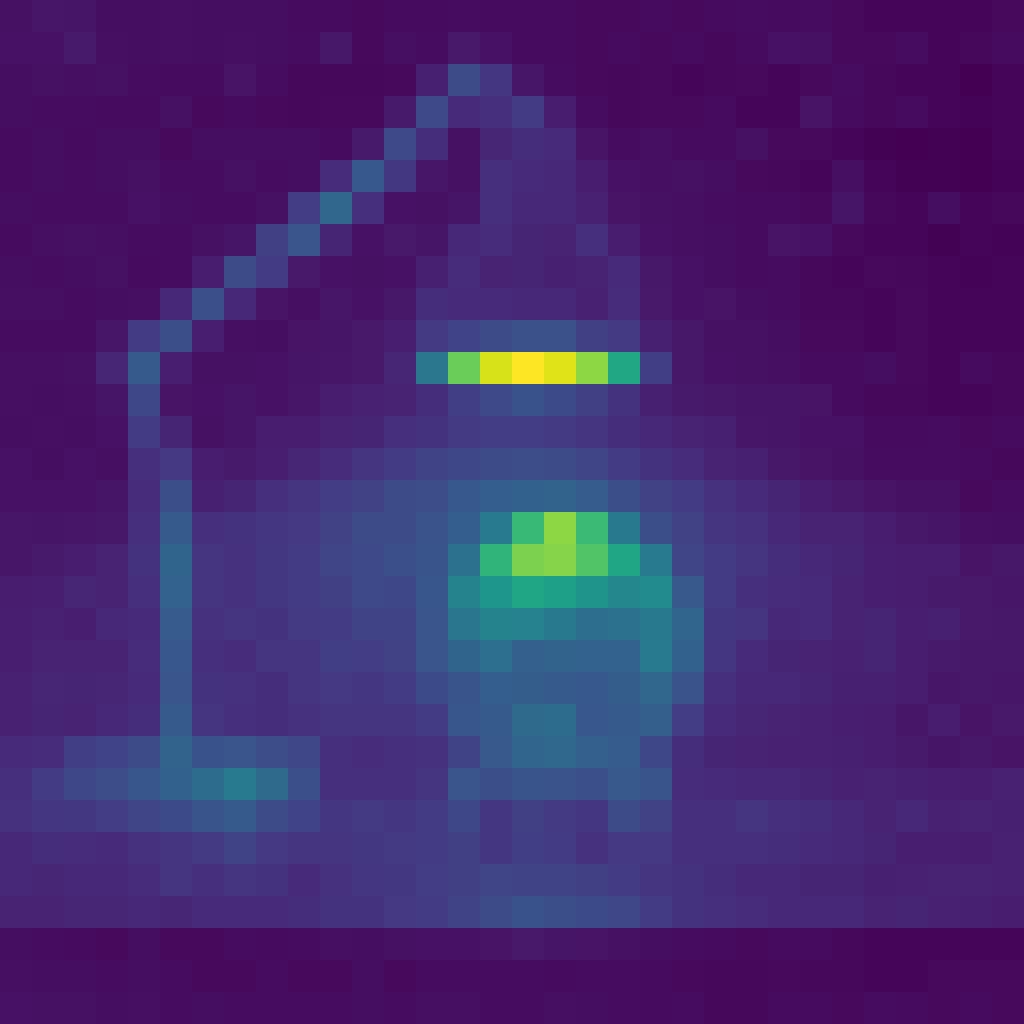} &
\includegraphics[width=0.12\textwidth]{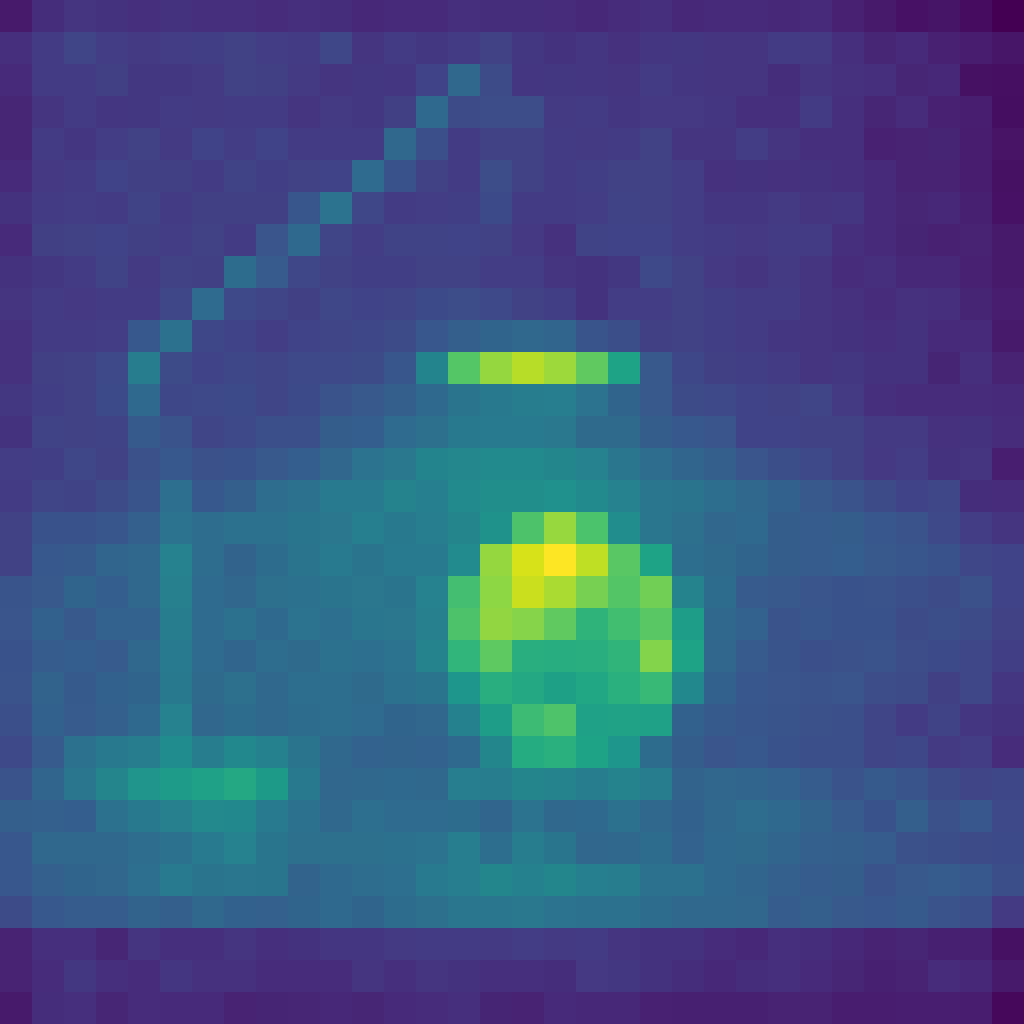}

\end{tabular}
}
\caption{\textbf{DiT and Light Transport.} The spatial tokens attending the text token ``light" in the generated image are shown as ``1st Bounce". We then propagate attention two more times using multi-bounce attention \cite{attention_chains}. Attention propagates in a similar manner to light transport, suggesting our method may be applied to DiT based generators.}
\label{fig:flux}
\end{figure}
While all experiments showcasing \ourmethod's relighting capabilities were performed using the UNet-based SD1.5 \cite{sd}, we believe the same approach could be used with other large diffusion models, including ones based on a diffusion transformer (DiT) architecture such as Flux \cite{flux}. This is because the many-to-many interaction in attention layers serve as a natural inductive bias for encoding light transport, as explained in the main paper, regardless of the architecture. In \Cref{fig:flux}, we show how spatial tokens directly attending the text token ``light" in Flux-schenll are highly concentrated around the light source. We then use the multi-bounce attention mechanism to expose image patches that attend these spatial light source tokens (2nd bounce tokens), and also these which attend the 2nd bounce tokens (3rd bounce tokens). As can be observed, the propagation of attention is similar to the propagation of light, which indicates that light transport phenomena are indeed encoded in attention layers even when the architecture is DiT based. Visualizations were conducted following the multi-bounce procedure defined by \citet{attention_chains}, averaging over the last 10 attention layers of the dual-channel branch of the transformer, and over all timesteps (4 timesteps in total). The prompt for generating this image was ``A lamp shining light on a toy ball on a wooden table.".

\section{More Relighting Results}
\begin{figure*}
    \centering
    {
    \setlength{\tabcolsep}{0pt} 
    \renewcommand{\arraystretch}{0} 
        \begin{tabular}{*{1}{>{\centering\arraybackslash}m{0.16\textwidth}} @{\hskip 5pt} *{5}{>{\centering\arraybackslash}m{0.16\textwidth}}}
        \\[0.5em]
        \includegraphics[width=1.0\linewidth]{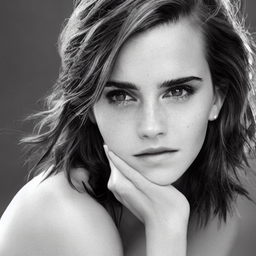} &
        \stackinset{r}{0pt}{b}{0pt}{\color{white}\frame{\includegraphics[width=0.33\linewidth]{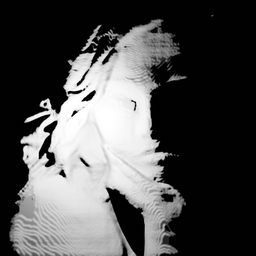}}}{\includegraphics[width=1.0\linewidth]{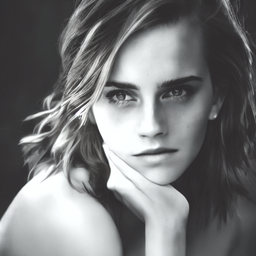}} & 
        \stackinset{r}{0pt}{b}{0pt}{\color{white}\frame{\includegraphics[width=0.33\linewidth]{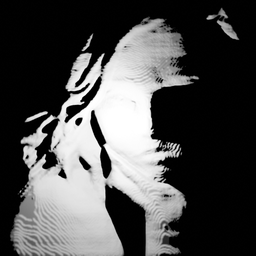}}}{\includegraphics[width=1.0\linewidth]{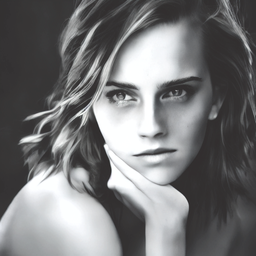}} & 
        \stackinset{r}{0pt}{b}{0pt}{\color{white}\frame{\includegraphics[width=0.33\linewidth]{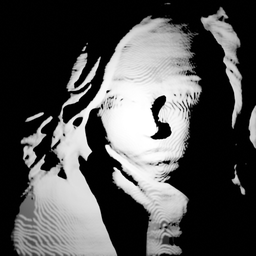}}}{\includegraphics[width=1.0\linewidth]{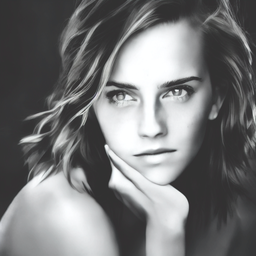}} &
        \stackinset{r}{0pt}{b}{0pt}{\color{white}\frame{\includegraphics[width=0.33\linewidth]{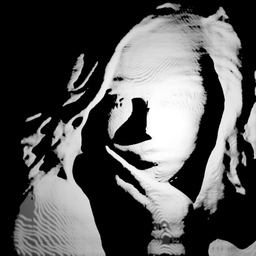}}}{\includegraphics[width=1.0\linewidth]{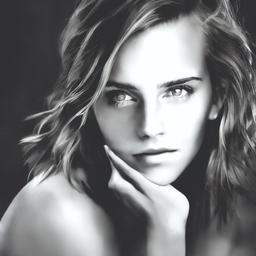}} & 
        \stackinset{r}{0pt}{b}{0pt}{\color{white}\frame{\includegraphics[width=0.33\linewidth]{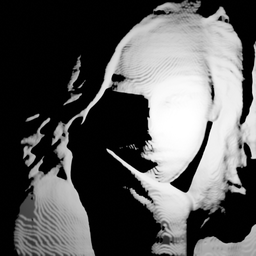}}}{\includegraphics[width=1.0\linewidth]{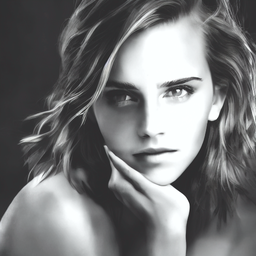}} \\
        \includegraphics[width=1.0\linewidth]{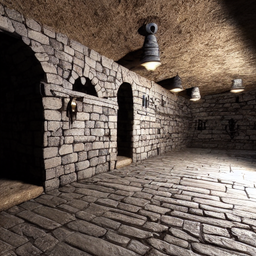} &
        \stackinset{r}{0pt}{b}{0pt}{\color{white}\frame{\includegraphics[width=0.33\linewidth]{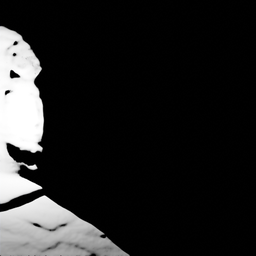}}}{\includegraphics[width=1.0\linewidth]{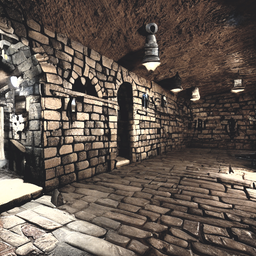}} & 
        \stackinset{r}{0pt}{b}{0pt}{\color{white}\frame{\includegraphics[width=0.33\linewidth]{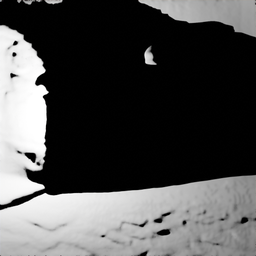}}}{\includegraphics[width=1.0\linewidth]{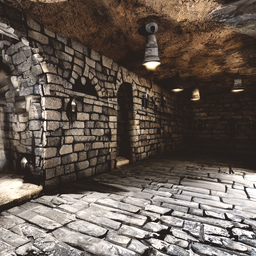}} & 
        \stackinset{r}{0pt}{b}{0pt}{\color{white}\frame{\includegraphics[width=0.33\linewidth]{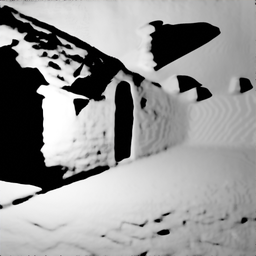}}}{\includegraphics[width=1.0\linewidth]{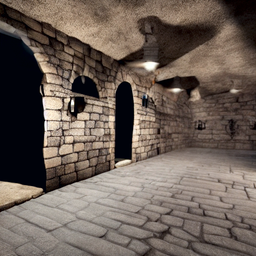}} & 
        \stackinset{r}{0pt}{b}{0pt}{\color{white}\frame{\includegraphics[width=0.33\linewidth]{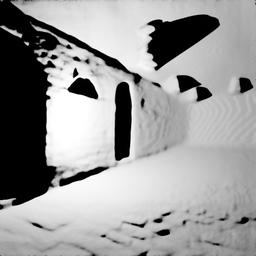}}}{\includegraphics[width=1.0\linewidth]{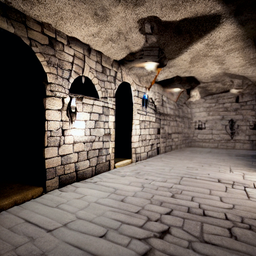}} &         
        \stackinset{r}{0pt}{b}{0pt}{\color{white}\frame{\includegraphics[width=0.33\linewidth]{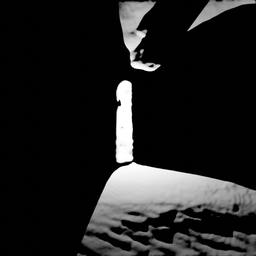}}}{\includegraphics[width=1.0\linewidth]{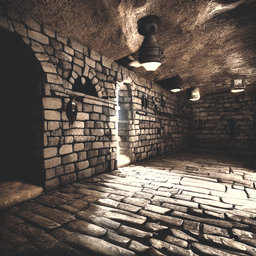}} \\
    \end{tabular}
    }
    \caption{\textbf{Light Probe.} We relight the source $I_S$ (left) using a single light probe in different locations.}
    \label{fig:many_lights_face}
\end{figure*}

\begin{figure*}
  \centering
    \begin{subfigure}{.24\textwidth}
    \centering
    \stackinset{r}{0pt}{b}{0pt}{\color{white}\frame{\includegraphics[width=0.33\textwidth]{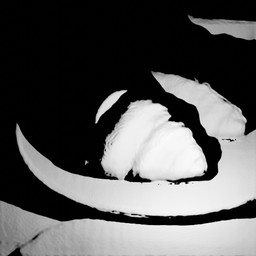}}}{\includegraphics[width=1.0\textwidth]{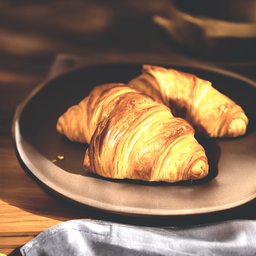}}
    \end{subfigure}
    \begin{subfigure}{.24\textwidth}
    \centering
    \stackinset{r}{0pt}{b}{0pt}{\color{white}\frame{\includegraphics[width=0.33\textwidth]{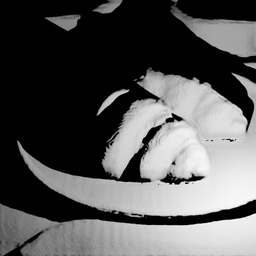}}}{\includegraphics[width=1.0\textwidth]{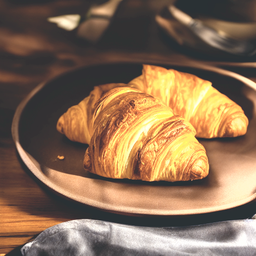}}
    \end{subfigure}
    \begin{subfigure}{.24\textwidth}
    \centering
    \stackinset{r}{0pt}{b}{0pt}{\color{white}\frame{\includegraphics[width=0.33\textwidth]{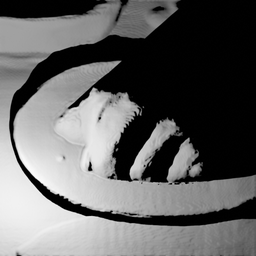}}}{\includegraphics[width=1.0\textwidth]{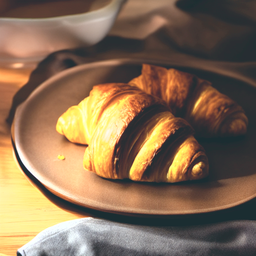}}
    \end{subfigure}
    \begin{subfigure}{.24\textwidth}
    \centering
    \stackinset{r}{0pt}{b}{0pt}{\color{white}\frame{\includegraphics[width=0.33\textwidth]{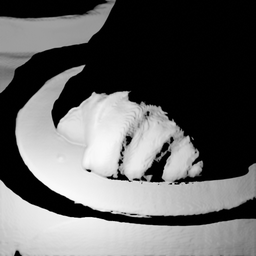}}}{\includegraphics[width=1.0\textwidth]{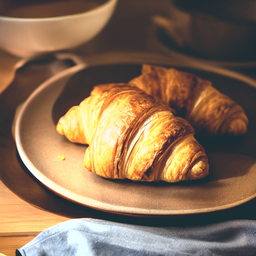}}
    \end{subfigure}
    \begin{subfigure}{.24\textwidth}
    \centering
    \stackinset{r}{0pt}{b}{0pt}{\color{white}\frame{\includegraphics[width=0.33\textwidth]{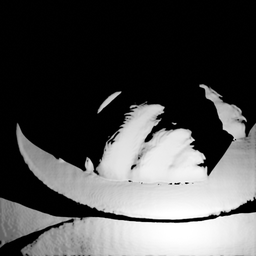}}}{\includegraphics[width=1.0\textwidth]{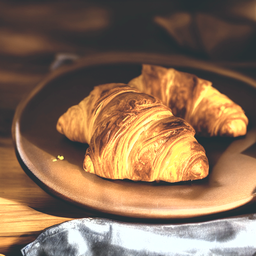}}
    \end{subfigure}
    \begin{subfigure}{.24\textwidth}
    \centering
    \stackinset{r}{0pt}{b}{0pt}{\color{white}\frame{\includegraphics[width=0.33\textwidth]{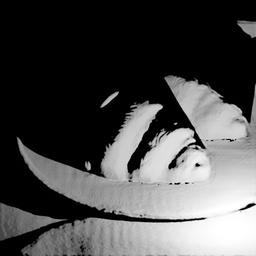}}}{\includegraphics[width=1.0\textwidth]{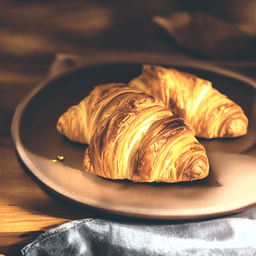}}
    \end{subfigure}
    \begin{subfigure}{.24\textwidth}
    \centering
    \stackinset{r}{0pt}{b}{0pt}{\color{white}\frame{\includegraphics[width=0.33\textwidth]{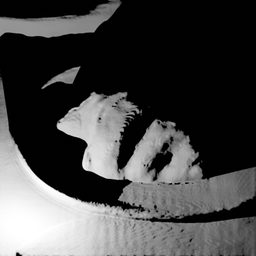}}}{\includegraphics[width=1.0\textwidth]{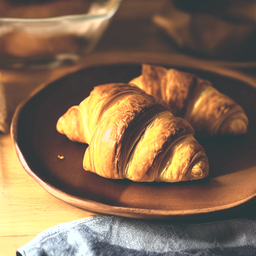}}
    \end{subfigure}
    \begin{subfigure}{.24\textwidth}
    \centering
    \stackinset{r}{0pt}{b}{0pt}{\color{white}\frame{\includegraphics[width=0.33\textwidth]{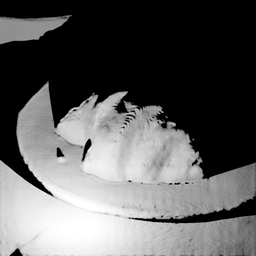}}}{\includegraphics[width=1.0\textwidth]{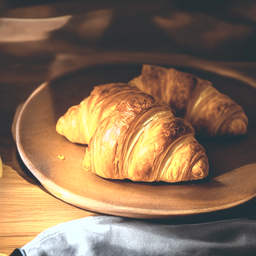}}
    \end{subfigure}
    \begin{subfigure}{.24\textwidth}
    \centering
    \stackinset{r}{0pt}{b}{0pt}{\color{white}\frame{\includegraphics[width=0.33\textwidth]{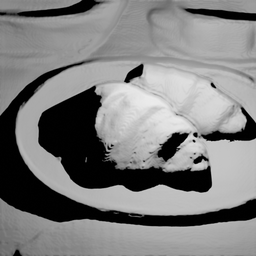}}}{\includegraphics[width=1.0\textwidth]{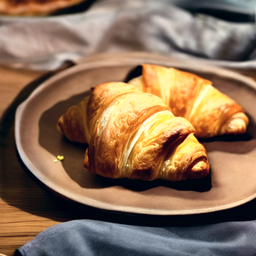}}
    \end{subfigure}
    \begin{subfigure}{.24\textwidth}
    \centering
    \stackinset{r}{0pt}{b}{0pt}{\color{white}\frame{\includegraphics[width=0.33\textwidth]{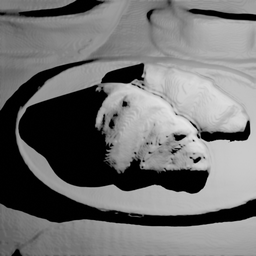}}}{\includegraphics[width=1.0\textwidth]{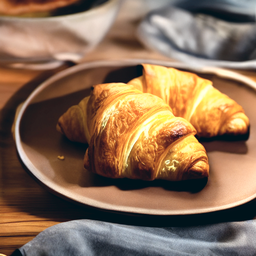}}
    \end{subfigure}
    \begin{subfigure}{.24\textwidth}
    \centering
    \stackinset{r}{0pt}{b}{0pt}{\color{white}\frame{\includegraphics[width=0.33\textwidth]{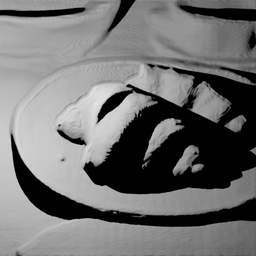}}}{\includegraphics[width=1.0\textwidth]{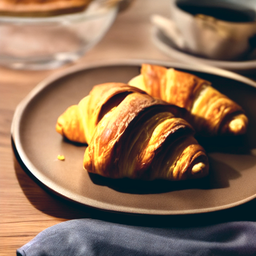}}
    \end{subfigure}
    \begin{subfigure}{.24\textwidth}
    \centering
    \stackinset{r}{0pt}{b}{0pt}{\color{white}\frame{\includegraphics[width=0.33\textwidth]{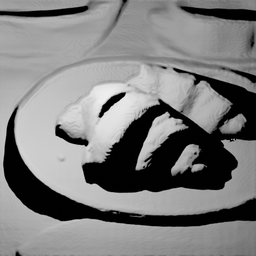}}}{\includegraphics[width=1.0\textwidth]{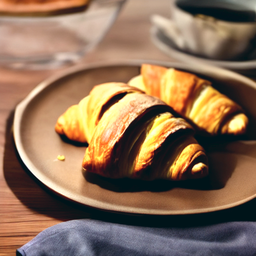}}
    \end{subfigure}
    \begin{subfigure}{.24\textwidth}
    \centering
    \stackinset{r}{0pt}{b}{0pt}{\color{white}\frame{\includegraphics[width=0.33\textwidth]{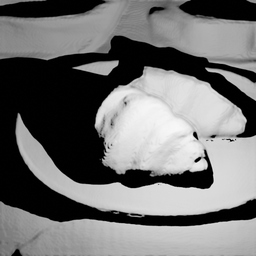}}}{\includegraphics[width=1.0\textwidth]{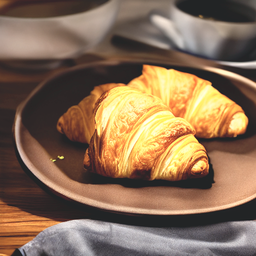}}
    \end{subfigure}
    \begin{subfigure}{.24\textwidth}
    \centering
    \stackinset{r}{0pt}{b}{0pt}{\color{white}\frame{\includegraphics[width=0.33\textwidth]{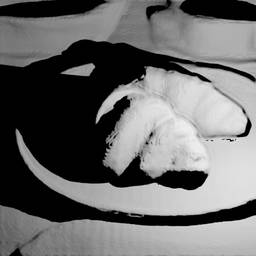}}}{\includegraphics[width=1.0\textwidth]{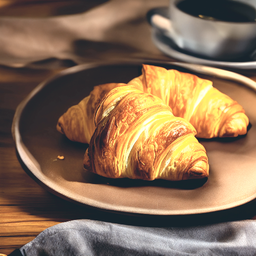}}
    \end{subfigure}
    \begin{subfigure}{.24\textwidth}
    \centering
    \stackinset{r}{0pt}{b}{0pt}{\color{white}\frame{\includegraphics[width=0.33\textwidth]{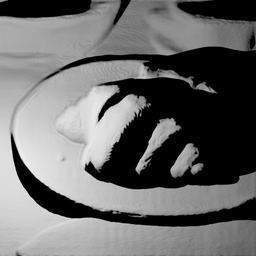}}}{\includegraphics[width=1.0\textwidth]{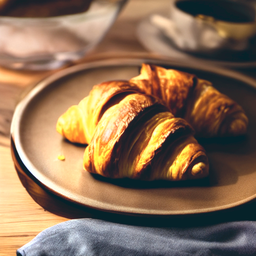}}
    \end{subfigure}
    \begin{subfigure}{.24\textwidth}
    \centering
    \stackinset{r}{0pt}{b}{0pt}{\color{white}\frame{\includegraphics[width=0.33\textwidth]{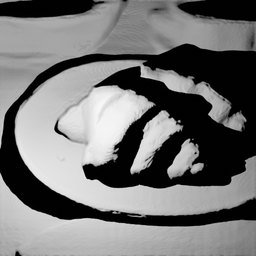}}}{\includegraphics[width=1.0\textwidth]{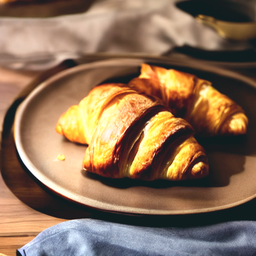}}
    \end{subfigure}

  \caption{\textbf{Light Probe.} Relighting examples using \ourmethod with insets showing $I_c$. Note the hard shadows edits.}
  \label{fig:many_lights2}
\end{figure*}

In \Cref{fig:many_lights_face,fig:many_lights2} we show a single image relit by placing a random light probe in the scene. \ourmethod~ is able to relight the scene consistently with plausible results, synthesizing shadows and highlights.

\begin{figure*}
    \centering
    {
    \setlength{\tabcolsep}{0pt} 
    \renewcommand{\arraystretch}{0} 
    \begin{tabular}{*{2}{>{\centering\arraybackslash}m{0.13\textwidth}} @{\hskip 5pt} *{5}{>{\centering\arraybackslash}m{0.13\textwidth}}}
    Source $I_s$& Condition $I_c$ & Readouts & ControlNet & RGB$\leftrightarrow$x & IC-Light & Ours  
    \end{tabular}
        \begin{tabular}{*{2}{>{\centering\arraybackslash}m{0.13\textwidth}} @{\hskip 5pt} *{5}{>{\centering\arraybackslash}m{0.13\textwidth}}}
        \\[0.5em]
        \includegraphics[width=\linewidth]{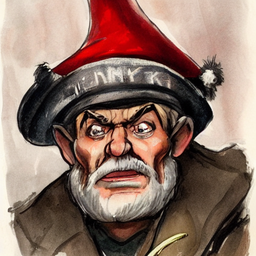} & 
        \includegraphics[width=\linewidth]{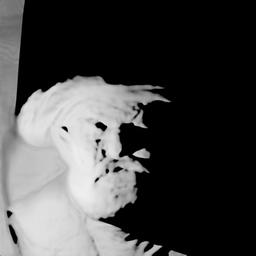} & 
        \includegraphics[width=\linewidth]{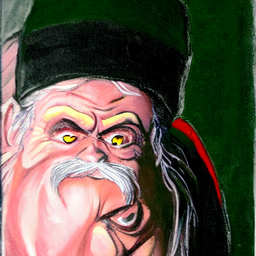} & 
        \includegraphics[width=\linewidth]{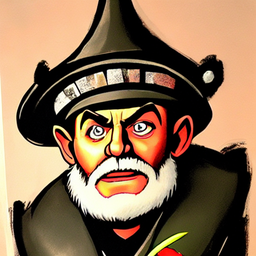} &
        \includegraphics[width=\linewidth]{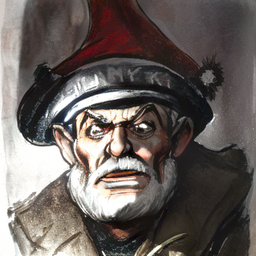} & 
        \includegraphics[width=\linewidth]{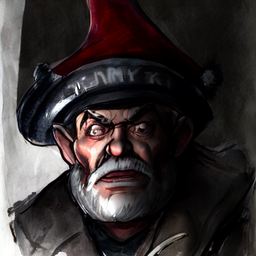} & 
        \includegraphics[width=\linewidth]{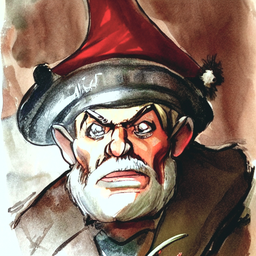} \\ 
        \includegraphics[width=\linewidth]{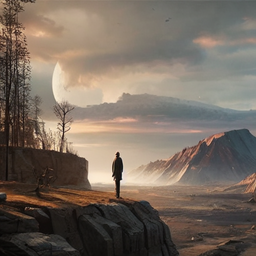} & 
        \includegraphics[width=\linewidth]{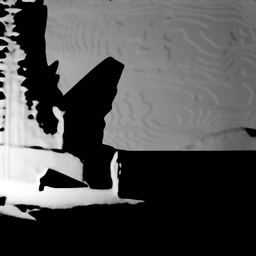} & 
        \includegraphics[width=\linewidth]{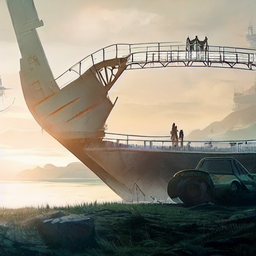} & 
        \includegraphics[width=\linewidth]{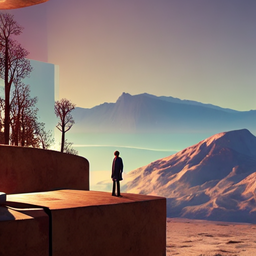} & 
        \includegraphics[width=\linewidth]{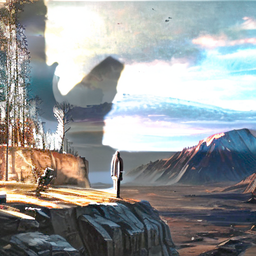} & 
        \includegraphics[width=\linewidth]{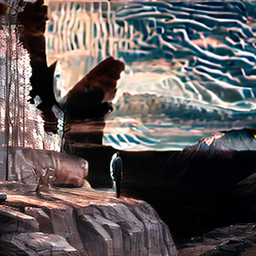} & 
        \includegraphics[width=\linewidth]{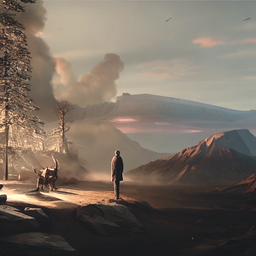} \\
        \includegraphics[width=\linewidth]{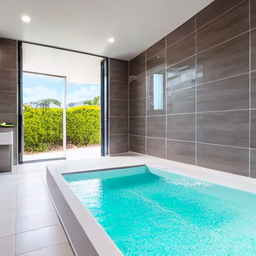} & 
        \includegraphics[width=\linewidth]{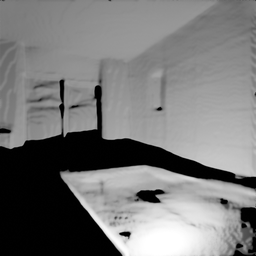} & 
        \includegraphics[width=\linewidth]{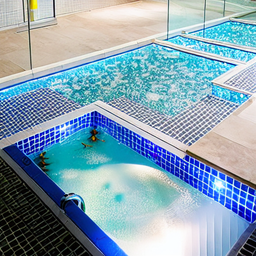} & 
        \includegraphics[width=\linewidth]{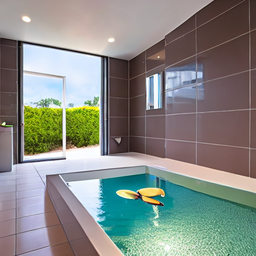} &
        \includegraphics[width=\linewidth]{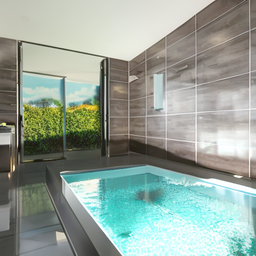} & 
        \includegraphics[width=\linewidth]{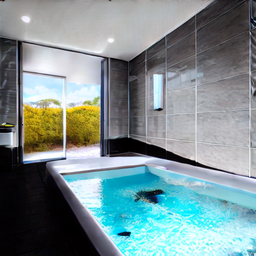} & 
        \includegraphics[width=\linewidth]{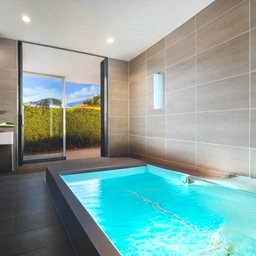} \\
        \includegraphics[width=\linewidth]{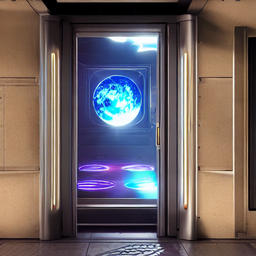} & 
        \includegraphics[width=\linewidth]{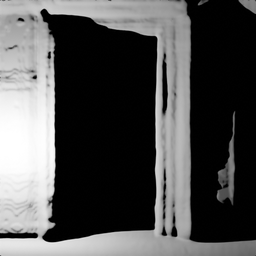} & 
        \includegraphics[width=\linewidth]{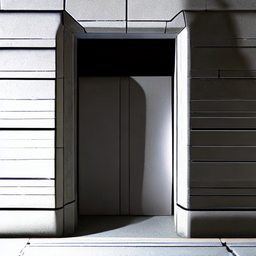} & 
        \includegraphics[width=\linewidth]{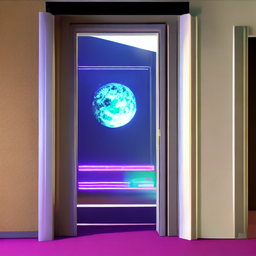} &
        \includegraphics[width=\linewidth]{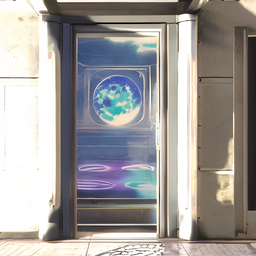} & 
        \includegraphics[width=\linewidth]{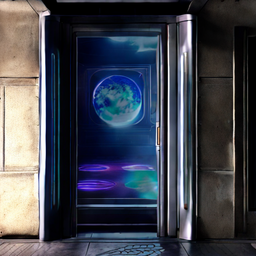} & 
        \includegraphics[width=\linewidth]{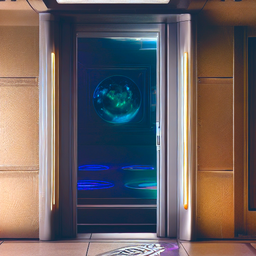} \\
        \includegraphics[width=\linewidth]{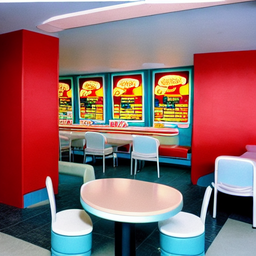} & 
        \includegraphics[width=\linewidth]{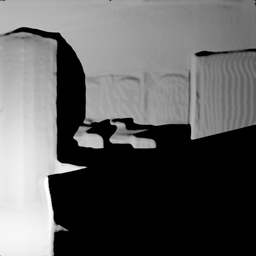} & 
        \includegraphics[width=\linewidth]{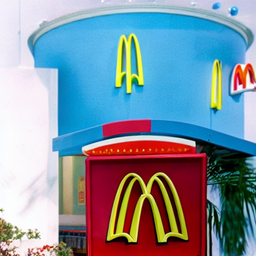} & 
        \includegraphics[width=\linewidth]{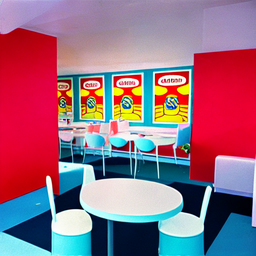} &
        \includegraphics[width=\linewidth]{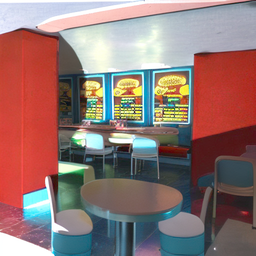} & 
        \includegraphics[width=\linewidth]{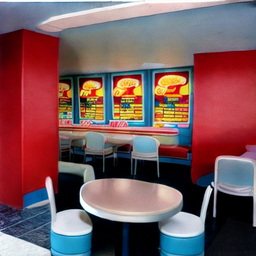} & 
        \includegraphics[width=\linewidth]{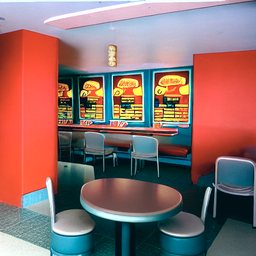} \\
        \includegraphics[width=\linewidth]{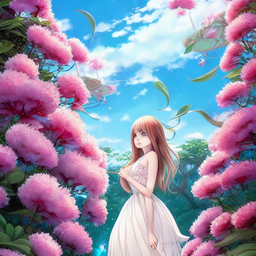} & 
        \includegraphics[width=\linewidth]{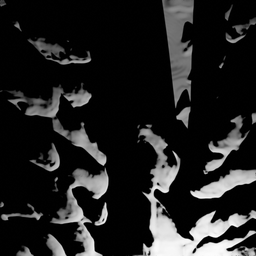} & 
        \includegraphics[width=\linewidth]{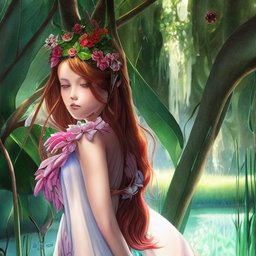} & 
        \includegraphics[width=\linewidth]{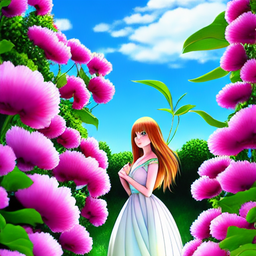} &
        \includegraphics[width=\linewidth]{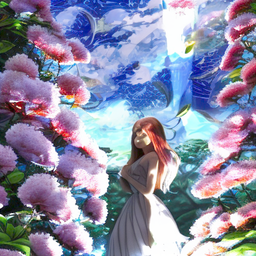} & 
        \includegraphics[width=\linewidth]{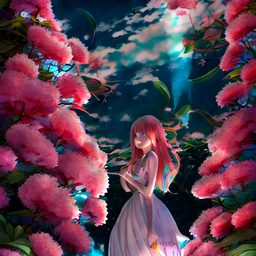} & 
        \includegraphics[width=\linewidth]{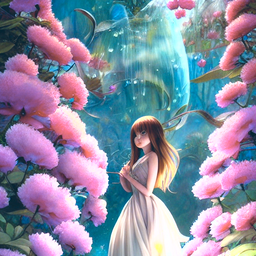} \\ 
        \includegraphics[width=\linewidth]{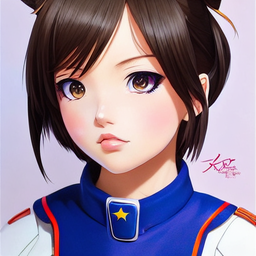} & 
        \includegraphics[width=\linewidth]{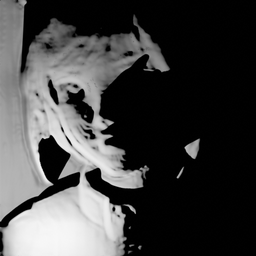} & 
        \raisebox{-.5\height}{
  \begin{tikzpicture}[scale=2.0]
    \draw[red, thick] (0,0) rectangle (1,1); 
    \draw[red, thick] (0,0) -- (1,1);       
    \draw[red, thick] (0,1) -- (1,0);       
  \end{tikzpicture}} & 
        \includegraphics[width=\linewidth]{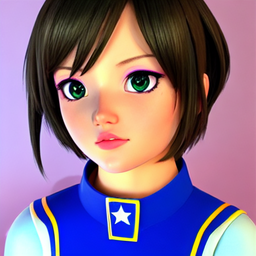} &
        \includegraphics[width=\linewidth]{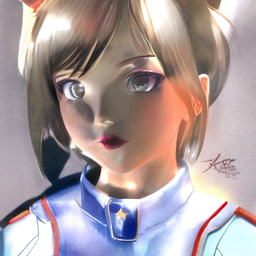} & 
        \includegraphics[width=\linewidth]{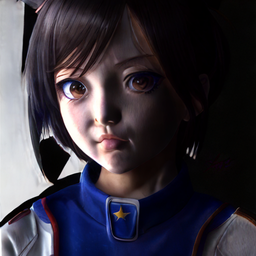} & 
        \includegraphics[width=\linewidth]{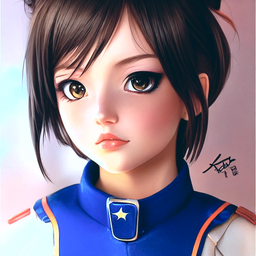} \\
        \includegraphics[width=\linewidth]{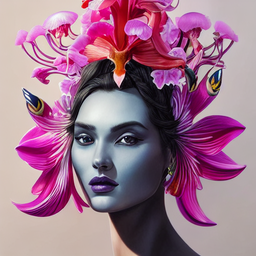} & 
        \includegraphics[width=\linewidth]{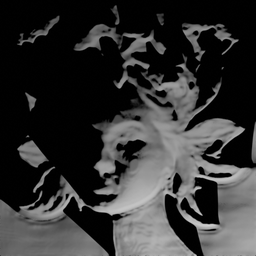} & 
        \includegraphics[width=\linewidth]{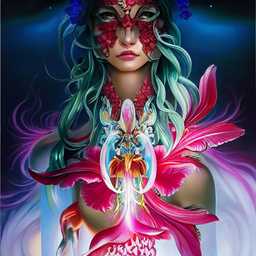} & 
        \includegraphics[width=\linewidth]{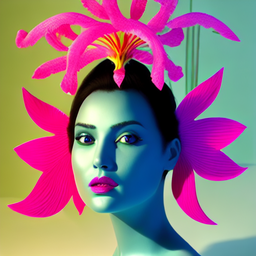} &
        \includegraphics[width=\linewidth]{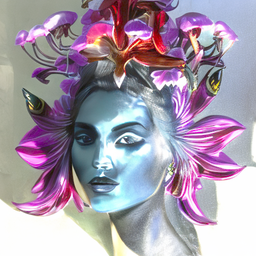} & 
        \includegraphics[width=\linewidth]{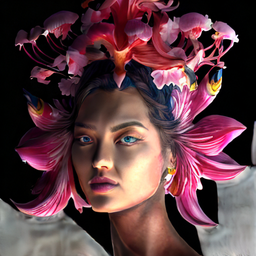} & 
        \includegraphics[width=\linewidth]{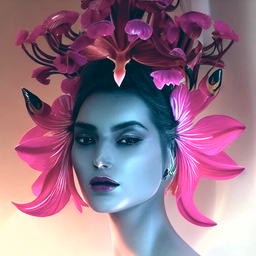} \\
        \includegraphics[width=\linewidth]{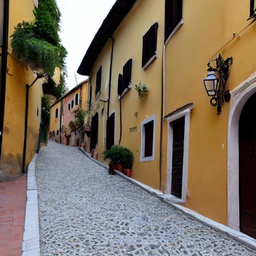} & 
        \includegraphics[width=\linewidth]{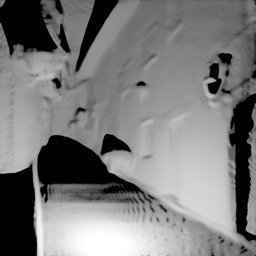} & 
        \includegraphics[width=\linewidth]{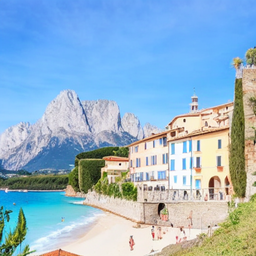} & 
        \includegraphics[width=\linewidth]{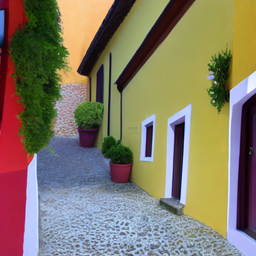} &
        \includegraphics[width=\linewidth]{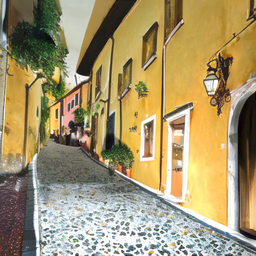} & 
        \includegraphics[width=\linewidth]{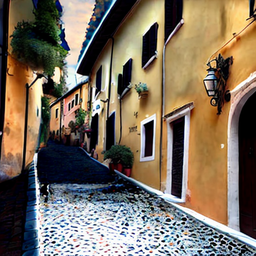} & 
        \includegraphics[width=\linewidth]{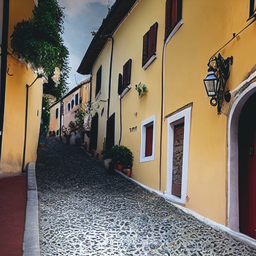} \\ 
    \end{tabular}
    }
    \caption{\textbf{Qualitative Results.} Relighting results on the evaluation dataset. Some images removed due to unsafe content. Zoomed-in viewing recommended.}
    \label{fig:zoo33}
\end{figure*}
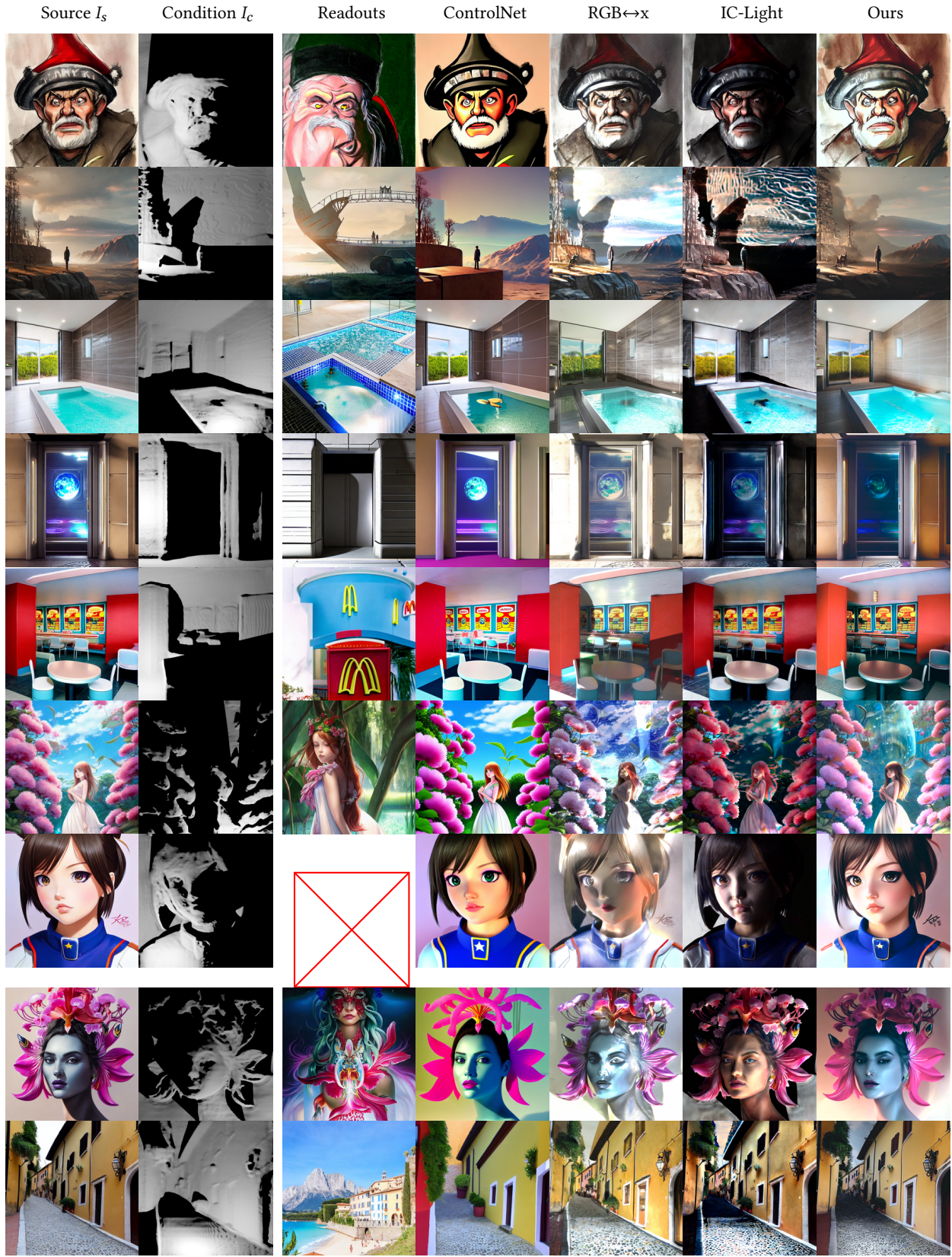

\begin{figure*}
    \centering
    {
    \setlength{\tabcolsep}{0pt} 
    \renewcommand{\arraystretch}{0} 
    \begin{tabular}{*{2}{>{\centering\arraybackslash}m{0.12\textwidth}} @{\hskip 5pt} *{5}{>{\centering\arraybackslash}m{0.12\textwidth}}}
    Source $I_s$& Condition $I_c$ & Readouts & ControlNet & RGB$\leftrightarrow$x & IC-Light & Ours  
    \end{tabular}
        \begin{tabular}{*{2}{>{\centering\arraybackslash}m{0.12\textwidth}} @{\hskip 5pt} *{5}{>{\centering\arraybackslash}m{0.12\textwidth}}}
        \\[0.5em]
        \includegraphics[width=\linewidth]{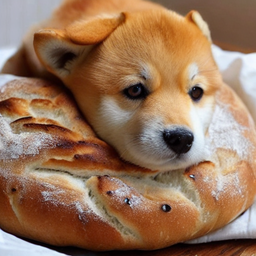} & 
        \includegraphics[width=\linewidth]{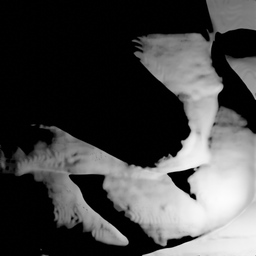} & 
        \includegraphics[width=\linewidth]{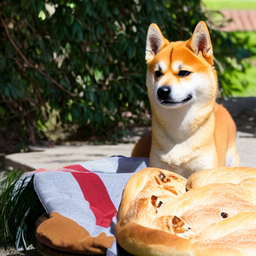} & 
        \includegraphics[width=\linewidth]{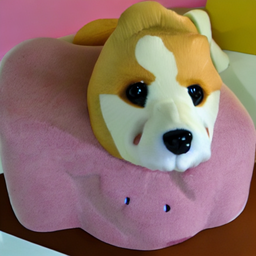} &
        \includegraphics[width=\linewidth]{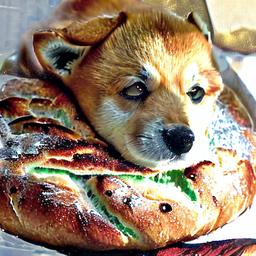} & 
        \includegraphics[width=\linewidth]{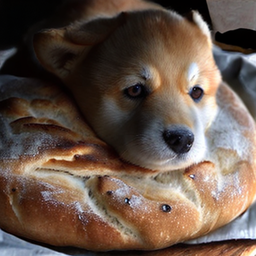} & 
        \includegraphics[width=\linewidth]{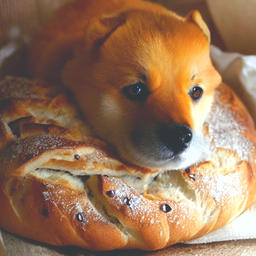} \\ 
        \includegraphics[width=\linewidth]{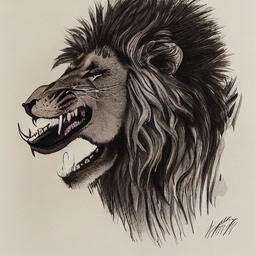} & 
        \includegraphics[width=\linewidth]{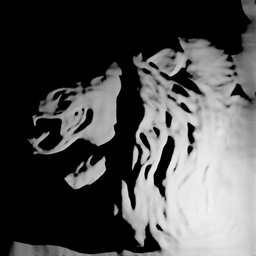} & 
        \includegraphics[width=\linewidth]{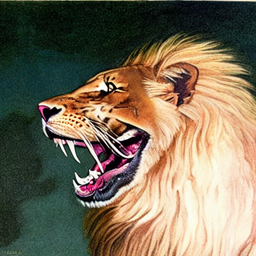} & 
        \includegraphics[width=\linewidth]{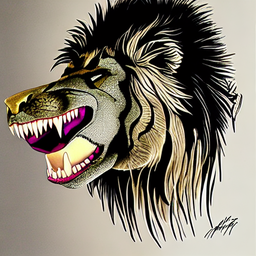} & 
        \includegraphics[width=\linewidth]{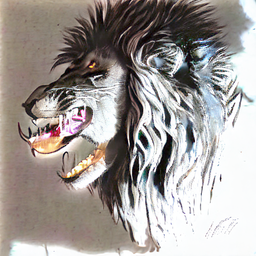} & 
        \includegraphics[width=\linewidth]{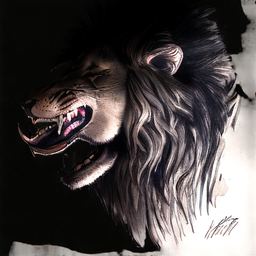} & 
        \includegraphics[width=\linewidth]{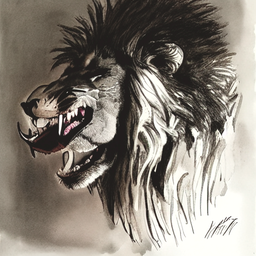} \\
        \includegraphics[width=\linewidth]{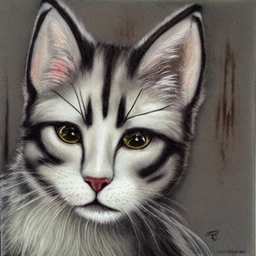} & 
        \includegraphics[width=\linewidth]{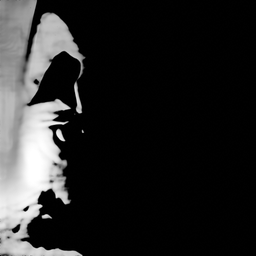} & 
        \includegraphics[width=\linewidth]{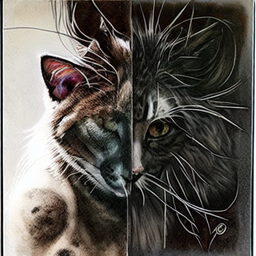} & 
        \includegraphics[width=\linewidth]{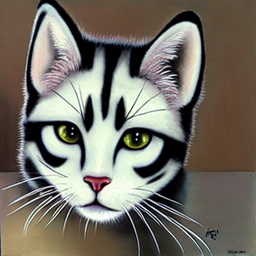} &
        \includegraphics[width=\linewidth]{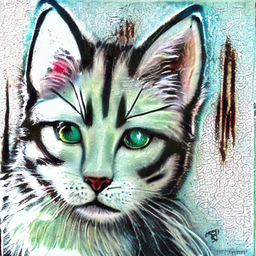} & 
        \includegraphics[width=\linewidth]{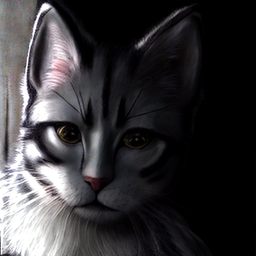} & 
        \includegraphics[width=\linewidth]{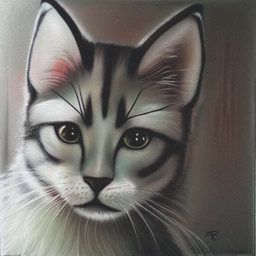} \\
        \includegraphics[width=\linewidth]{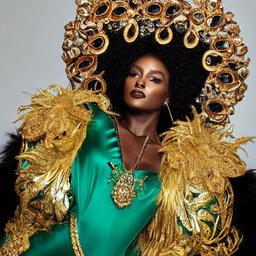} & 
        \includegraphics[width=\linewidth]{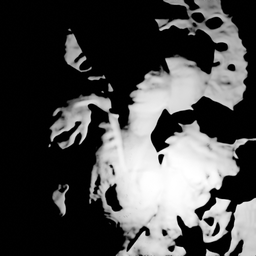} & 
        \includegraphics[width=\linewidth]{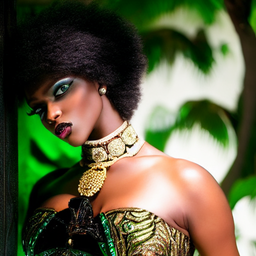} & 
        \includegraphics[width=\linewidth]{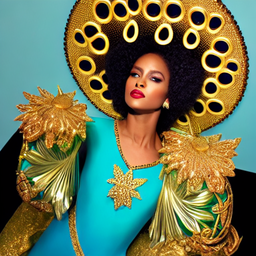} &
        \includegraphics[width=\linewidth]{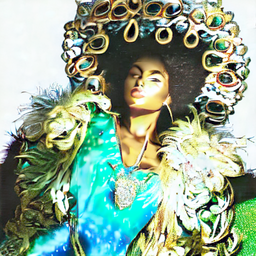} & 
        \includegraphics[width=\linewidth]{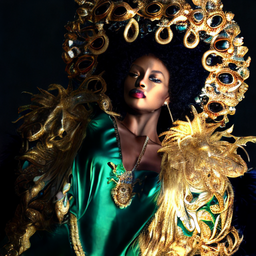} & 
        \includegraphics[width=\linewidth]{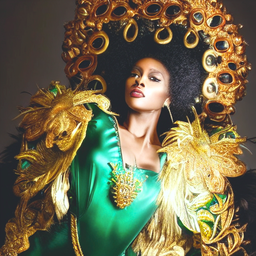} \\
        \includegraphics[width=\linewidth]{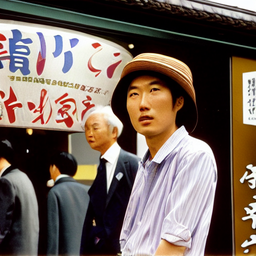} & 
        \includegraphics[width=\linewidth]{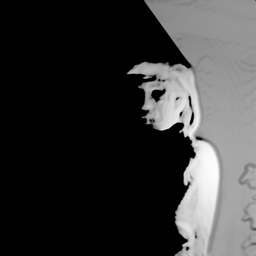} & 
        \includegraphics[width=\linewidth]{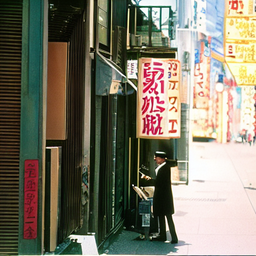} & 
        \includegraphics[width=\linewidth]{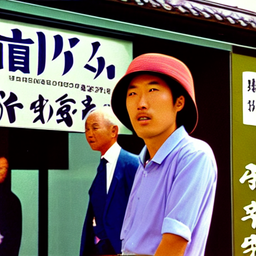} &
        \includegraphics[width=\linewidth]{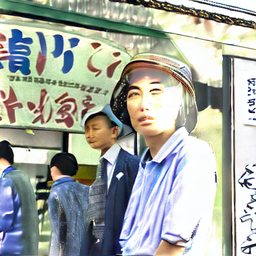} & 
        \includegraphics[width=\linewidth]{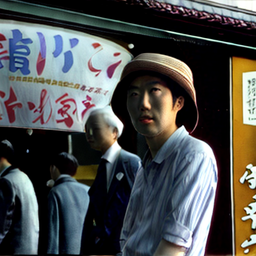} & 
        \includegraphics[width=\linewidth]{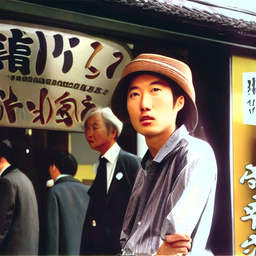} \\
        \includegraphics[width=\linewidth]{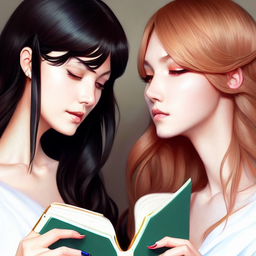} & 
        \includegraphics[width=\linewidth]{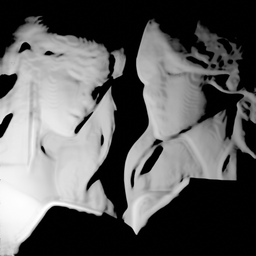} & 
        \includegraphics[width=\linewidth]{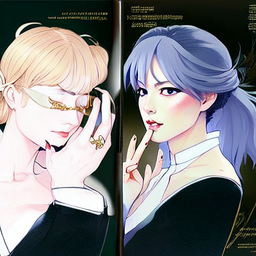} & 
        \includegraphics[width=\linewidth]{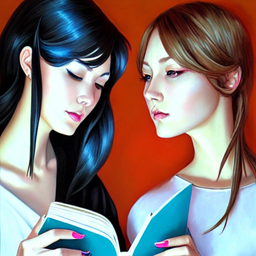} &
        \includegraphics[width=\linewidth]{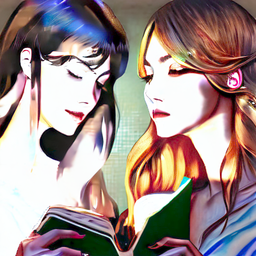} & 
        \includegraphics[width=\linewidth]{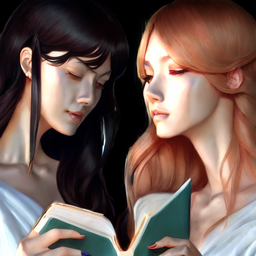} & 
        \includegraphics[width=\linewidth]{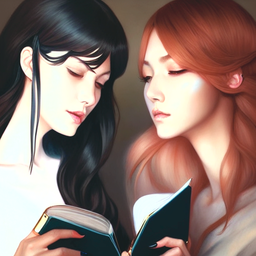} \\
    \end{tabular}
    }
    \caption{\textbf{Qualitative Results.} Relighting results on the evaluation dataset. Zoomed-in viewing recommended.}
    \label{fig:zoo4}
\end{figure*}

\begin{figure*}
    \centering
    {
    \setlength{\tabcolsep}{0pt} 
    \renewcommand{\arraystretch}{0} 
    \begin{tabular}{*{2}{>{\centering\arraybackslash}m{0.10\textwidth}} @{\hskip 1pt} *{2}{>{\centering\arraybackslash}m{0.10\textwidth}} @{\hskip 1pt}*{2}{>{\centering\arraybackslash}m{0.10\textwidth}} @{\hskip 1pt}*{2}{>{\centering\arraybackslash}m{0.10\textwidth}} @{\hskip 1pt}}
    DiLightNet & Ours & DiLightNet & Ours & DiLightNet & Ours & DiLightNet & Ours
    \end{tabular}
        \begin{tabular}{*{2}{>{\centering\arraybackslash}m{0.10\textwidth}} @{\hskip 1pt} *{2}{>{\centering\arraybackslash}m{0.10\textwidth}} @{\hskip 1pt}*{2}{>{\centering\arraybackslash}m{0.10\textwidth}} @{\hskip 1pt}*{2}{>{\centering\arraybackslash}m{0.10\textwidth}} @{\hskip 1pt}}
        \\[0.5em]
        \includegraphics[width=\linewidth]{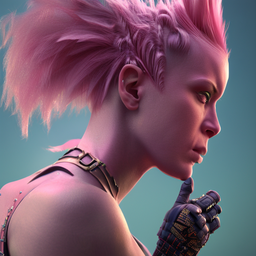} & 
        \includegraphics[width=\linewidth]{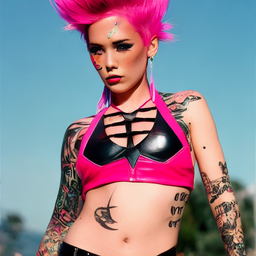} & 
        \includegraphics[width=\linewidth]{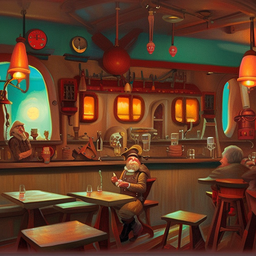} &
        \includegraphics[width=\linewidth]{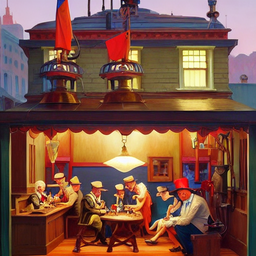} & 
        \includegraphics[width=\linewidth]{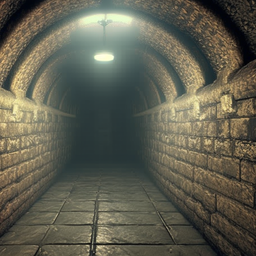} & 
        \includegraphics[width=\linewidth]{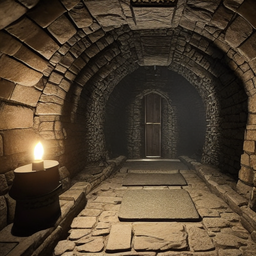} & 
        \includegraphics[width=\linewidth]{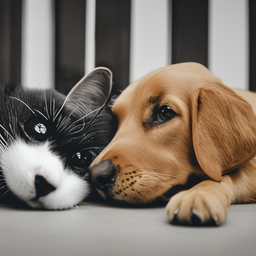} &
        \includegraphics[width=\linewidth]{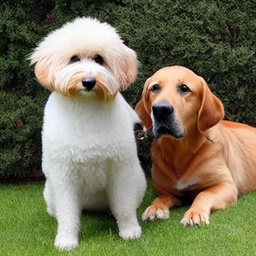} \\ 
        \includegraphics[width=\linewidth]{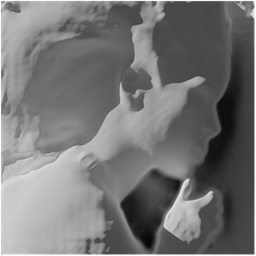} & 
        \includegraphics[width=\linewidth]{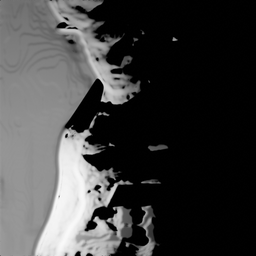} & 
        \includegraphics[width=\linewidth]{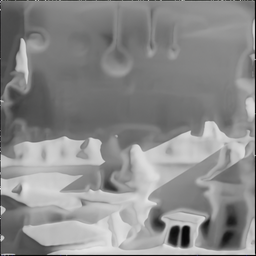} &
        \includegraphics[width=\linewidth]{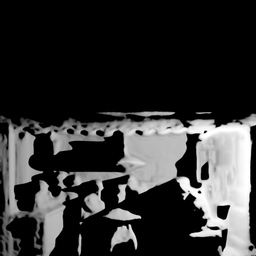} & 
        \includegraphics[width=\linewidth]{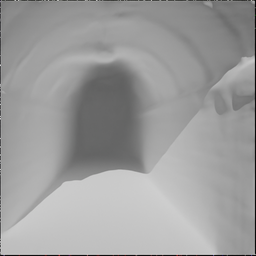} & 
        \includegraphics[width=\linewidth]{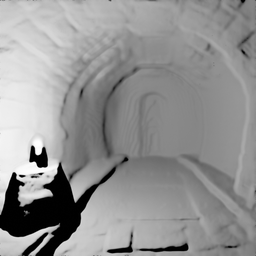} & 
        \includegraphics[width=\linewidth]{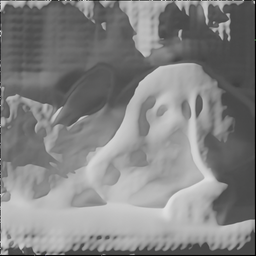} &
        \includegraphics[width=\linewidth]{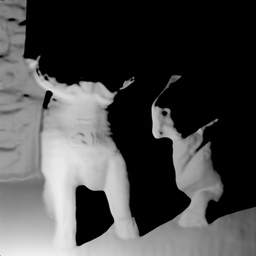} \\
        \includegraphics[width=\linewidth]{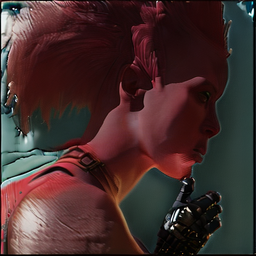} & 
        \includegraphics[width=\linewidth]{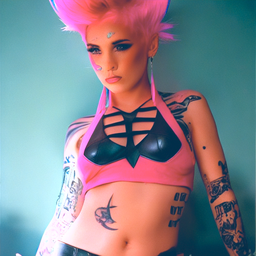} & 
        \includegraphics[width=\linewidth]{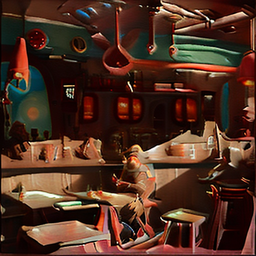} &
        \includegraphics[width=\linewidth]{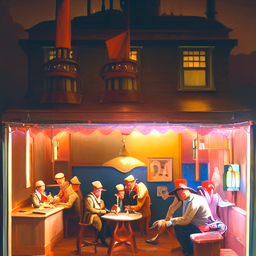} & 
        \includegraphics[width=\linewidth]{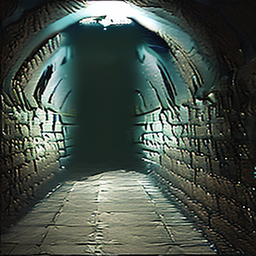} & 
        \includegraphics[width=\linewidth]{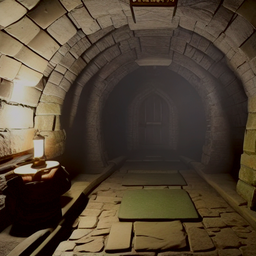} & 
        \includegraphics[width=\linewidth]{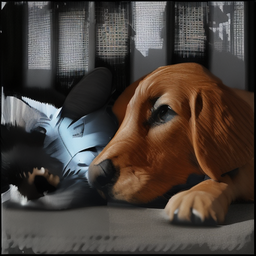} &
        \includegraphics[width=\linewidth]{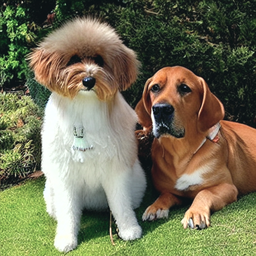} \\
    \end{tabular}
    }
    \caption{\textbf{DiLightNet Comparisons. Top:} source $I_s$, \textbf{middle:} condition $I_c$, \textbf{bottom:} relit $I_r$. Zoomed-in viewing recommended.}
    \label{fig:zoo5}
\end{figure*}

In \Cref{fig:zoo33,fig:zoo4,fig:zoo5} we show more comparative results, all generated using the same parameters as in the main paper.

\section{Head Activations Analysis}
\begin{figure}[h]
  \centering
  \includegraphics[width=0.99\linewidth]{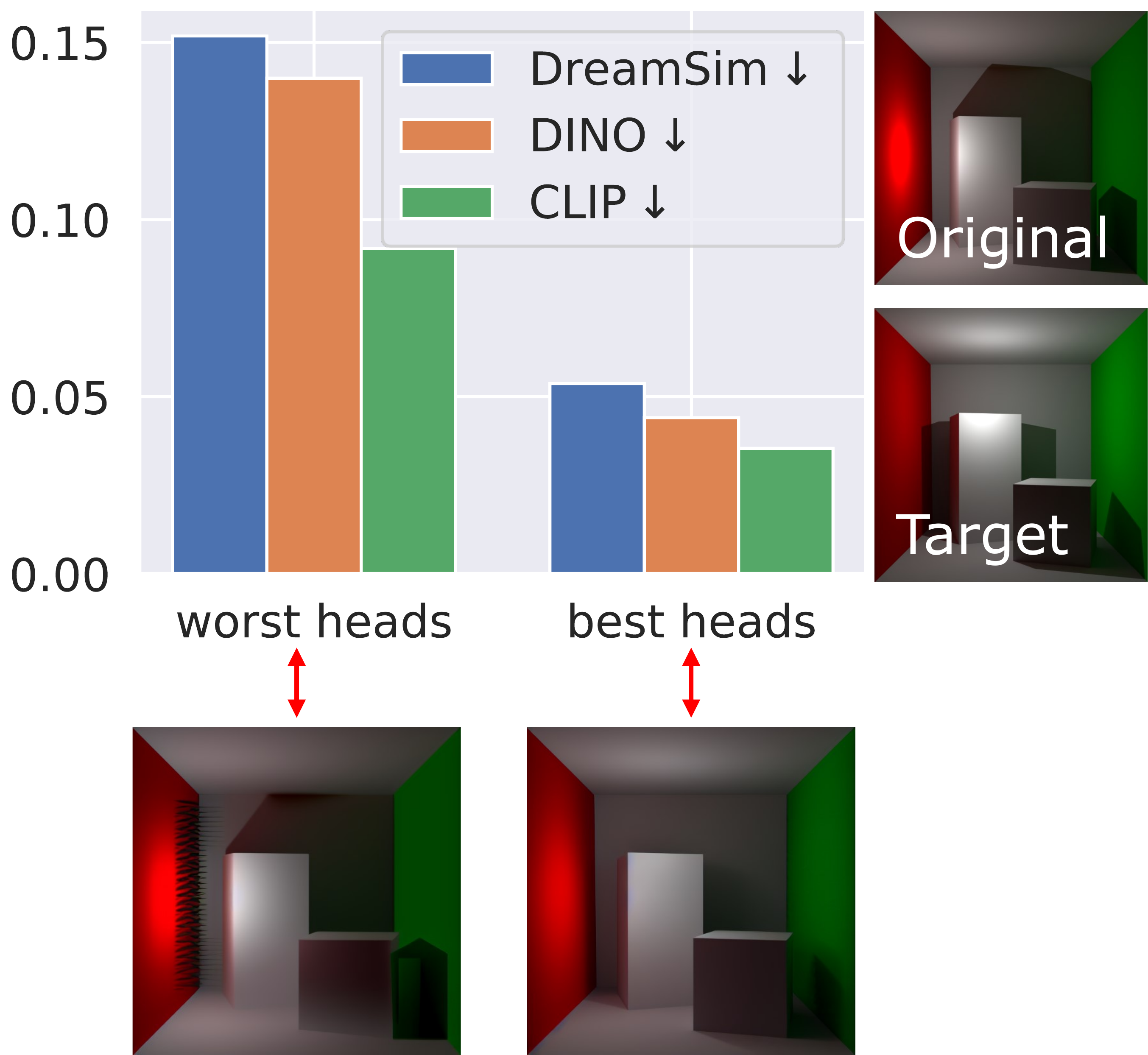}
  \caption{\textbf{Head Activation Analysis.} The self-attention heads associated with the largest regressor change ("best heads") and the smallest change ("worst heads") were injected with features from \textit{Target} when denoising \textit{Original}. Our regressor correctly altered heads which correlate with light interaction.}
  \label{fig:head_analysis}
\end{figure}

A rather unexplored yet important granularity level is the per-attention-head correlation with light (SD 1.5 has 8 heads per layer). The different heads in the self-attention layers are believed to encode different information because they are exposed to different parts of the input features. However, whether or not this holds true for light is unknown.
After training the LoRA regressor to relight, we compute the norm of the expanded weights per head inside the adapter: $||\Delta W||^2_2 = ||AB||^2_2$. This follows the assumption that heads which changed more after optimization, must correlate with light. We then followed the same experiment described in Sec 4.1 in the paper, but injected features only to the top performing head (highest norm) of each self-attention layer, versus injecting features to the worst performing head per layer (\Cref{fig:head_analysis}). The results suggest it may be beneficial to work with finer level of granularity (i.e. optimizing specific heads rather than full layers), and also that our optimization yielded a regressor tapping into the right place.

\section{Stability Under Reconstruction}
\begin{figure}[h]
\centering
{
\setlength{\tabcolsep}{1pt} 
\renewcommand{\arraystretch}{0.5} 
\begin{tabular}{ccc}
Source $I_s$ & Ours & IC-Light \\
[0.5em]
\includegraphics[width=0.15\textwidth]{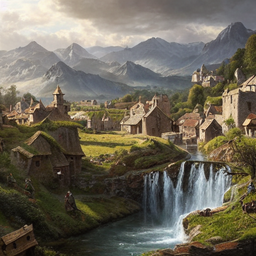} &
\includegraphics[width=0.15\textwidth]{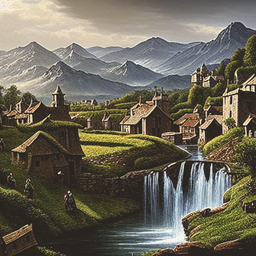} &
\includegraphics[width=0.15\textwidth]{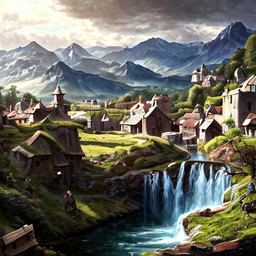}
\end{tabular}
}
\caption{\textbf{Reconstruction.} The source $I_s$ is used as the condition $I_c$, showing stability in the relighting process.}
\label{fig:stable}
\end{figure}
We show in \Cref{fig:stable} an example of relighting an image using itself as a control signal. The result is a near identical image, and actually demonstrates sharper details. While this is not a necessary outcome for good relighting, this does attest to the reconstruction stability of \ourmethod.

\section{Creating the Control Signal}
\begin{figure}[h]
  \centering
    {
    \setlength{\tabcolsep}{1pt} 
    \renewcommand{\arraystretch}{0.5} 
    \begin{tabular}{cccc}
    Source $I_s$ & Depth & Displacement & Condition $I_c$\\
    [0.5em]
    \includegraphics[width=0.25\linewidth]{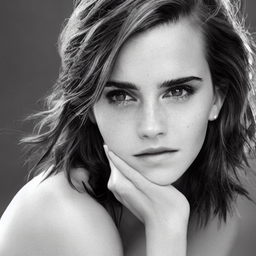} &
    \includegraphics[width=0.25\linewidth]{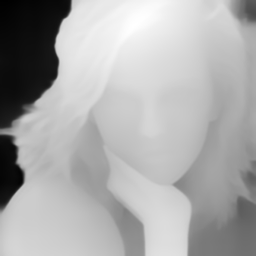} &
    \includegraphics[width=0.25\linewidth]{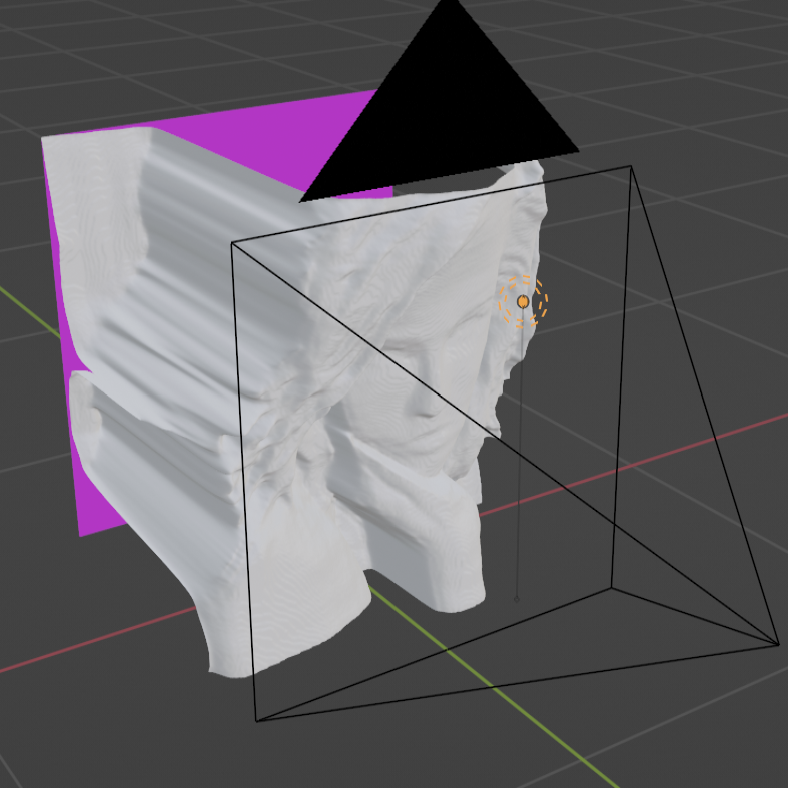} &
    \includegraphics[width=0.25\linewidth]{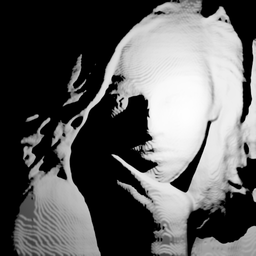}
    \end{tabular}
    }
  \caption{\textbf{Creating the Control Signal.} The process of creating $I_c$ is depicted. We extract a depth image from the source $I_s$, followed by displacing a fine mesh, and rendering the direct light channel using an orthographic camera.}
  \label{fig:lantern_control}
\end{figure}

See \Cref{fig:lantern_control} for an illustration of creating a control signal. We use Blender \cite{blender} to displace a mesh using an estimated depth map, and render the direct light map $I_c$.

\section{Cartoonification}
\begin{figure*}
    \centering
    {
    \setlength{\tabcolsep}{0pt} 
    \renewcommand{\arraystretch}{0} 
    \begin{tabular}{*{1}{>{\centering\arraybackslash}m{0.14\textwidth}} @{\hskip 5pt} *{6}{>{\centering\arraybackslash}m{0.14\textwidth}}}
    Source $I_s$& Scale=$0.0$ & Scale=$0.2$ & Scale=$0.6$ & Scale=$1.0$ & Scale=$2.0$ & Scale=$4.0$
    \end{tabular}
        \begin{tabular}{*{1}{>{\centering\arraybackslash}m{0.14\textwidth}} @{\hskip 5pt} *{6}{>{\centering\arraybackslash}m{0.14\textwidth}}}
        \\[0.5em]
        \begin{tikzpicture}[spy using outlines={circle,red,magnification=3,size=1.5cm, connect spies}]
        \node [inner sep=0pt, anchor=center]{\stackinset{l}{0pt}{t}{0pt}{\color{white}\frame{\includegraphics[width=0.33\linewidth]{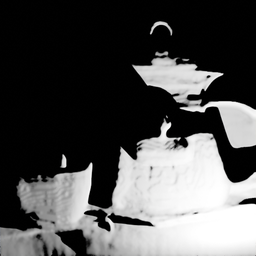}}}{\includegraphics[width=1.0\linewidth]{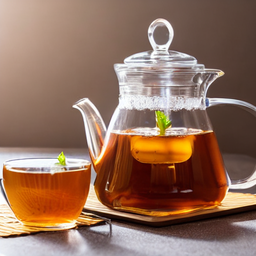}}};
        \spy on (-0.3,-0.1) in node [left] at (0.5,-1.5);
        \end{tikzpicture} & 
        \begin{tikzpicture}[spy using outlines={circle,red,magnification=3,size=1.5cm, connect spies}]
        \node [inner sep=0pt, anchor=center]{\includegraphics[width=1.0\linewidth]{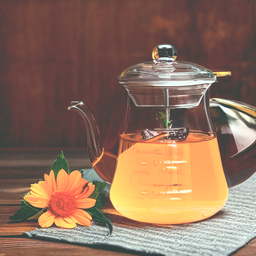}};
        \spy on (-0.3,-0.1) in node [left] at (0.5,-1.5);
        \end{tikzpicture} & 
        \begin{tikzpicture}[spy using outlines={circle,red,magnification=3,size=1.5cm, connect spies}]
        \node [inner sep=0pt, anchor=center]{\includegraphics[width=1.0\linewidth]{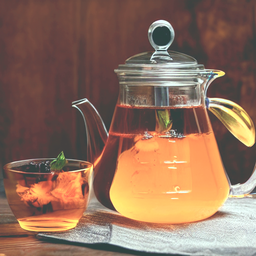}};
        \spy on (-0.3,-0.1) in node [left] at (0.5,-1.5);
        \end{tikzpicture} & 
        \begin{tikzpicture}[spy using outlines={circle,red,magnification=3,size=1.5cm, connect spies}]
        \node [inner sep=0pt, anchor=center]{\includegraphics[width=1.0\linewidth]{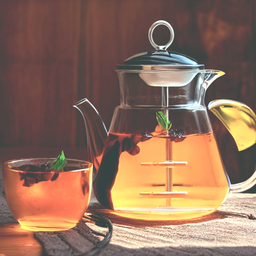}};
        \spy on (-0.3,-0.1) in node [left] at (0.5,-1.5);
        \end{tikzpicture} & 
        \begin{tikzpicture}[spy using outlines={circle,red,magnification=3,size=1.5cm, connect spies}]
        \node [inner sep=0pt, anchor=center]{\includegraphics[width=1.0\linewidth]{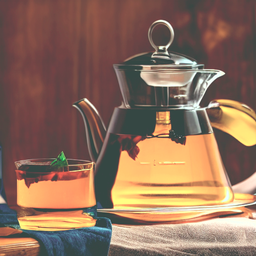}};
        \spy on (-0.3,-0.1) in node [left] at (0.5,-1.5);
        \end{tikzpicture} & 
        \begin{tikzpicture}[spy using outlines={circle,red,magnification=3,size=1.5cm, connect spies}]
        \node [inner sep=0pt, anchor=center]{\includegraphics[width=1.0\linewidth]{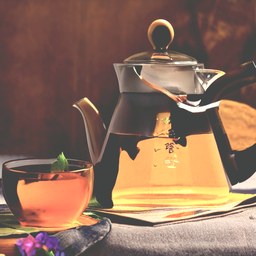}};
        \spy on (-0.3,-0.1) in node [left] at (0.5,-1.5);
        \end{tikzpicture} & 
        \begin{tikzpicture}[spy using outlines={circle,red,magnification=3,size=1.5cm, connect spies}]
        \node [inner sep=0pt, anchor=center]{\includegraphics[width=1.0\linewidth]{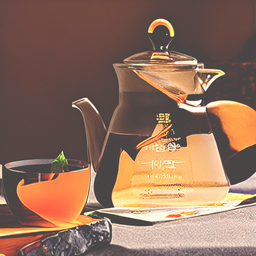}};
        \spy on (-0.3,-0.1) in node [left] at (0.5,-1.5);
        \end{tikzpicture} \\
        \stackinset{l}{0pt}{t}{0pt}{\color{white}\frame{\includegraphics[width=0.33\linewidth]{images/floral_tea/guidance.png}}}{\includegraphics[width=1.0\linewidth]{images/floral_tea/orig.png}} & 
        \includegraphics[width=1.0\linewidth]{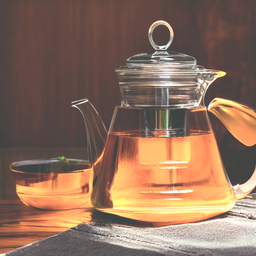}& 
        \includegraphics[width=1.0\linewidth]{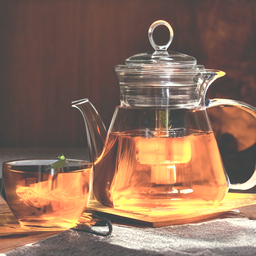} & 
        \begin{tikzpicture}
          \node[inner sep=0pt] (img) at (0,0) {\includegraphics[width=1.0\linewidth]{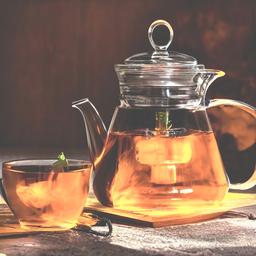}};
          \node[anchor=north west, text=white, font=\normalsize] at (img.north west) {Ours};
        \end{tikzpicture} & 
        \includegraphics[width=1.0\linewidth]{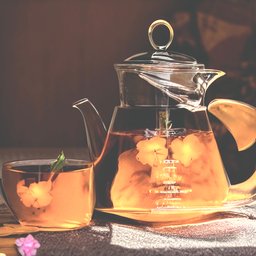} & 
        \includegraphics[width=1.0\linewidth]{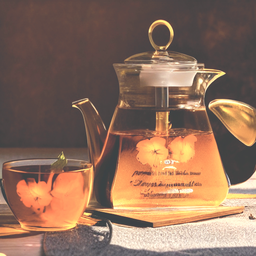}& 
        \includegraphics[width=1.0\linewidth]{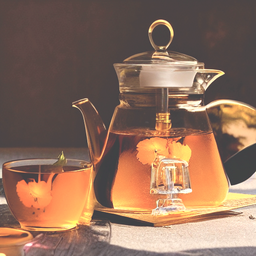} \\
    \end{tabular}
    }
    \caption{\textbf{Cartoonification. Top:} the source $I_s$ (and condtiion $I_c$, inset), relit by our method with different ControlNet guidance scales without query injection. \textbf{Bottom:} with query injection. ControlNet alone somewhat preserves the structure but causes cartoonification and color flattening. Adding query injection helps remove these artifacts at all ControlNet guidance scales, and preserve identity better.}
    \label{fig:cartoon}
\end{figure*}

In \Cref{fig:cartoon}, we show results of various combinations of ControlNet guidance scale with and without query injection. Without query injection, results tend to cartoonify as described in the main paper, whereas with no controlnet, identity preservation is harmed.

\section{Guidance Scheduling}
\begin{figure*}
    \centering
    {
    \setlength{\tabcolsep}{0pt} 
    \renewcommand{\arraystretch}{0} 
    \begin{tabular}{*{1}{>{\centering\arraybackslash}m{0.19\textwidth}} @{\hskip 5pt} *{4}{>{\centering\arraybackslash}m{0.19\textwidth}}}
    Source $I_s$& $t=[0.05, 0.5]$ & $t=[0.25, 0.75]$ & $t=[0.5, 1.0]$ & $t=[0.0, 1.0]$ 
    \end{tabular}
    \begin{tabular}{*{1}{>{\centering\arraybackslash}m{0.19\textwidth}} @{\hskip 5pt} *{4}{>{\centering\arraybackslash}m{0.19\textwidth}}}
        \stackinset{r}{0pt}{b}{0pt}{\color{white}\frame{\includegraphics[width=0.33\linewidth]{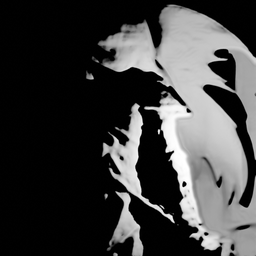}}}{\includegraphics[width=1.0\linewidth]{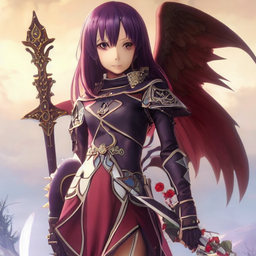}} &
        \begin{tikzpicture}
          \node[inner sep=0pt] (img) at (0,0) {\includegraphics[width=1.0\linewidth]{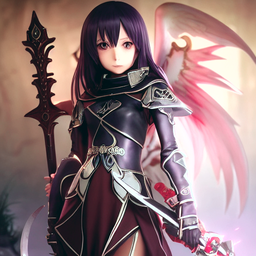}};
          \node[anchor=north west, text=white, font=\normalsize] at (img.north west) {Ours};
        \end{tikzpicture} & 
        \includegraphics[width=1.0\linewidth]{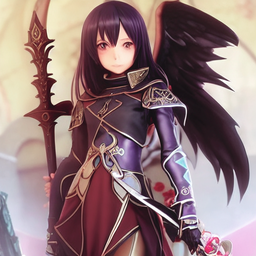} &
        \includegraphics[width=1.0\linewidth]{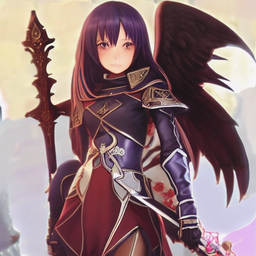} &
        \includegraphics[width=1.0\linewidth]{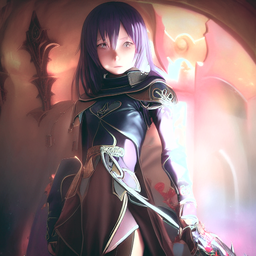}
    \end{tabular}
    }
    \caption{\textbf{Guidance Scheduling.} We use different subsets of timesteps $t$ (normalized timestep) to perform guidance. The earlier guidance is applied, the better light transfers from the condition $I_c$ (left, inset). However too much guidance fails to preserve identity (e.g. $t=[0.0,1.0]$). All examples were created using the same hyper-parameters.}
    \label{fig:timesteps}
\end{figure*}

In \Cref{fig:timesteps}, we show a source image $I_s$ relit using different guidance timesteps. In all our observations starting the edits early (when the image is still noisy) is beneficial for overall relighting performance, consistent with the analysis in the main paper. Applying guidance through all timesteps severely degrades identity preservation.

\end{document}